\documentclass[
	english,	
	BCOR,		
]{WissTemplate}

\usepackage{bookmark}

\renewcommand{\name}{Md Motiur Rahman Sagar}
\renewcommand{\matrikel}{Matriculation number: 217205651}
\renewcommand{\course}{M.Sc. in Electrical Engineering \\(specialization in Information Technology)}
\renewcommand{\supervisor}{Dr. rer. hum. Martin Dyrba\\Prof. Dr.-Ing. Thomas Kriste}
\renewcommand{\type}{Master Thesis} 
\renewcommand{\Title}{Learning Shape Features and \\Abstractions in 3D Convolutional Neural Networks for Detecting Alzheimer’s Disease}

\newcommand{\deliveryDate}{04.08.2020}

\begin{document}
\pdfbookmark[0]{Title Page}{title page}	
	\begin{titlepage}
		\newlength{\BCORoffset}
		\ifdefined\BCOR
			\setlength{\BCORoffset}{0.5cm}
		\else
			\setlength{\BCORoffset}{0cm}
		\fi
		\linespread{1.2} 
		\begin{tikzpicture}[remember picture,overlay, anchor = west]
			\coordinate (d) at ([yshift=-6cm, xshift=-2cm+\BCORoffset] current page.north east);
			\coordinate (c) at ([yshift=-6cm, xshift=2cm+\BCORoffset] current page.north west);
			\coordinate (b) at ([yshift=2.25cm, xshift=-2cm+\BCORoffset] current page.south east);
			\coordinate (a) at ([yshift=2.25cm, xshift=2cm+\BCORoffset] current page.south west);
			\coordinate (middle) at ([yshift=-6cm, xshift=\paperwidth/2+\BCORoffset] current page.north west);
			\draw[line width=2,rounded corners=10pt,uniblau] (a) -- (c) -- (d) -- (b);
			\node[anchor=west,inner sep=0,outer sep=0] (logo) at ([yshift=-3cm, xshift=2cm+\BCORoffset] current page.north west) {
				\includegraphics[width=115mm]{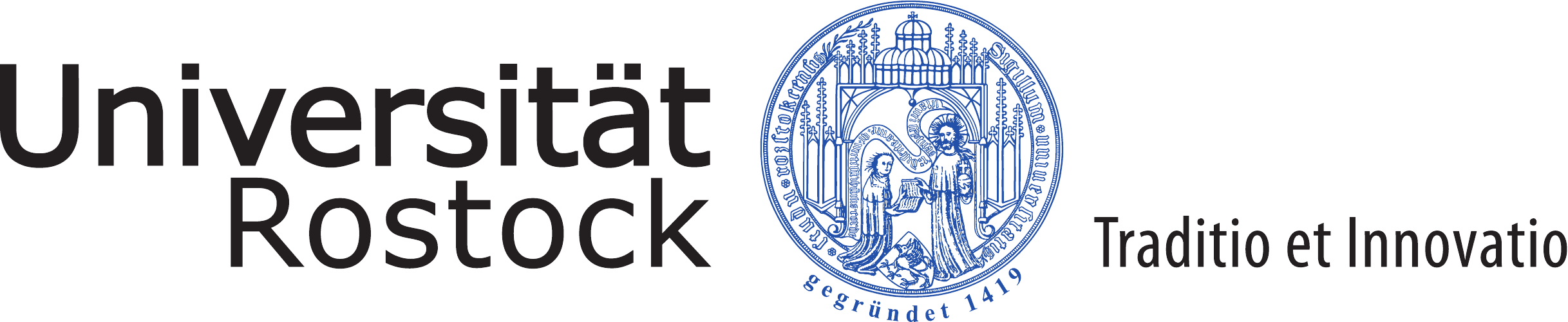}
			};
			\node[fill=uniblau,draw=uniblau,line width=2,anchor=north,minimum height=1.5cm, minimum width=\paperwidth-4cm,text=white,font=\footnotesize,text width=\paperwidth-4.5cm] (names)
			at ([yshift=2.25cm,xshift=\BCORoffset]current page.south) {
				\iflanguage{english}{
					\textsc{Universität Rostock} \textbar \ \textsc{Department of Computer Science and Electrical Engineering}\\
				}	
				
			};
			\node[font=\Huge\bfseries,anchor=center,] (typenode) at ([yshift=-3cm] middle) {\type}; 
			\node[font=\Large\bfseries,anchor=center,text width=\paperwidth-5.5cm,align=center] (titelnode) at ([yshift=-2.5cm] typenode) {\Title}; 
			\node[font=\large,anchor=center,text width=\paperwidth-4.5cm,align=center] (authornode) at ([yshift=-3.5cm] titelnode) {
				\iflanguage{english}{
					\textsc{Submitted by}:\\\bfseries\name\\\mdseries \textsc{\matrikel}\\\course
				}{
					\textsc{Vorgelegt von}:\\\bfseries\name\\\mdseries \textsc{\matrikel}\\\course
				}	
			};
			\node[font=\large,anchor=center,text width=\paperwidth-4.5cm,align=center] (abgabenode) at ([yshift=-4.5cm] authornode) {
				\iflanguage{english}{
					\textsc{Submitted on:}\\\deliveryDate
				}{
					\textsc{Eingereicht am:}\\\deliveryDate
				}			
			};
			\node[font=\large,anchor=center,text width=\paperwidth-4.5cm,align=center] (supervisornode) at ([yshift=-4.0cm] abgabenode) {
				\iflanguage{english}{
					\textsc{Supervisor:}\\\supervisor
				}{
					\textsc{Betreuer:}\\\supervisor
				}
			};
		\end{tikzpicture}
	\end{titlepage}
	\restoregeometry

	\frontmatter


	\newpage
	\chapter{Approval}\label{approval}	
	The thesis titled “Learning Shape Features and Abstractions in 3D Convolutional Neural Networks for Detecting Alzheimer’s Disease” has been submitted by \textit{Md Motiur Rahman Sagar} (Matrikel-Nr.: 217205651) of the Faculty of Computer Science and Electrical Engineering, Universität Rostock, in partial fulfillment of the requirement for the degree of Master of Science in Electrical Engineering (EE) on August 2020 and has been accepted as satisfactory.

\vspace{1.8cm}
	
\noindent
{\large \textbf{Supervisor:} \par}

\vspace{0.5cm}
\noindent
\rule{10cm}{0.5pt}\\
{\large \textbf{Dr. rer. hum. Martin Dyrba}\par} 
\noindent
Research Associate,\\
German Center for Neurodegenerative Diseases(DZNE).\\
Postdoctoral Researcher, \\
Faculty of Computer Science and Electrical Engineering,\\
Universität Rostock,\\
Rostock, Germany 

\vspace{1.0cm}

\noindent
{\large \textbf{Corresponding Professor:}\par} 

\vspace{0.5cm}
\noindent
\rule{10cm}{0.5pt}\\
{\large \textbf{Prof. Dr.-Ing. Thomas Kirste} \par}
\noindent
Chair of the working group MMIS, \\
Institute Director, \\
Visual \& Analytic Computing, \\
Faculty of Computer Science and Electrical Engineering, \\
Universität Rostock, \\
Rostock, Germany 


	\newpage
	\pdfbookmark[0]{Statutory Declaration}{statutory}
	
	\thispagestyle{empty}
	\vspace*{4cm}
	{\parindent 0pt
		\iflanguage{english}{
			\textbf{\Huge{Statutory Declaration}}\vspace{10mm}\\
			I herewith declare that I have composed the present thesis myself and without use of any other than the cited sources and aids. Sentences or parts of sentences quoted literally are marked as such; other references with regard to the statement and scope are indicated by full details of the publications concerned. The thesis in the same or similar form has not been submitted to any examination body and has not been published. This thesis was not yet, even in part, used in another examination or as a course performance.
		}
		
	\\[2cm]}
	Rostock, \deliveryDate
	\\[1.5cm]
	\rule{6cm}{0.5pt}\\
	\parbox[l][1cm][c]{6cm}{\hfill\name\hfill\vbox{}}
	

\newpage
\pdfbookmark[0]{Acknowledgments}{acknowledgments}
\thispagestyle{empty}
\vspace*{3cm}
\begin{center}{\huge\textit{Acknowledgments}\par}\end{center}
\vspace{0.2cm}
First and foremost, I would like to express my gratitude to Dr. rer. hum. Martin Dyrba for his immense support, encouragement and invaluable guidance as my supervisor throughout the whole thesis work. I feel very grateful to have the opportunity to work in the field of deep learning for medical domain. Martin’s idea, enthusiasm and vision have always been motivating for me to overcome the problems that I faced for this thesis work. Without his indispensable guidance this thesis work would not have been completed successfully.

I would also like to show my gratitude to Prof. Dr.-Ing. Thomas Kirste for allowing me to work on this thesis project. His leadership is always encouraging for the whole group of MMIS to have a perfect research environment. 

A special thanks to all the teachers throughout my Master study at Universität Rostock, who contributed to the development of my knowledge that I possess today. 

I would also like to thank Alzheimer’s Disease Neuroimaging Initiative (ADNI) and Australian Imaging Biomarkers and Lifestyle Study of Ageing (AIBL) for providing the data for the study.  

Last but not the least, I am forever grateful to my father, \textit{Md Ataur Rahman}  for his blessings and prayers. Without his unconditional love, support and motivation, it would have not been possible to be what I am today.



\newpage
\pdfbookmark[0]{Abstract}{abstract}
\thispagestyle{empty}
\begin{center}{\large\textsc{{Universität Rostock}}\par}\end{center}
\begin{center}{\huge\textit{{Abstract}}\par}\end{center}
\begin{center}{\textrm{{Faculty of Computer Science and Electrical Engineering,}}\par}{{Universität Rostock}\par}\end{center}

\begin{center}{{Master of Science in Electrical Engineering \\(Specialization in Information Technology)}\par}\end{center}

\begin{center}{\textbf{Learning Shape Features and Abstractions in 3D Convolutional Neural Networks for Detecting Alzheimer’s Disease}\par}\end{center}

\begin{center}{\textit{{by}} {Md Motiur Rahman Sagar}\par}\end{center}
\vspace{0.3cm}
\noindent
Deep Neural Networks – especially Convolutional Neural Network (ConvNet) has become the state-of-the-art for image classification, pattern recognition and various computer vision tasks. ConvNet has a huge potential in medical domain for analyzing medical data to diagnose diseases in an efficient way. Based on extracted features by ConvNet model from MRI data, early diagnosis is very crucial for preventing progress and treating the Alzheimer’s disease. Despite having the ability to deliver great performance, absence of interpretability of the model’s decision can lead to misdiagnosis which can be life threatening. In this thesis, learned shape features and abstractions by 3D ConvNets for detecting Alzheimer’s disease were investigated using various visualization techniques. How changes in network structures, used filters sizes and filters shapes affects the overall performance and learned features of the model were also inspected. LRP relevance map of different models revealed which parts of the brain were more relevant for the classification decision. Comparing the learned filters by Activation Maximization showed how patterns were encoded in different layers of the network. Finally, transfer learning from a convolutional autoencoder was implemented to check whether increasing the number of training samples with patches of input to extract the low-level features improves learned features and the model performance.               

\vspace{0.5cm}
\noindent
\textbf{\textit{Keywords:}} 3D ConvNet; Deep Learning; Neural Network; Alzheimer's; Visualization; MRI; LRP; Activation Maximization; Transfer Learning; Convolutional Autoencoders.


	
	\newpage
	\pdfbookmark[0]{Contents}{contents}
	\tableofcontents
	

	\chapter*{List of Abbreviations}
		\addcontentsline{toc}{chapter}{List of Abbreviations}
		\begin{itemize}
    		\item {\bfseries AI:} {\textbf{A}rtificial \textbf{I}ntelligence }
    		\item {\bfseries ML:} {\textbf{M}achine \textbf{L}earning }
    		\item {\bfseries ANN:} {\textbf{A}rtificial \textbf{N}eural \textbf{N}etwork}
    		\item {\bfseries DNN:} {\textbf{D}eep \textbf{N}eural \textbf{N}etwork}
    		\item {\bfseries ConvNet or CNN:} {\textbf{C}onvolutional \textbf{N}eural \textbf{N}etwork }
    		\item {\bfseries FC:} {\textbf{F}ully \textbf{C}onnected }
    		\item {\bfseries ReLU:} {\textbf{Re}ctified \textbf{L}inear \textbf{U}nit }
    		\item {\bfseries Adam:} {\textbf{Ada}ptive \textbf{M}oments estimation }
    		\item {\bfseries SGD:} {\textbf{S}tochastic \textbf{G}radient \textbf{D}escent }
    		\item {\bfseries ADNI:} {\textbf{A}lzheimer’s \textbf{D}isease \textbf{N}euroimaging \textbf{I}nitiative }
    		\item {\bfseries AIBL:} {\textbf{A}ustralian \textbf{I}maging \textbf{B}iomarkers and \textbf{L}ifestyle study of ageing }
    		\item {\bfseries AD:} {\textbf{A}lzheimer's \textbf{D}isease }
    		\item {\bfseries MCI:} {\textbf{M}ild \textbf{C}ognitive \textbf{I}mpairment}
    		\item {\bfseries NC:} {\textbf{N}ormal \textbf{C}ontrol }
    		\item {\bfseries LRP:} {\textbf{L}ayer-wise \textbf{R}elevance \textbf{P}ropagation}
    		\item {\bfseries AE:} {\textbf{A}uto\textbf{e}ncoder }
    		\item {\bfseries CAE:} {\textbf{C}onvolutional \textbf{A}uto\textbf{e}ncoder }
    		
		\end{itemize}

	\newpage
	\listoffigures

	\newpage
	\listoftables

	\mainmatter


	\chapter{Introduction}\label{chapter:1}
Since the introduction of Neural Network (NN), the idea to mimic the signaling of organic neurons \cite{McCulloch1943}; the intelligence and complexity of Machine Learning algorithms have increased gradually. Although the idea of Neural Network was proposed more than half a century ago due to limited computational resources and availability of datasets, there was not significant improvements in this field for a long time. In late 1990s, advances in various kernel methods \cite{Boser1992} \cite{Muller1997} \cite{Scholkopf1998} and classical Machine Learning techniques such as Support Vector Machines \cite{Smola.1999} \cite{Cortes1995} caught much attention because of its efficient computation algorithm which enabled optimal problem solving capability. In 2000s, the attraction of (Deep) Neural Network can be attributed to increased raw computer power and availability of large datasets. With the integration of available large datasets and capability of processing these extreme amounts of computation via advanced GPUs (Graphics Processing Unit), a spectacular performance achieved in the diverse filed of computer vision \cite{Krizhevsky2012} \cite{Russakovsky2015}. From then, refinement of Neural Network architecture and improvement of model performance became more and more prominent. Soon, high-performance methods of Neural Network made their way out of research phase and grab huge attention in the industry and in media. Artificial Intelligence (AI) and Machine Learning (ML) methods have revolutionized the learning-based techniques in many research areas. By analyzing the data, AI can recognize patterns more efficiently and with enough training they can outperform human expert which enables businesses and industries to have more insight into the data. Some of the application of AI and ML includes speech recognition, Natural Language Processing (NLP), robotics, medical imaging, image classification, autonomous-driving etc. Image classification has become the core in many computer vision applications. Meaningful information extraction from 2D/3D medical imaging (e.g. Magnetic Resonance Image (MRI), X-ray) plays an important role for diagnosis of many diseases in medical domain.

	\section{Alzheimer’s Disease}\label{chapter:1.1}
Alzheimer’s Disease (AD) is a neurodegenerative progressive and irreversible brain disorder which is the most common cause of dementia among the elderly people. Although dementia mainly affect older people, it is not a normal part of ageing. AD results to the death of nerve cells and tissue loss thus reduces the volume of whole brain and overtime affects most of its common function that can even cause death. According to WHO, there are about 50 million cases of dementia around the world and nearly 10 million new cases every year where AD may contribute to 60-70\% of the cases. The number of dementia cases is estimated to be increased to 82 in 2030 and 152 million by 2050 \cite{WHO2019}. There is no conclusive drug or treatment has been revealed to stop or reverse the progression of AD and the cause is still unknown. However, early diagnosis can play a crucial role to slow down the progression to AD. Mild Cognitive Impairment (MCI) is considered as the prodromal of transitional stage of AD and most likely to progress to AD \cite{Markesbery2010}. So, detecting MCI cases who are at high risk of progressing AD can give an essential insight for the treatment. Study showed that, neuropathological changes can be found several years before the diagnosis of AD which caused by the loss of grey matter in the brain \cite{Karas2004}.  

	\begin{figure}[H]
     \centering
     \begin{subfigure}[b]{0.7\textwidth}
         \centering
         \includegraphics[width=\textwidth]{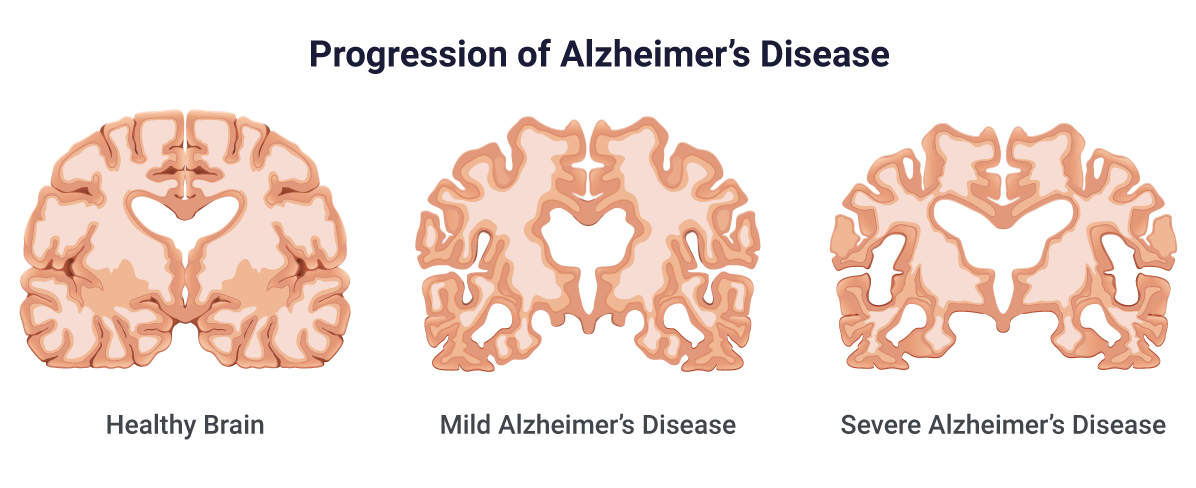}
         \caption{Abstract view of coronal slices from healthy to AD \cite{National2020}.}
         \label{fig:fig1_1a}
     \end{subfigure}
	
     \begin{subfigure}[b]{0.9\textwidth}
         \centering
         \includegraphics[width=\textwidth]{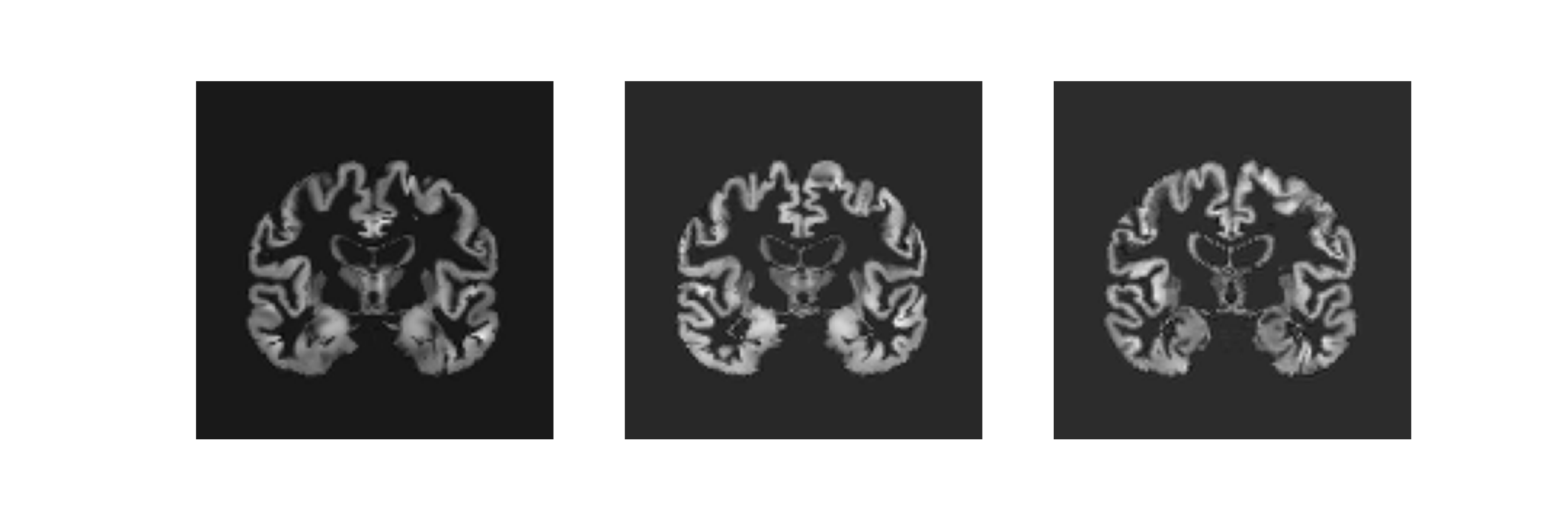}
         \caption{Actual brain scan (coronal view) of healthy control, MCI and AD subject(left to right). }
         \label{fig:fig1_1b}
     \end{subfigure}
     
    \label{fig:fig1_1}
	\caption[Significant tissue loss from healthy to MCI and severe AD.]{Significant tissue loss from healthy to MCI and severe AD. }
	\end{figure}

\noindent	
During the preclinical stage toxic changes take place in the brain but still it is symptom free. Eventually healthy neuron cells of the brain stop functioning properly, lose connection with other neurons and die overtime affecting the essential part of the brain such as hippocampus and entorhinal cortex \cite{National2020} which are responsible for forming memories. Besides that, many other critical brain changes are also considered responsible for AD. Figure \ref{fig:fig1_1a} shows how by the brain tissue has shrunk from healthy to MCI and severe AD. In many research \cite{Liu2015} \cite{Suk2014} \cite{Zhang2011} \cite{Cuingnet2011} neuroimaging biomarkers has been used to detect different stages of AD or to predict MCI to AD progression and MRI is the most used imaging technique because of its high resolution, non-invasive and cost effectiveness. Figure \ref{fig:fig1_1b} displays actual MRI scan of a healthy, MCI and AD subject.

	\section{Motivation}\label{chapter:1.2}
Over the past decade Machine Learning (ML) has greatly benefited many applications in the field of computer vision especially in image classification. Traditional ML methods such as Support Vector Machine (SVM) and Random Forest (RF) depends on various hand-crafted feature extraction like Local Surface Patches \cite{Chen2007}, Intrinsic Shape \cite{Yu2009}, Heat Kernel Signature \cite{Sun2009}, Spin Images \cite{Johnson1999}, etc. Despite having the ability to deliver great performance, these methods have been criticized because of their inability to extract adaptive features and always requires human intervention. Unlike traditional ML methods, Deep Learning methods which is based on Deep Neural Networks learns the feature as a part of training process. The Convolutional Neural Network (CNN or ConvNet) is a part of Deep Learning that has proven and become the state-of-the-art for detecting pattern and shapes in grid-like 2D/3D data such as RGB images and MR images. Starting with the great success on ImageNet classification problem of AlexNet \cite{Krizhevsky2012}, ConvNet has soon found its way into many diverse research fields. Application of CNN in medical image analysis domain also achieved successful result in 2D images such as X-ray \cite{Bar2015} and retinal scans and extended for 3D data like MRI as well. Although Deep Learning or ConvNet extract the features for a particular classification problem from the given training data by itself, it does not necessarily mean that the model has to match the desired meaningful features, thus these systems are considered as \textit{black box}. As AI applications impacting our everyday life by supporting in many important tasks, it is crucial to make sure that the system is trustworthy. Maybe in some situation simple binary \textit{yes/no} decision or prediction is enough but when it comes to life, health and death of humans and animals such as in medical domain or in autonomous driving, transparency in the decision making is inevitable. Absence of interpretability of the model’s decision can lead to misdiagnosis which can be life threatening in medical domain and can jeopardize the protection of passengers and pedestrians in case of autonomous driving. The reasons like \textit{what} and \textit{how} an automation system makes the decisions are the critical significance in research area. Besides, Deep Learning based models are known to outperform humans in same task, for instance playing game \cite{Silver2016} \cite{Firoiu2017} \cite{Moravcik2017} \cite{Mnih2015} or in image classification \cite{He2016} \cite{Szegedy2016}, having the information about the model’s decision reasoning could open up a whole new way of thinking for us. So, to have an automated system that extract meaningful features for a specific application and also interpretable and transparent is an open research option.

	\section{Objective}\label{chapter:1.3}
Deep Neural Networks - especially Deep ConvNets have become a powerful tool for detecting pattern and shape features in image data. Remarkable successes have been achieved over the past few years with 2D image/data like classification, pattern recognition and various computer vision tasks. Outstanding performance starting with AlexNet \cite{Krizhevsky2012} followed by GoogLeNet \cite{Szegedy2015}, VGG16 \cite{Simonyan2015}, ResNet \cite{He2016}, etc. achieved because of the fact that (a) enormous amount of research effort designing core network architectures (b) availability of large scale annotated datasets such as ImageNet \cite{JiaDeng2009}, MNIST \cite{LeCun1998}, CIFAR \cite{Krizhevsky}, etc. Compared to 2D, available researches are limited in the field of 3D data as there has been no standard representation method developed yet, lack of labeled large datasets and required computational power is also very high to process the 3D data. Most popular 3D dataset called ModelNet \cite{Wu2015} has been used in many researches such as 3D ShapeNet \cite{Wu2015}, VoxNet \cite{Maturana2015}, DeepPano \cite{Shi2015}, Multi-view CNN \cite{Su2015}, for 3D shape detection/classification. For each of the studies 3D data has been preprocessed with different strategies. Several methods have also been developed and researched to open up the black box nature and learning abstraction (e.g. learned filters, feature maps, heatmap, activation maximization, layer-wise relevance propagation) of the 2D classifier models. 

By Extracting meaningful features/patterns directly from raw or minimally processed 3D MRI scans (e.g. hippocampus shape, ventricle size, cortical thickness, brain shape-volume), ConvNets can offer a great potential for improving automatic diagnosis of early stage and predicting progression of the Alzheimer’s Disease. Recent research papers \cite{Liu2015} \cite{Suk2014} \cite{Zhang2011} \cite{Cuingnet2011} \cite{Marzban2019} \cite{Bohle2019} \cite{Rieke2018} \cite{Gupta2013} also report promising results in the domain of disease detection using brain MRI data. Despite the high accuracy obtained from CNN models for MRI data so far, almost no papers provided information on the features or image regions driving this accuracy. Currently, there is a great demand for a systematic and theory-driven evaluation on how different 3D shapes can be encoded and recognized by 3D CNNs with respect to the required layer structure and input image properties (e.g. size, contrast).

\vspace{0.5cm}
\noindent
Therefore, this thesis has two main goals:	
	\begin{itemize}
		\item Systemically evaluate \textit{state-of-the-art} 3D CNN models in various fields of research (e.g. industry, robotics, cars, medical domain).
	\end{itemize}
	
	\begin{itemize}
		\item To implement various 3D CNN approaches and evaluate shape recognition using simulated and brain MRI data.
	\end{itemize}

	\section{Related Work}\label{chapter:1.4}
Although CNNs have achieved excellent classification accuracy, one of the major shortcomings of these networks is the lack of transparency in decision making, particularly the limited understanding of the internal features extracted by each convolutional layer. During training process convolutional filters (also known as kernels) are optimized in each layer and convolved output is recorded as feature maps. Initial Conv filters are expected to learn lower level features (e.g. edges, colors) and deeper layers extract higher label features (e.g. shapes, corners, faces, eyes, noses, objects) \cite{Krizhevsky2012} \cite{Zeiler2014} and eventually converge to the final classification. As the filters are formatted as a simple matrices and optimization of these filters during training process is unpredictable, it is impossible to explain these filters directly. Thus, the inadequate network transparency restricts the further optimization possibility as well as evaluation robustness. Furthermore, transferability and adaptability to other applications of the network is also hindered \cite{Pan2010} \cite{Shin2016}.         
	
	 \begin{figure}[H]
		\centering
		\includegraphics[width=\textwidth]{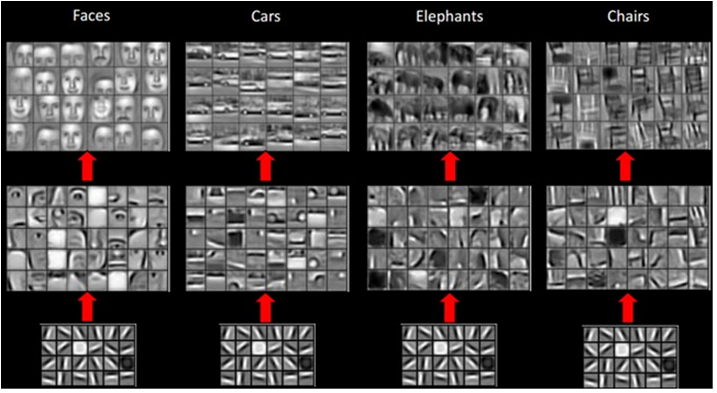}
		\caption[CNN firsts identify low level features (bottom) and then in deeper layers more complex features (middle and top) are recognized by combining the low-level features.]{CNN firsts identify low level features (bottom) and then in deeper layers more complex features (middle and top) are recognized by combining the low-level features. source: [\url{https://i.stack.imgur.com/Hl2H6.png}]}
		\label{fig:fig1_2}
	\end{figure} 

\noindent
Much like human visual cortex which has multiple vision neuron areas and each of them respond to different features such as shapes, edges and colors \cite{Kay2008}, networks interpretability can also be improved by visualizing learned features into recognizable pattern in the image. Recently several effective visualization techniques have been proposed and many of them are also adopted. Layer-wise Relevance Propagation (LRP) \cite{Bach2015} is one of the most plausible visualization options, which conserves the relevance like activation strength of an output neuron for a certain class per layer. Another proposed algorithm called Activation Maximization and the idea behind is to generate an input image that maximizes the CNN filter activation of a single neuron \cite{Erhan2009}. Further improvement of the idea was implemented by applying different regularization methods \cite{Nguyen2016}. The DeconvNet based visualization technique which can highlight the pattern of an input image that activates a specific neuron thus creates a direct link between the neuron and input data \cite{Zeiler2014}. Several other visualization methods such as Saliency Map \cite{Baehrens2010} \cite{Simonyan2014}, SmoothGrad \cite{Smilkov2017}, IntegratedGradients \cite{Sundararajan2017}, GuidedBackprop \cite{Springenberg2015}, PatternNet \cite{Kindermans2018}, DeepTaylor \cite{Montavon2017} etc. were also proposed. Most of the visualization methods described above are investigated under natural 2D images but visualizations of 3D image like MRI are still in very early stage. Due to the complex representation of 3D data it is quite difficult to interpret/visualize learned feature of a 3D CNN. Nevertheless, there are some resent studies \cite{Bohle2019} \cite{Rieke2018} \cite{Zintgraf2019} showing CNN visualization of 3D MRI data. Research regarding visualization of the learned feature for disease detection for 3D data can play a significant role for the adaptability of the network in the medical domain.   

	\section{Thesis Structure}\label{chapter:1.5}
Over the course of the thesis work learning abstraction of 3D ConvNet is described with various visualization methods. So far, the motivation, objective and related work regarding this thesis has been described with little background about Alzheimer’s disease. The rest of the chapters of thesis report are organized as follows:

\vspace{5mm}
\textbf{Chapter \ref{chapter:2} (Fundamentals)} describes the background theory and concepts that are indispensable for the understanding of this work. Starting with the ANN of a single neuron the chapter builds up the core theory of CNN with its training approaches and evaluation technique as well. This chapter also explain the concept behind the transfer learning and its application.

\textbf{Chapter \ref{chapter:3} (Methods and Concepts)} explains methods that were used for extracting the learned features by the network. Filters and feature map visualization, Layer-wise Relevance Propagation and Activation maximization were the choice of methods for interpretability of the network. This chapter describes the theory behind these visualization techniques. 
 
\textbf{Chapter \ref{chapter:4} (Implementation)} represents all the implementation steps for various experiments. First data acquisition and pre-processing steps are described then methods for implementing 3D ConvNet with their structure are explained. This chapter also introduced the concept of Convolutional Autoencoder and using this for transfer learning. Finally, training environment and tools used throughout the thesis work are mentioned.

\textbf{Chapter \ref{chapter:5} (Results and Discussion)} shows standout results from the experiments that were conducted and discusses the results with respect to the goal of the thesis. First the training results are evaluated depending on different experiments then visualization techniques for interpreting the networks learned and illustrated and discussed. The results from different method are also compared. 

Finally, \textbf{Chapter \ref{chapter:6 } (Summary and Outlook)} concludes this thesis with overall summary and results that were significant outcomes of this work. Here, some of the limitation that were faced throughout the thesis are mentioned with room for improvements and future scope of work as well.

	\chapter{Fundamentals}\label{chapter:2}	
This chapter provides a brief overview and background of various concepts that are essential for the understanding of this thesis work. Deep learning is a sub-category of Machine Learning. Among various approaches in Deep learning Convolutional Neural Network  is the most used image classification technique. The fundamental idea behind the Convolutional Neural Network will be described which is the core of this study.

	\section{Artificial Neural Network}\label{chapter:2.1}
The algorithm of ANN is inspired by the structure of human brain that contains billions of neurons. The neurons inside the brain are connected with each other by axon-synapses-dendrite connection to transmit the information in the form of electrical impulses which cause the rise of potential of respective neuron. If the potential is strong enough then the neuron fires an electrical impulse to all connected neurons. Figure \ref{fig:fig2_1a} below shows a single biological neuron.The mathematical model is designed on the basis of biological neuron which can produce an output signal according to the input signal(s) that it receives. Figure \ref{fig:fig2_1b}  illustrates a comparative mathematical model of the biological neuron that often known as artificial neuron or perceptron. The inputs of the constructed model are corresponding to the dendrites, transfer function, net input and activation function refer to the cell body and activation or output refers to the axon and synaptic terminal. The strength of the output impulse is controlled by the synaptic weights. The output of the \(jth\) neuron can be formulated as,

\begin{equation}\label{eq:eq2_1}
  o_j = \phi( \theta_j + \sum\limits_{i=1}^n w_{ij}x_i)
\end{equation}

where \( \phi(\bullet) \) is the activation function, \(\theta_j\) is the bias for \textit{jth} neuron, \textit{x} is the input, \textit{w} is the corresponding weight and \textit{n} is the number of inputs. 

	\begin{figure}[H]
     \centering \offinterlineskip
     \begin{subfigure}[b]{0.8\textwidth}
         \centering
         \includegraphics[width=\textwidth]{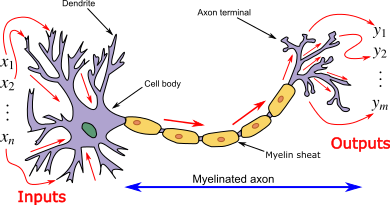}
         \caption{}
         \small source: [\url{http://wikipedia.org/wiki/File:Neuron3.png}]
         \label{fig:fig2_1a}\vspace{0.5em}
     \end{subfigure}\vspace{1em}
			
     \begin{subfigure}[b]{0.9\textwidth}
         \centering
         \includegraphics[width=\textwidth]{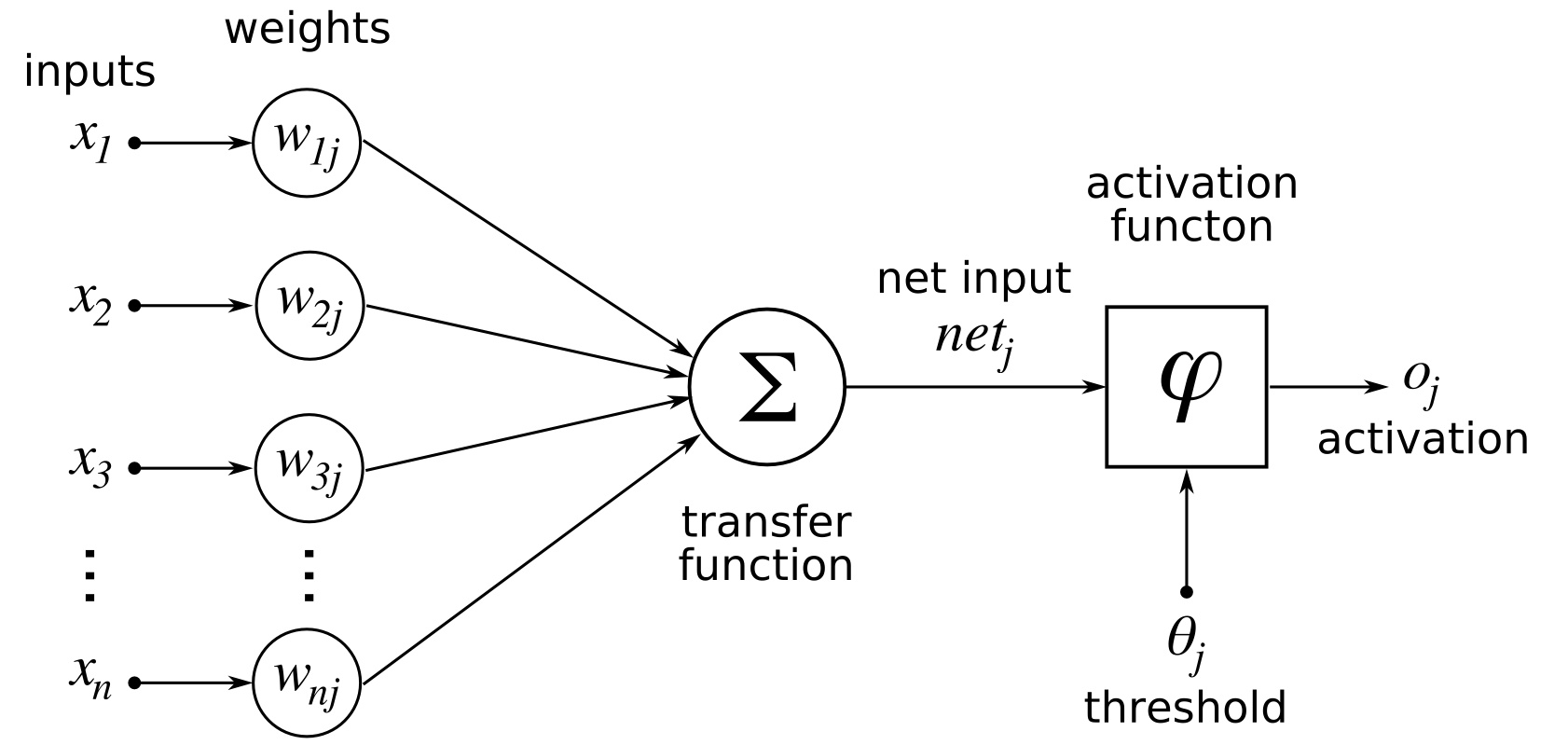}
         \caption{}
         \small source: [\url{https://commons.wikimedia.org/wiki File:ArtificialNeuronModel_english.png}]
         \label{fig:fig2_1b}
     \end{subfigure}\vspace{1em}
     
    \label{fig:fig2_1}
	\caption[(a) A biological neuron. (b) Mathematical model of a neuron]{(a) A biological neuron. (b) Mathematical model of a neuron.}
	\end{figure}

\begin{figure}[H]
		\centering
		\includegraphics[width=0.8\textwidth]{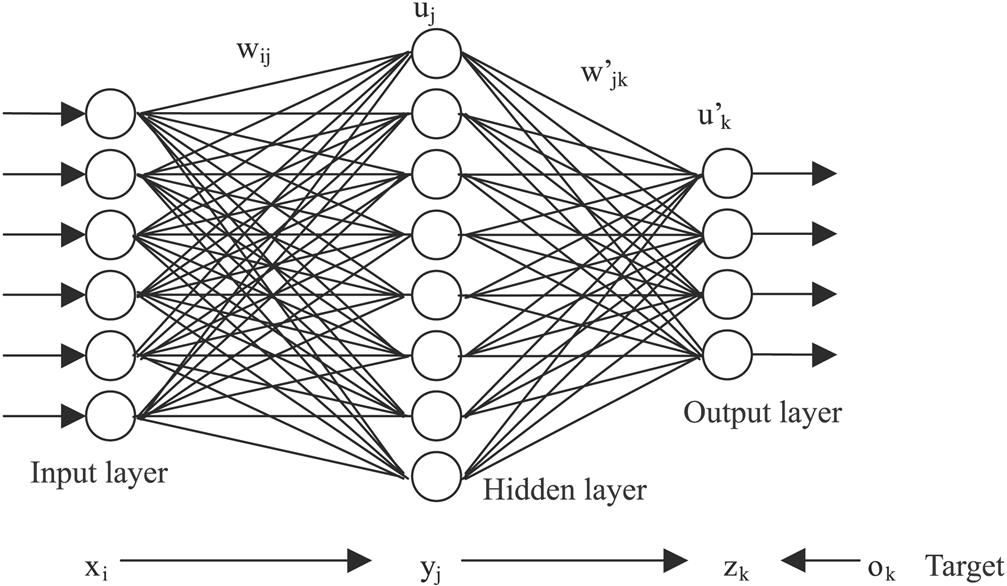}
		\caption[Architecture of an ANN with one fully-connected (FC) hidden layer. Each neuron of a FC layer is connected to each neuron of its previous and next layer.]{Architecture of an ANN with one fully-connected (FC) hidden layer. Each neuron of a FC layer is connected to each neuron of its previous and next layer. Source: [\url{http://www.extremetech.com/wp-content/uploads/2015/07/NeuralNetwork.png}}
		\label{fig:fig2_2}
	\end{figure}

\noindent
This basic mathematical neuron is connected with multiple neurons to create a network hence called Neural Network or Artificial Neural Network. An ANN can be as simple as having a single perceptron but often it consists of an Input layer with multiple inputs, a hidden layer and an output layer. Figure \ref{fig:fig2_2} shows a basic structure of an ANN.

\noindent
Each node of the network corresponds to a neuron. A deeper ANN consists of multiple hidden layers with different number of nodes thus it is called Deep Neural Network (DNN). The inputs of a node come from previous nodes (except input layer) and the output becomes the inputs for next nodes (except output layer). During training process weights are adjusted through backpropagation with a loss function (described in section \ref{chapter:2.3.1} and \ref{chapter:2.3.2}) to make the final decision of the network.

	\section{Convolutional Neural Network}\label{chapter:2.2}
Convolutional Neural Network (CNN or ConvNet) is a class of Deep Neural Network that is well known for various computer vision applications. CNNs are very similar to the classical Neural Networks as they are made up of neurons with learnable weights and biases, but neuron of a hidden layer is connected in an efficient way to extract pattern in images. In regular Neural Networks each neuron of a hidden layer is fully connected with all the neurons in the previous layer (Fully-Connected or Dense layer) and the last fully-connected layer called output layer which represents classification scores. Clearly this idea does not scale for image data. For example, an RGB image with resolution of \( (224\times224\times3) \) would make 150,528 weights for a single fully-connected neuron in first hidden layer. As each hidden layer has a set of fully-connected neurons and in a single layer connection are independent so the parameters would add up (first hidden layer with 64 neuron results over \textit{9 million} parameters) and become prone to overfitting and it requires huge memory for the computation. Moreover, as the pixel of the image is flattened all spatial information would be lost. Unlike classical NNs, in ConvNets each neuron of a hidden layer would be connect with a small region of the input layer keeping the spatial information of the image and reducing the number of parameters significantly compared to fully-connected layers. 

	 \begin{figure}[H]
		\centering
		\includegraphics[width=0.9\textwidth]{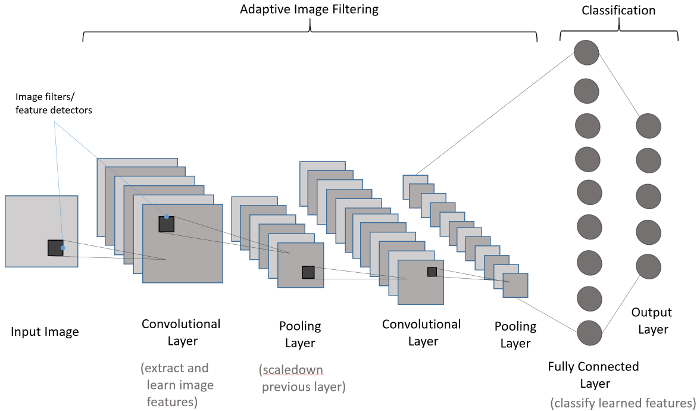}
		\caption[A basic ConvNet with two convolutional layers followed by pooling layer for feature extraction and two FC layers for classification.]{A basic ConvNet with two convolutional layers followed by pooling layer for feature extraction and two FC layers for classification. Source: [\url{http://www.deepnetts.com/blog/wp-content/uploads/2019/02/ConvNet.png}]}
		\label{fig:fig2_3}
	\end{figure}

\noindent
ConvNet architecture is comprised of several hidden layers that include Convolutional layers, Pooling layers, Activation functions and Fully-Connected (FC) layers (same as regular NN) that produces class score. Figure \ref{fig:fig2_3} shows a basic architecture of a ConvNet. The following sections briefly discuss the basic building blocks of ConvNet.

	\subsection{Convolutional Layer }\label{chapter:2.2.1 }
\begin{figure}[H]
		\centering
		\includegraphics[width=0.8\textwidth]{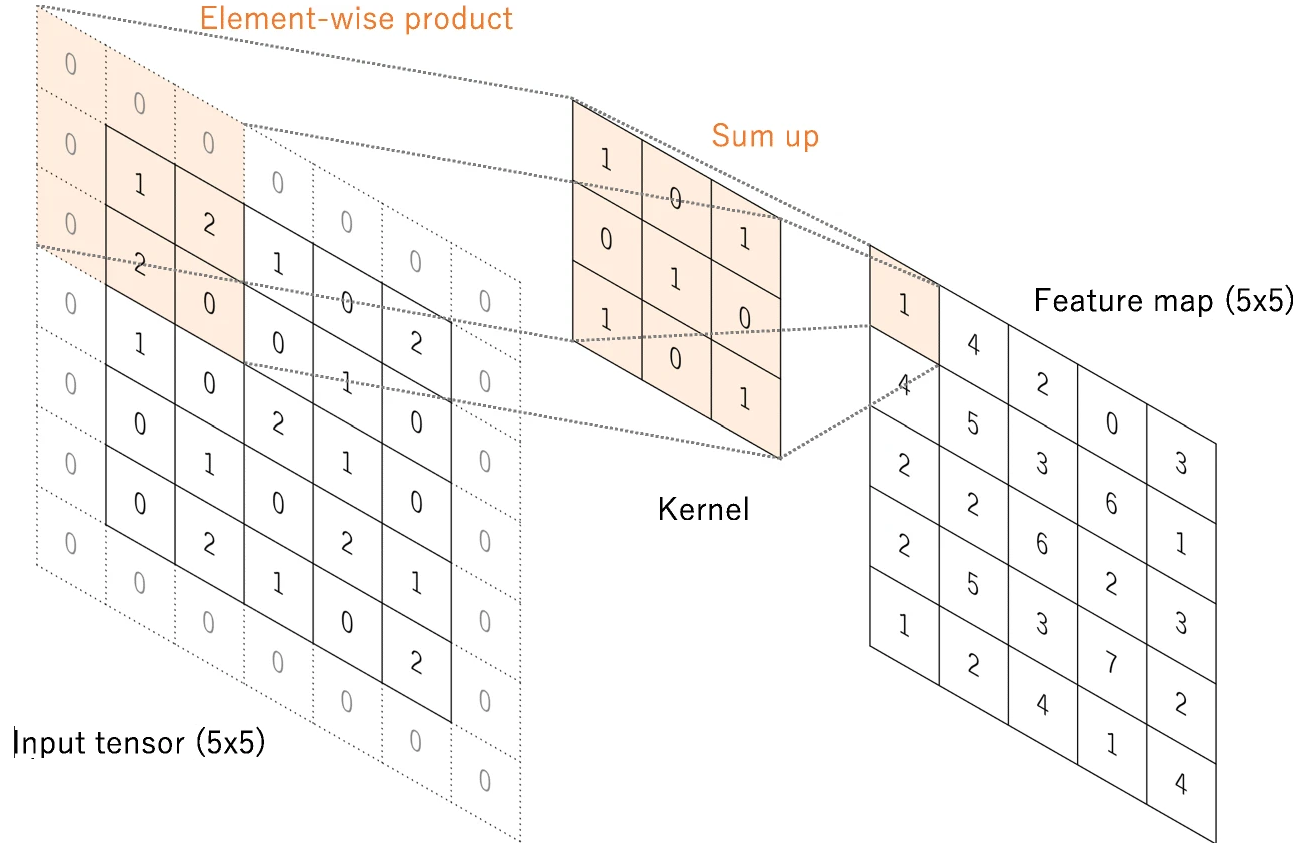}
		\caption[Convolution operation by a \((3\times3)\) filter with stride=1, padding=1, results same size feature map of input image \((5\times5)\).]{Convolution operation by a \((3\times3)\) filter with stride=1, padding=1, results same size feature map of input image \((5\times5)\).  Feature map size = \( ((\frac{5+2\times1-3}{1}+1)\times(\frac{5+2\times1-3}{1}+1))\) = \((5\times5)\) Highlighted feature map result, 1 = \((0\times1)+(0\times0)+(0\times1)+(0\times0)+(1\times1)+(2\times0)+(0\times1)+(2\times0)+(0\times1)\). Figure source \cite{Yamashita2018}.}
		\label{fig:fig2_4}
	\end{figure}

Convolutional layer or Conv layer usually is the first and core building block of ConvNets that extract feature from an image and outputs feature maps by convolution operation. Conv layer consists a set of filters/kernels/weights and each filter convolves over the entire input image and computes dot products between the filter and input of the image at that position. Figure \ref{fig:fig2_4} shows an example of convolution operation. Typically, the kernel sizes are of \( (3\times 3)\),\((5\times 5)\),\((7\times7)\) and number of kernels can be any number which refers to the depth of feature map. To preserve the input image size padding (usually 0) is used which known as \textit{same} mode. Number of padding to be used depends on the kernel size. If no padding is used the output reduces the resolution known as \textit{valid} mode. 

\noindent
The distance between consecutive kernel position is known as stride. The kernel slides one pixel after each operation by default (stride = 1), but this can also be controlled by changing the stride. Stride 2 means kernel would shift 2 pixels after each operation. The convolutional output size of an \((n\times n) \) image with filter size \((f\times f)\), padding \(p\) and stride \(s\) can be calculated as \( ((\frac{n+2p-f}{s}+1)\times(\frac{n+2p-f}{s}+1)) \). In case of color images, the input size becomes \((n\times n\times n_c)\) and filter size \((f\times f\times n_c)\), where \(n_c\) is the number of channels. The process described above known as 2D convolution. The depth of the feature map is equal to the number of filters used. In 3D convolution feature map produced by a single filter is also 3D. The output after convolution operation are passed through a non-linear activation function (more on section \ref{chapter:2.2.3 }) which gives a nonlinear representation to learn complex functions that maps the input data to the classification output. The kernels/filters/weights associated with a ConvNet are adjusted while training to extract features from the input image.  

	\subsection{Pooling Layer }\label{chapter:2.2.2 }
\begin{figure}[H]
		\centering
		\includegraphics[width=0.8\textwidth]{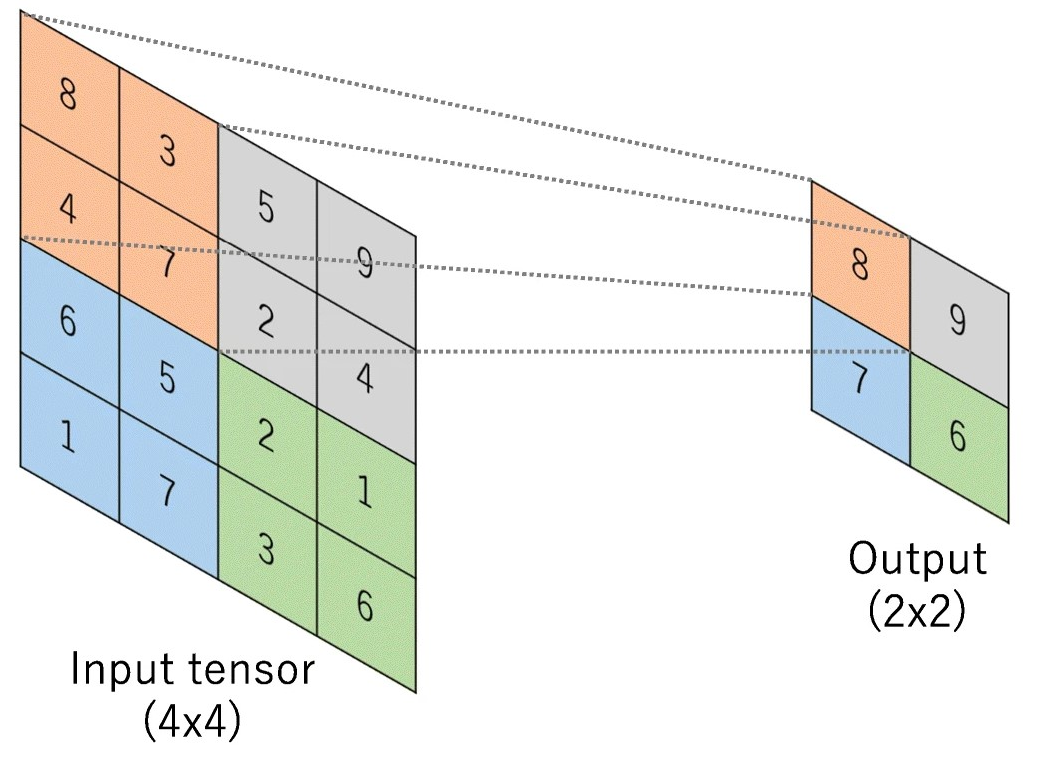}
		\caption[MaxPooling operation on a \((4\times4)\) image with \((2\times2)\) window size where, stride=2 and padding=0.]{MaxPooling operation on a \((4\times4)\) image with \((2\times2)\) window size where, stride=2 and padding=0 which results \( ((\frac{4+2\times0-2}{2}+1)\times(\frac{4+2\times0-2}{2}+1))\) = \((2\times2)\) output. Figure source \cite{Yamashita2018}.}
		\label{fig:fig2_5}
	\end{figure}

After a Conv layer a pooling layer is applied that provides typical downsampling to reduce the dimensionality of the feature maps and while keeping the spatial information. Because of the downsampling the number learnable parameters also decrease for the subsequent Conv layers and introduces a translation invariance to small shifts and distortions. As this is only a mathematical operational layer there is no learnable parameters involved, although filter size, padding and strides works in a similar way that of a Conv layer. Max pooling is the most used pooling operation which extract the maximum value out of each patch from the input feature map. Figure \ref{fig:fig2_5} shows a max pooling operation with filter size \( (2\times2) \) with stride 2 so that patches does not overlap. Other types of pooling include average pooling, min pooling, etc. which has their specific application.

	\subsection{Activation Function }\label{chapter:2.2.3 }
As mentioned before, the output of each convolutional operation passes through an activation function. This is true for fully-connected layers as well. Activation function maps a linear input to a non-linear output which helps the network to learn more complex data classification. Choosing the right activation function is very crucial for a network to learn during backpropagation which tries to minimize the loss function by adjusting learnable parameters using gradient descent (details in section \ref{chapter:2.3}). There are several activation functions that can be used such as Sigmoid, tanh, ReLU, Leaky ReLU, etc. 
	
	\begin{equation}\label{eq:eq2_2}
  		sigmoid,\ \sigma(x) = \frac{1}{1+e^{-x}}
	\end{equation}  

	\begin{equation}\label{eq:eq2_3}
  		tanh,\ \sigma(x) = tanh(x)
	\end{equation}
	
	\begin{equation}\label{eq:eq2_4}
  		ReLU,\ \sigma(x) = max(0,x)
	\end{equation}
	
	\begin{equation}\label{eq:eq2_5}
  		softmax,\ \sigma(x) = \frac{e^{x_i}}{\sum_j e^{x_i}}
	\end{equation}
	
	\vspace{0.5cm}
	
	\begin{figure}[H]
		\centering
		\includegraphics[width=\textwidth]{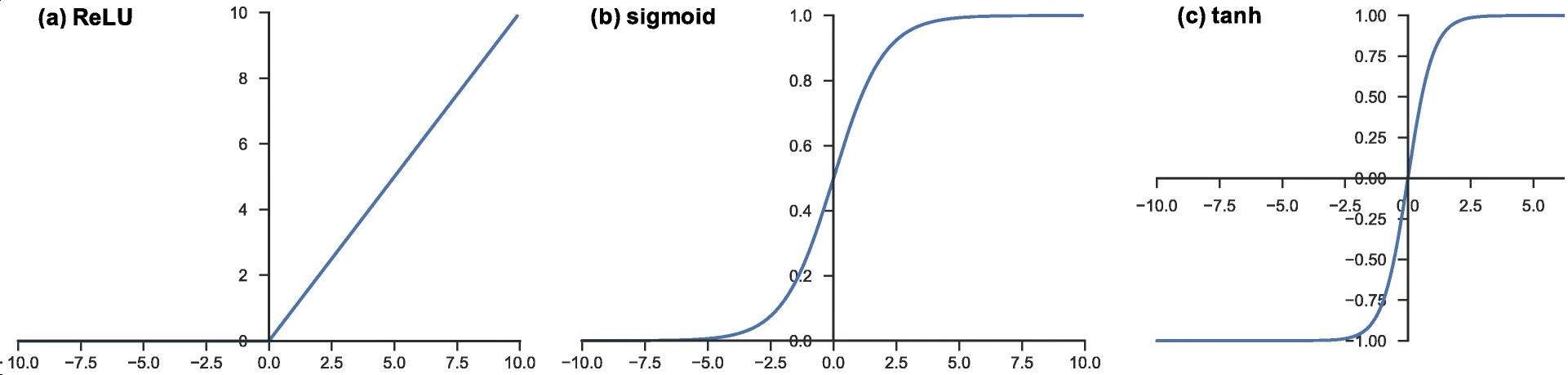}
		\caption[Commonly applied activation functions to neural networks. (a) ReLU, (b) Sigmoid and (c) tanh.]{Commonly applied activation functions to neural networks. (a) ReLU, (b) Sigmoid and (c) tanh \cite{Yamashita2018}.}
		\label{fig:fig2_6}
	\end{figure}

\noindent
Sigmoid activation function (eq.\ref{eq:eq2_2}) maps all input between 0 and 1 that means data is squeezed and gradient for very large positive or negative value becomes close to 0. Once the gradient becomes zero the neuron becomes useless as there is no learning. This problem is known as \textit{vanishing gradient problem}. While tanh activation function (eq.\ref{eq:eq2_3}), increase the output range between -1 to 1, it does not solve the problem. To solve the issue activation function named Rectified Linear Unit or ReLU (eq.\ref{eq:eq2_4}) is used which does not squeeze the input. ReLU passes all positive value of the input as it is and all negative value is mapped to 0. ReLU is proven to be the most used and best activation function in the inner layers of the network  \cite{Krizhevsky2012} \cite{Russakovsky2015} \cite{He2016} \cite{Simonyan2015} \cite{YannLeCunYoshuaBengio2015} \cite{Rama2018} \cite{Nair2010}. Sometimes during training, bias of some neurons can become very negative and since ReLU blocks negative value these neurons are off even worse if the training input is not strong enough to overcome this negative bias then the neuron becomes dead. This is known as dying ReLU problem which can be solve using Leaky ReLU, similar to ReLU but negative numbers are not completely blocked instead mapped to a small negative value. However, this is very rare and for most of the cases ReLU would work just fine.    
The last layer of neural network usually consists an activation function called Softmax which is similar to Sigmoid function but gives a probabilistic output for each class of the classification. Each element in the output depends on the entire set of elements of the input. Softmax activation function (eq.\ref{eq:eq2_5}) normalizes the outputs from the last fully-connected layer to target class between 0 and 1 and summation of all values equal to 1.

	\subsection{Normalization Layer }\label{chapter:2.2.4 }
Normalization layers are proposed to use in ConvNets to force the output of each layer into a certain range which empirically shows faster convergence. However, this is only observable in a deep network and with size of the normalization is large enough. There are many normalization techniques introduced such as batch normalization, group normalization, weight normalization, etc. to normalize the values in a different way. As the contribution has been shown to be very minimal, these normalization layers have fallen out of interest in practice.

	\subsection{Fully-Connected Layer }\label{chapter:2.2.5 }	
The last layer of ConvNets is a fully-connected layer where all the neurons of the layer have connection from the previous activations. This is same as the hidden layers of Neural Networks except it has different activation function (Softmax) than the hidden layers which gives probabilistic out of all the output neurons. Before connecting to the fully-connected layers the output of last Conv layer is flattened to make it a 1D array. Sometimes ConvNets also have couple of fully-connected layers (with ReLU) similar to hidden layers of Neural Networks before the last fully-connected classification layer. See section \ref{chapter:2.1} for more details.

	\section{Training of CNNs}\label{chapter:2.3}
Training a CNN means adjusting the kernels/filters of Conv layers and weights and biases of fully-connected layers to minimize the difference between predicted output and ground truth labels on the training dataset. The difference between ground truth and predicted output is calculated using a loss function. First, the network is initialized with random kernels, weights and biases of a certain range and during forward pass model performance is calculated for particular kernels and weights by a loss function on a training dataset. The learnable parameters are then updated to minimize the loss calculated during forward pass by an optimization algorithm and the process is known as backpropagation.  

	\subsection{Loss Function }\label{chapter:2.3.1}	
While forward propagation the difference between the ground truth or the labels of the training dataset and the predicted output is calculated using a loss function which is also known as cost function. Main goal of training a network is to minimize these costs by updating the learnable parameters in each iteration of training (epoch). Loss function needs to be determined according to the specific classification task such as Cross Entropy is used for multiclass classification, Mean Squared Error (MSE) is used for regression to continuous values. The loss function is one of the hyperparameters that needs to be determined while optimizing the model for a given task.      

	\subsection{Backpropagation }\label{chapter:2.3.2}
The process of updating the learnable parameters (e.g., kernels, weights) to minimize the loss calculated in forward propagation hence improving the model performance is known as backpropagation. Model performance is calculated as a function of loss function such as MSE where loss for each input of the training set is obtained and squared then average of these squared errors is taken. If \(P\) is the performance of the network calculated with MSE then,

	\begin{equation}\label{eq:eq2_6}
  		P = \frac{(e^2_1+e^2_2+...+e^2_n)}{n}
	\end{equation}

\noindent
where, \(n\) is the number of training input and \(e\) is the error calculated for each input. Model performance is optimized using an algorithm called Gradient Descent which tries to minimize the loss function. The gradient of the loss function determines in which direction the function has the sharpest increase rate so each learnable parameter is updated in the negative direction of the gradient to minimize the loss hence the name Gradient Descent. Each of the learnable weights needs to be adjusted because loss is the function of all the weights in the network so we need to apply partial derivative. The step size by which the network improve is also determined by a hyperparameter called learning rate which is a very small number (between 0.01 to 0.0001). The weight updates are calculated as,

	\begin{equation}\label{eq:eq2_7}
  		\Delta \vec{w} = r\times (\frac{\partial P}{\partial w_1},\frac{\partial P}{\partial w_2},\frac{\partial P}{\partial w_3},...,\frac{\partial P}{\partial w_q})
	\end{equation}

\noindent
where, \( \Delta \vec{w} \) is the vector of all the updated weights, \textit{P} is the performance of the model, \textit{r} is the learning rate and \textit{q} is the number of weights in the network. Usually training dataset consist a huge number of inputs which can be challenging to calculate gradients for all the inputs at once due to the memory limitation. So gradient of the loss function is calculated using a subset of the input at once which is known as mini-batch. This method is called mini-batch Gradient Descent or often referred as Stochastic Gradient Descent (SGD). Learning rate along with mini-batch size are very important hyperparameters for optimizing a neural network. Many improvements of SGD have been proposed and widely used such as RMSprop, AdaGrad, AdaDelta, AdaMax, NAdam, Adam, \cite{Ruder2016} etc. among them Adam optimizer is most popular because of its robust for a wide range of application \cite{Kingma2015}.   

	\subsection{Overfitting }\label{chapter:2.3.3}
Overfitting is one of the most challenging factors in the Machine Learning. Sometimes when a model is trained it tries to memorize irrelevant regularities specific to the training set rather than learning. Due to overfitting the model fails to generalize on new unseen data and therefore performs very poorly. So, it is a good idea to evaluate a model’s performance with a test set that is not a part of training dataset. Most common way to check overfitting is to monitor loss and accuracy on the training and validation sets. If model performance on training and validation differ significantly then model is considered most likely overfitted. Several methods can be applied to tackle or minimize the overfitting problem. The best practice is to train on large dataset which helps the model to generalize better. Sometimes obtaining large dataset is not possible for example in medical domain. The other solutions such as batch normalization, dropout, data augmentation can help the model not to be overfitted. Model complexity can also be a root cause for overfitting as complex model has larger number of parameters to learn and without enough training sample model can be overfitted easily. So, it is always a good practice to reduce the model architecture complexity with smaller training data. Regularization technique such as dropout where randomly selected neurons are set to off state so that model is not sensitive to a particular activation. To increase the training data by random transformation like rotating, shifting, cropping, flipping, etc. is the process called data augmentation. Data augmentation helps the model not to train on same data during training epochs. Batch Normalization normalizes the input values of the following layer which helps better gradient flow, allow higher learning rate to mitigate the overfitting problem. Nonetheless overfitting can still be present in the model because of information leakage during hyperparameter optimization. Therefore, the best way to report a model performance is to evaluate on test data set that are separate from training and validation dataset.

	\subsection{Parameters and Hyperparameters }\label{chapter:2.3.4}
Training a ConvNet or Neural Network means adjusting the learnable parameters such kernels, weights and biases of the network. These are called the ultimate parameters of the network. As already discussed in the previous sections that there are some parameters that controls the ultimate parameters of the network which are called hyperparameters. Tuning hyperparameters is very crucial for training a network. There are many hyperparameters in a network such as learning rate, training iterations (epochs), activation functions, number of hidden layers, number of neurons in a layer, number of kernels, size of the kernel, etc. Task of tuning hyperparameter can be very daunting because there can be thousands of combinations depending on the network architecture and application. Hyperparameters tuning can take weeks specially for the network that is deep and has a huge training dataset. In cases where the training data is evolving optimized hyperparameter does not guaranty the best performance for newer data so optimizing hyperparameter is not a onetime process. It is also important to perform cross-validation during optimization which split the dataset into \textit{k-fold (k=5, 10, 20…)} and for each fold different validation set is used to evaluate the model performance.

	\subsection{Data Preparation and Evaluation }\label{chapter:2.3.5}
Data preparation is a vital part of training of a model. Generally, ConvNets are trained on large dataset, where the entire dataset is split into three sets: a training set, a validation and a test set. After training a model its performance is evaluated on test set. However, for small data leaving out a portion of the dataset for validation can cause loss of important feature learning from the validation set. The gold standard for model evaluation in practice is \textit{k-fold} cross validation. It provides a robust estimation on model performance on unseen data. \textit{K-fold} splits the training dataset into \textit{k} subsets and each fold appears in the training set \textit{(k-1)} times, thus enables the model to learn from entire dataset.    

	\begin{figure}[H]
		\centering
		\includegraphics[width=0.8\textwidth]{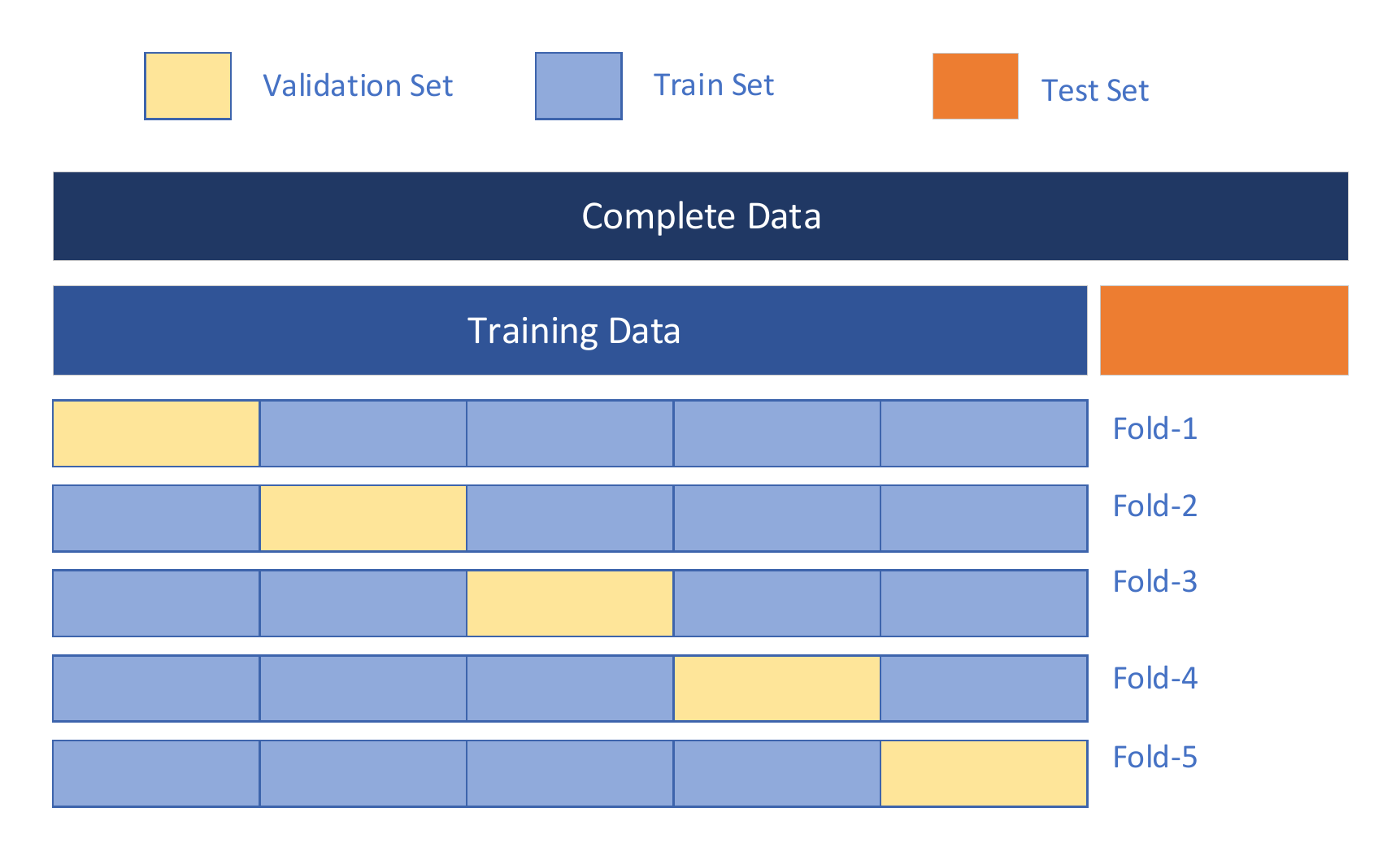}
		\caption[\textit{5-Fold} cross validation.]{\textit{5-Fold} cross validation.}
		\label{fig:fig2_7}
	\end{figure}

\noindent
Figure \ref{fig:fig2_7} represents a \textit{5-fold} cross validation on the training data. For each iteration the model is trained on a different fold with a validation set different from other folds. The performance is captured for each model then by averaging and standard deviation a robust performance of the model is estimated. To make sure the ratio of the observations in each class remain same in each fold another technique called Stratified \textit{k-fold}  is used.

	\section{Learning Approaches  }\label{chapter:2.4}
Deep learning approaches can be classified into Supervised, Semi-supervised and Unsupervised learning. Supervised learning means having a full set of labeled data while training a model, whereas unsupervised learning does not any labeled training dataset. Semi-supervised holds the middle ground where data is partially labeled and this influences the final outcome of the trained model. Depending on the task learning approaches are selected.

	\subsection{Supervised Learning}\label{chapter:2.4.1}
There are two main application where supervised learning is used: classification and regression problems. Classification simply means identifying the input data as the member of a particular class. The learned model is evaluated on how accurately the model can classify an unseen data. In regression problems continuous data which depends on lots of variables are inspected. Classic example of regression is linear regression where given a particular input would predict the output in future. That is why supervised learning is best suited to problems where a dataset is available with the reference point or the ground truth.

	\subsection{Unsupervised Learning}\label{chapter:2.4.2}
Although clean and perfectly labeled datasets are desirable in case of learning, it is not always the case. In unsupervised learning, the model learns the internal structure to discover unknown relationships within the unlabeled input data. The training dataset does not have any ground truth or correct answer. Unsupervised learning extracts useful features automatically by analyzing the data structure. Depending on the task, unsupervised learning can organize the data in different ways: Clustering, Anomaly detection, Association, Autoencoders, etc. In clustering, the unsupervised learning model looks for similar data and groups them. Anomaly detection means find the outlier in the dataset. By looking at some attributes of a data points, an unsupervised learning model can predict others key attributes to which they are associated. Autoencoders takes the input then compress the data into a code and try to recreate the input from the code. There are various types of autoencoders that are useful for different application. Denoising Auto-encoders can remove noises from a corrupted image. Autoencoders has various application image processing and analysis such as dimensionality reduction where data from high feature space is projected to a low feature map \cite{GoodfellowlanBengioYoshua2016}.  	
	
	\section{Transfer Learning}\label{chapter:2.5}
A large data is always desirable for any classification task to build a robust model which has learned generic feature. But in most cases, such well-labeled data is very rare for example in medical domain getting proper annotated data is expensive and requires necessary workload of experts in radiology. Transfer learning is the most common and effective strategy to overcome the problem of training a model on small dataset. The idea behind is to take the weights of a pretrained network that is trained on large datasets like ImageNet which contains \textit{1.4 million} images with 1000 classes and reuse them for a task of interest. Generic features learned on a large dataset by a network may be shared among disparate datasets as well. The learned generic feature from a large dataset is a unique advantage of ConvNets that allows its adaptability in many applications with small data. Many models trained on large datasets like ImageNet are publicly available such as AlexNet \cite{Krizhevsky2012}, VGG \cite{Simonyan2015}, RestNet \cite{He2016}, GoogLeNet \cite{Szegedy2015} that can be transferred for different applications.

	\begin{figure}[H]
		\centering
		\includegraphics[width=0.7\textwidth]{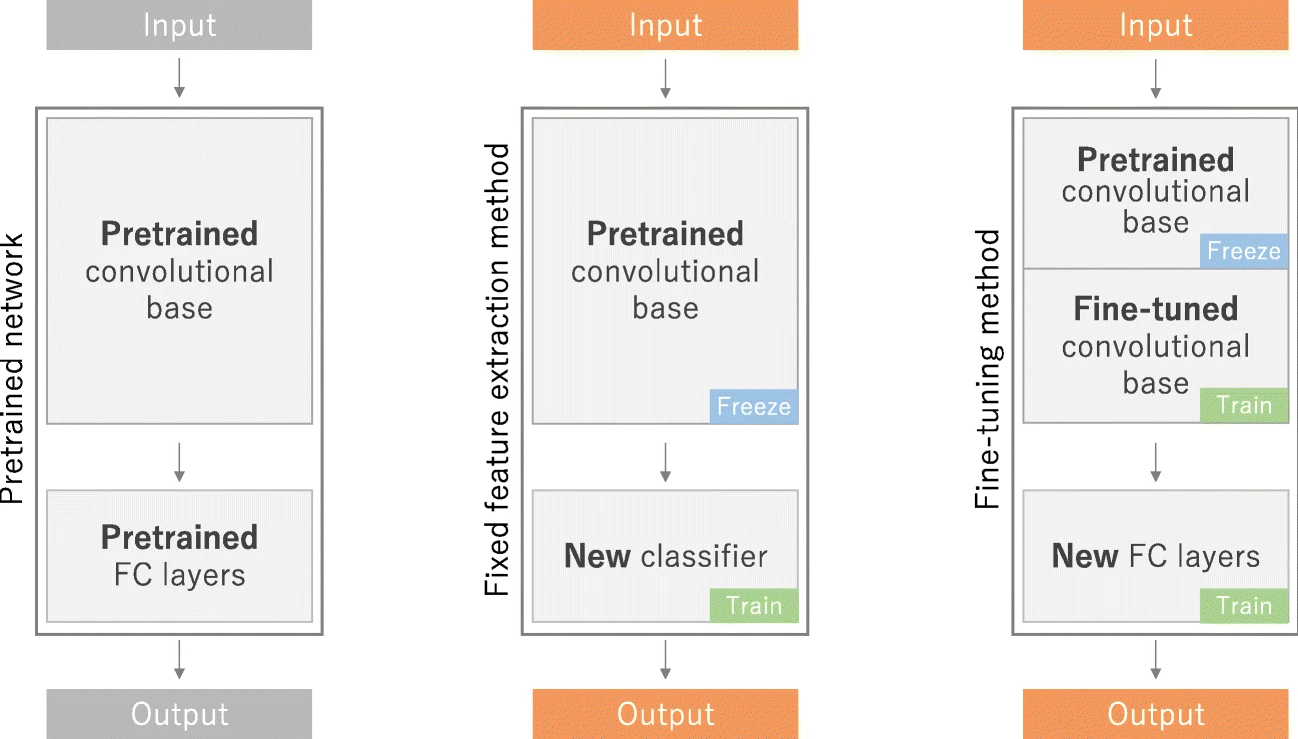}
		\caption[Fixed feature extraction and fine-tuning method of transfer learning.]{Fixed feature extraction and fine-tuning method of transfer learning \cite{Yamashita2018}.}
		\label{fig:fig2_8}
	\end{figure}

\noindent	
Transferring weights from a pretrained network can be fixed or partial feature extraction. The former is the process of transferring all the layers from the trained model which contains series of Conv layers, pooling layers except fully-connected layers. On top of the fixed feature extractor any classifier network can be added so suit the application specification. Partial feature extractor also known as fine-tuning method extract feature layers of the pretrained model similar to the previous method but only some earlier layers are kept frozen while training.

	\chapter{Methods and Concepts}\label{chapter:3}
This chapter explains the overall approaches and ideas that were utilized to achieve the goal of the thesis. Although the work started with implantation of various 3D ConvNets for classification task which required 3D data acquisition, data preprocessing, training and evaluation of the models, this chapter only discusses the methods and concepts that were used to extract the learning abstractions of the trained 3D ConvNet model. The implantation parts are discussed in details in the next section of the report (Chapter \ref{chapter:4}). Simply having a higher accuracy model without understanding the inner working is not enough in the domain of medical imaging as slight misdiagnosis can lead to severe costs. The methods discussed here will give us an in depth look into the model’s inner workings that would help us interpret and understand them better. The popularity and adaptability behind the ConvNets are the characteristic of how they actually work. Unlike other deep learning models, ConvNets are supposed to be less inscrutable because of their structure and function.  Although complete transparency of ConvNets are still debatable, there are many studies have been proposed that help us to have a better understanding of the network and optimize for better performance as well.

	\section{Filters/Kernels and Feature Maps}\label{chapter:3.1}
As discussed earlier ConvNet architecture is comprised of several Conv layers along with some pooling and fully-connected layers. Each Conv layer performs convolution operation on the given input with some defined kernels and outputs a feature map. The feature maps are also known as intermediate activations as output of a layer is called activation. The kernels are learned during training of the network that extract features from an input image. The feature map after the first Conv layer which is obtained from the convolution operation on input image becomes the input for the next Conv layers and finally reaches classification layer through (optional) fully-connected layers (See figure \ref{fig:fig2_3}). Feature maps after each Conv layer can be visualized to see how the input is transformed passing through the network. The filters/kernel of a layer determines what pixel values of an input image will be focused for that Conv layer. The depth of feature maps depends on the number of filters used in a Conv layer. For 2D ConvNet each filter will generate a 2D feature map but as multiple filters are used for a single layer usually feature map ends up as a volume that has depth equals to number of filters used for that layer. In case of 3D ConvNet generated feature map for each filter is also 3D. As defined filters (can be of different size) and extracted feature maps are matrices that are encoding independent features by visualizing them as image we can have an intuition about the learned features by the network. Visualizing filters and feature maps are quite simple. We can extract filters of each Conv layer and can plot them as 2D (slices in case of 3D filters) image. An input image is passed through Conv layers and record the intermediate activations (feature maps) that can be plotted as 2D (slice of 3D volumetric feature maps) image as well. Similarly feature map of pooling layers can also be extracted and investigated.    

	\section{Layer-wise Relevance Propagation }\label{chapter:3.2 }
Layer-wise Relevance Propagation of LRP \cite{Bach2015} is a technique for better interpretability of a neural network. LRP follows a conservation property which is analogous to Kirchhoff’s conservation laws for electrical circuits. LRP’s conservation property propose that what has been received by a neuron in a layer must be equally redistributable to the lower layer in the network. LRP technique can be applied structured model like ConvNets and other neural networks \cite{Bohle2019} \cite{Bach2015} \cite{Anders2019}. LRP propagates a class prediction backwards in the network following the designed local propagation rules.

	\begin{figure}[H]
		\centering
		\includegraphics[width=0.9\textwidth]{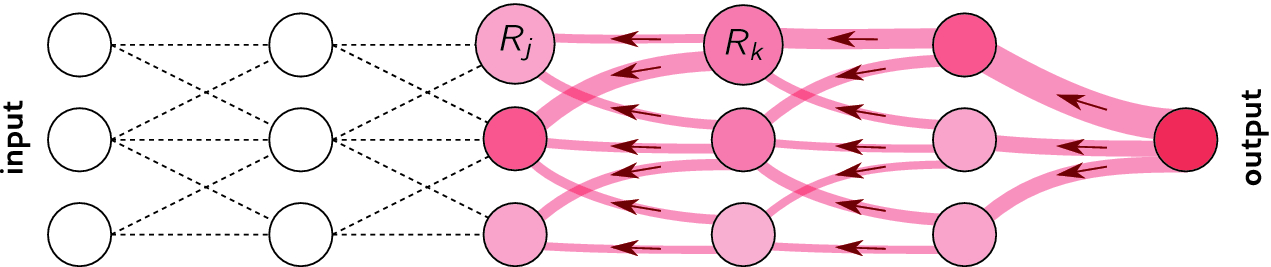}
		\caption[LRP procedure showing higher layer neurons are redistributing as much as it has received from lower layer.]{LRP procedure showing higher layer neurons are redistributing as much as it has received from lower layer \cite{Montavon2019}.}
		\label{fig:fig3_1}
	\end{figure}
\noindent
In the figure \ref{fig:fig3_1}, \textit{j} and \textit{k} are the neuron of two consecutive layers of the network. Propagating relevance score of a higher layer \( (R_k)_k \) on to a consecutive lower layer neuron can be calculated as,

	\begin{equation}\label{eq:eq3_1}
  		R_j = \sum_k\frac{a_j w_{jk}}{\sum_j a_j w_{jk}}\times R_k
	\end{equation}

\noindent
Where, \( a_j w_{jk} \)  is the activation of neuron \textit{j} multiplied by the weight between \textit{j} and \textit{k}, also \( \sum_j a_j w_{jk} \) keeps the conservation property. The propagation runs until the input feature found. If all the neurons on the network is considered then we can verify the conservation property by,  \( \sum_j R_j = \sum_k R_k \)  and global conservation property by, \( \sum_j R_j = f(x) \), where \( f(x) \) is a class prediction of the network. As a result, the class prediction of the network is decomposed into pixel-wise relevance which indicates how much a node contributes to the final prediction decision. There are other improvements has been also proposed for more robust redistribution like LRP-\(\epsilon \) \cite{Bach2015}, where a small positive number \( \epsilon \) is added to the denominator which absorb some relevance when a neuron has weak or contradictory activation. Another rule is known as LRP-\( \gamma \), that favor the positive contribution over negative contribution. Treating positive and negative contribution is known as LRP-\(\alpha\beta \) rule. If we choose \( \gamma\rightarrow\infty \) in LRP-\(\gamma \) then it becomes equivalent to LRP-\(\alpha_1\beta_0 \) \cite{Bach2015}. All the rules keep the conservation property and gives the insight on how a node in a network contributes to the final decision of the model. Using LRP methods in 3D ConvNet we can generate an 3D output providing an 3D input sample then comparing the output with original input would give used regions that are strongly contributes to the final classification decision. Comparing 3D images is not feasible so multiple 2D slices of input image that are relevant are compared with generated relevance map.

	\section{Activation Maximization}\label{chapter:3.3 }
Activation maximization aims to maximize the activation of certain neurons. The reason behind this idea is to find a pattern to which a unit responds maximally could provide initial insight of what the unit is doing or has learnt. When a ConvNet is trained to, it tries to minimize the loss function by adjusting weights and biases where input and desired output is remained unchanged. After training, for a given neuron has maximum activation for some input sample(s). In order to find maximum activation of the unit, input sample(s) can be searched to which the unit responds maximally. But this is difficult because there is no way of choosing how many input samples to combine and moreover finding common patterns among the inputs are desired rather than the input itself. To solve this difficulty, more general view that has been already been used to optimize the weights can be utilized. Instead of trying to find patterns from the input datasets, the input can be modified such that it maximizes certain neurons by keeping the weights and desired output constant – much like an optimization problem. If \( \theta \) is denoted as the learnt weights and biases of the network and \(x\) is the input then, activation can be defined as \( h_{ij} \) \( (\theta,x) \) for unit \textit{i} of layer \textit{j}. In activation maximization, the weights, biases and desired output are kept constant so, idea of looking for input that has maximum activation for given unit can be defined as \cite{Erhan2009}, 

	\begin{equation}\label{eq:eq3_2}
  		x^* = arg_{x s.t. \ ||x||=\rho} \ max \ h_{ij}(\theta,x) 
	\end{equation}

\begin{figure}[H]
		\centering
		\includegraphics[width=\textwidth]{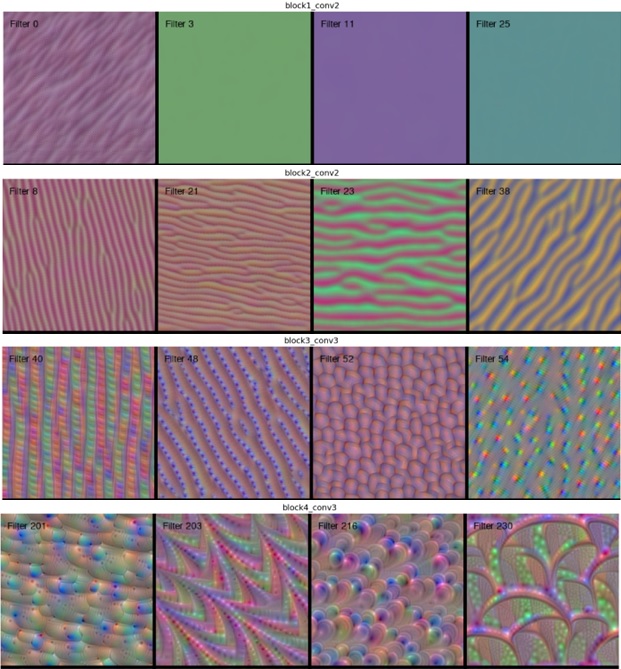}
		\caption[Filter visualization at different layer of VGG16 network.]{Filter visualization at different layer of VGG16 network.}
		\small Image adapted from [\url{https://towardsdatascience.com/applied-deep-learning-part-4-convolutional-neural-networks-584bc134c1e2} ]
		\label{fig:fig3_2}
	\end{figure}

\noindent
Now, after finding the gradient of \( h_{ij} \) \( (\theta,x) \) in input space, \textit{x} can be moved in the direction of gradient which is opposite to what backpropagation does. Minimization of the activation during training by moving to the opposite of gradient is called gradient descent. Similarly, maximization of the activation  can be found by moving in the direction of the gradient which is gradient ascent. Visualizing learned filters are easy but as this is often very small in dimension (i.e. \( (3\times3) \), \( (5\times5) \)) they are barely interpretable. Visualizing the filters of the network with \textit{Activation Maximization} technique could be beneficial to understand what features and pattern the network has learned. Starting with an input image rather than random noise can also generate maximum activation patterns in the image. Output of this technique is also referred as \textit{DeepDream} \cite{GoogleAlexander2015} images. Figure \ref{fig:fig3_2} shows some activation maximization results at different Conv block of VGG16 \cite{Simonyan2015} network starting with a random noise as input. It can be observed from figure \ref{fig:fig3_2} that initial Conv blocks of the network are encoding simple patterns and more deeper layers are encoding complex patterns. Although these patterns are not interpretable   directly, but are indication of encoding high level features in deeper and low level features in initial layers.      

	\chapter{Implementation  }\label{chapter:4}
This section of the thesis describes the overall implementation steps of different 3D ConvNets that were used for the classification and learned feature extraction task. In the previous section (Chapter \ref{chapter:3}), the methods that were used to extract learning abstraction of the ConvNets after training are discussed and here, earlier part that is training of ConvNets are explained. The conducted work for the thesis project is outlined step by step in this chapter. The work started with data acquisition and followed by data preprocessing, building CNN structures and training them. Although there were multiple versions of CNN model with different network architecture was used for the experiment not all of them are described individually, instead a standard 3D CNN model is described. Variations on this structure are described in the next chapter along with their evaluation.    
	
	\section{Data Acquisition}\label{chapter:4.1 }
The ultimate goal of this thesis work was to learn the shape features and abstraction in 3D ConvNet to detect the Alzheimer’s Disease. Keeping that in mind at first, 3D ConvNets with simulated 3D data were investigated. The reason behind choosing simulated simple 3D data was to have better understanding of the shape feature learned by the network. Finally, 3D MRI data from Alzheimer’s Disease Neuroimaging Initiative (ADNI) and Australian Imaging Biomarkers and Lifestyle Study of Ageing (AIBL) were used to train the 3D ConvNets and learned features were investigated with different methods explained in chapter  \ref{chapter:3}. As discussed before, available annotated 3D datasets are quite limited compared to 2D datasets. Moreover, there is no standard representation methods for 3D data and available tools for processing as well as visualizing these data are also limited. Princeton ModelNet \cite{Wu2015} is one of the most used 3D dataset for 3D object detection that provides comprehensive clean 3D CAD models. For the initial task, subset of ModelNet40 dataset and CAD models that are available online were used. Details about these data are discussed in the following section.      
	
	\subsection{3D Object/Models}\label{chapter:4.1.1 }	
\begin{figure}[H]
		\centering
		\includegraphics[width=\textwidth]{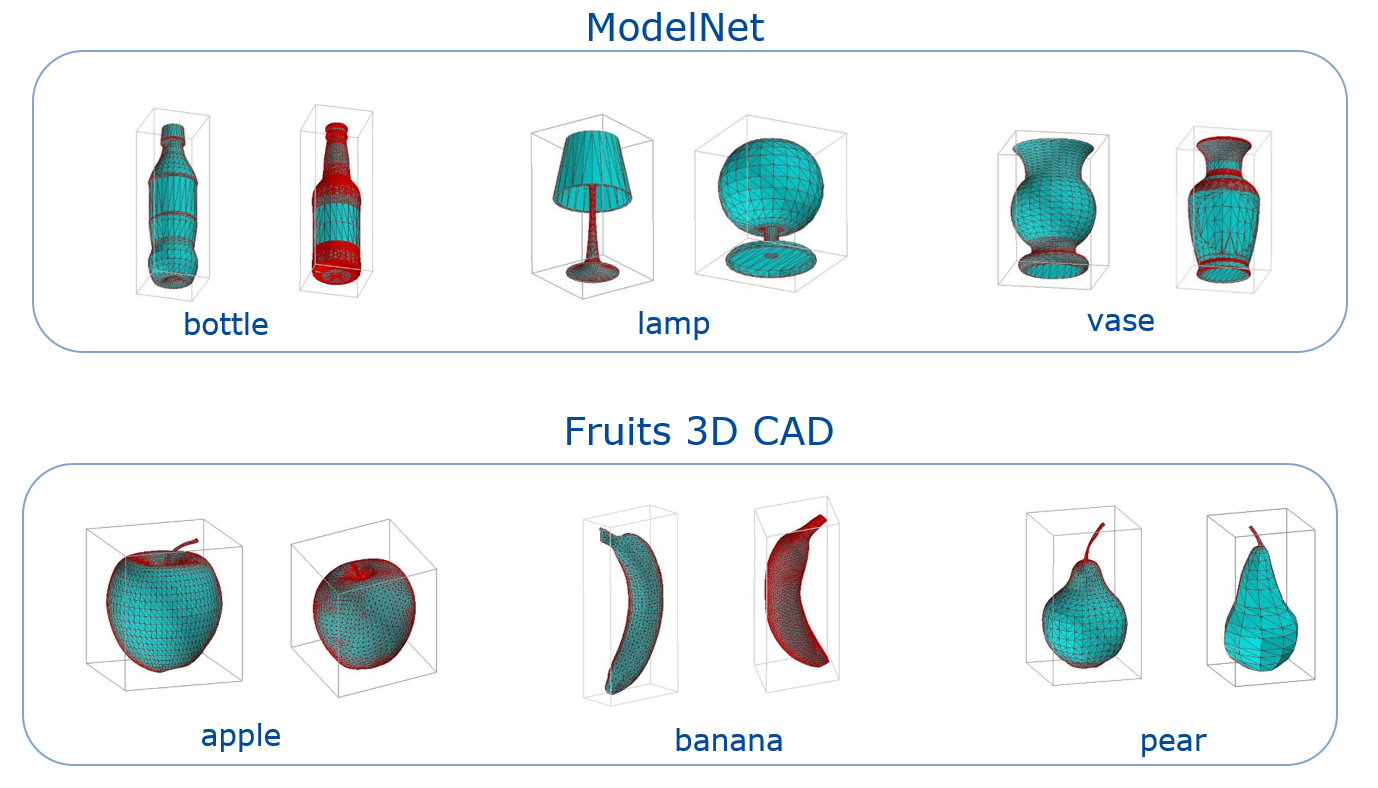}
		\caption[ModelNet dataset and 3D CAD fruits objects.]{ModelNet dataset and 3D CAD fruits objects.}
		\label{fig:fig4_1}
	\end{figure}

Princeton ModelNet consist a list of most common object categories, 662 to be exact. Although the entire dataset can be downloaded, they also offer ModelNet10 and ModelNet40 subset which consist 10 and 40 categories respectively. The CAD models offered are in Object File Format (OFF) which represent the geometry of a model by defining model’s surface polygon. Tools (Matlab functions) required to read and visualize the CAD model are provided as well in Princeton Vision \& Robotics Toolkit (PVRT). Although ModelNet40 consist of many complex structures only three categories named Bottle, Lamp and Vase which were chosen for the thesis task. All three categories have somewhat round and elliptical shape which can be useful to understand the learned features by a ConvNet that is trained on them. Each CAD model was manually handpicked to make sure they are human distinguishable and different from each other. Similarly, three types of fruits (apple, banana and pear) 3D CAD models freely available online were also downloaded. Downloaded CAD models format includes OFF, OBJ, STL which are common in 3D printing uses. Figure  \ref{fig:fig4_1} shows some of the CAD models from ModelNet and Fruits that were used. 

\noindent	
Table \ref{table:table4_1} sshows how many objects for each category were selected for training 3D ConvNet. ModelNet dataset comes with train and test data separated. For fruits there were few training samples than ModelNet.

	\begin{table}[H]
	\centering
	\begin{tabular}{||c|c|c|c||}
	\hline
	\multicolumn{4}{||c||}{ModelNet}\\ \hline
    \hline
	Categories & no. train sample & no. of test sample & Format \\ \hline
	Bottle & 80 & 10 & \multirow{2}{*}{OFF}\\
	\cline{1-3} 
	Lamp & 60 & 10 & \\ \cline{1-3}
	Vase & 80 & 10 & \\ \hline
 	\end{tabular}
 	
	\begin{tabular}{||c|c|c|c||}
	\hline
	\multicolumn{4}{||c||}{Fruits}\\ \hline
    \hline
	Categories & no. train sample & no. of test sample & Format \\ \hline
	Apple & 19 & 6 & \multirow{2}{*}{\vtop{\hbox{\strut OFF,}\hbox{\strut STL,} \hbox{\strut OBJ}}}\\ \cline{1-3}
	Banana & 19 & 6 & \\ \cline{1-3}
	Pear & 15 & 2 & \\ \hline
 	\end{tabular} 	
 	
	\caption[ModelNet and Fruits CAD models from training the 3D CNN.]{ModelNet and Fruits CAD models from training the 3D CNN. }
  	\label{table:table4_1}
	\end{table}

	\subsection{3D MRI – ADNI and AIBL}\label{chapter:4.1.2 }
\begin{figure}[H]
		\centering
		\includegraphics[width=0.9\textwidth]{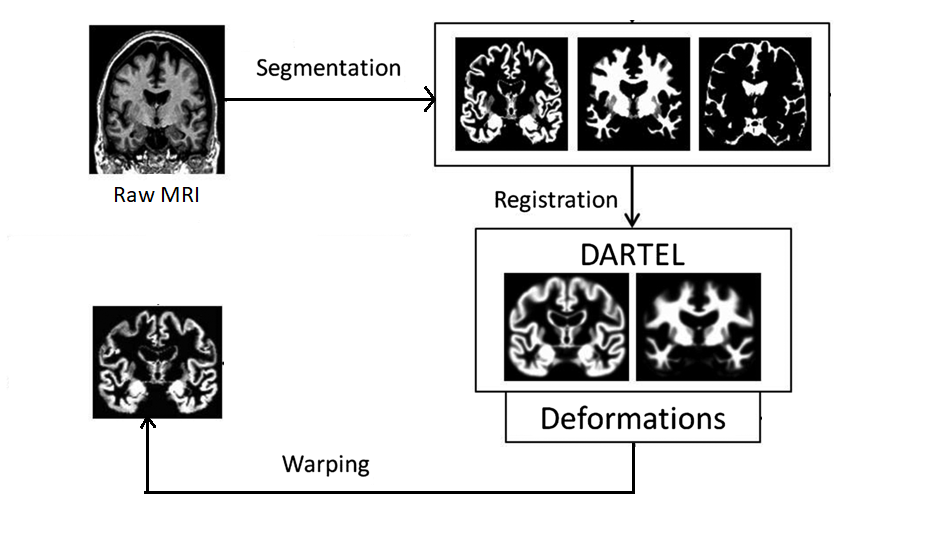}
		\caption[ADNI data preprocessing steps.]{ADNI data preprocessing steps. Figure is adopted from \cite{Grothe2016}.}
		\label{fig:fig4_2}
	\end{figure}

In this study, publicly available dataset from Alzheimer’s Disease Neuroimaging Initiative (ADNI) was used. Since the lunch of ADNI more than a decade ago, ADNIs study aims to develop clinical, imaging, genetic and biochemical biomarkers that would help to early detect and tracking of Alzheimer’s Disease (AD). ADNI dataset undergo standardized image processing steps to improve the data uniformity as they were acquired across multicenter scanner platforms. Original \(T_1\)- weighted MPRAGE sequences of ADNI-GO/-2 MRI data were scanned on multiple 3T MRI scanners following different platform specific protocols \cite{Grothe2016}. Preprocessing starts with automatically segmenting MRI scans into gray matter, white matter and cerebrospinal fluid partitions of 1.5 mm isotropic voxel size with \textit{VBM8} segmentation toolbox. Using \textit{DARTEL} algorithm these gray matter and white matter are registered high-dimensionally to an ageing/AD specific reference template. To preserve the total amount of gray matter before warping, resulting template are used to warp the gray matter segments and voxel values are modulate for volumetric changes introduced by normalization. Finally, overall segmentation and registration accuracy is confirmed by visual inspection of the gray matter maps. Figure \ref{fig:fig4_2} demonstrates simplified ADNI data preprocessing steps. After preprocessing these MRI scans are used for various research for Alzheimer’s Disease. Australian Imaging Biomarkers and Lifestyle Study of Ageing (AIBL) is also available from ADNI which undergoes same preprocessing steps before available for the public to be used for different research projects. Both ADNI and AIBL datasets were used for this study. 
	
\noindent	
ADNI and AIBL dataset consists of 662 and 621 MRI scans respectively which are categorized as Alzheimer’s Disease (AD), Mild Cognitive Impairment (MCI) and Normal Control (NC) subjects base on their test score to measure cognitive impairment known as Mini Mental State Examination (MMSE) \cite{Creavin2016}. Combining both data set a total 1283 (574 AD, 709 NC) MRI scan were found out of which 1038 (464 AD, 574 NC) scan were used for training followed by 126 (52 AD, 64 NC) for validation and 129 (58 AD, 71 NC) scans for testing. Both AD and MCI subjects were considered as positive AD for the training which helped to mitigate data imbalance problem. Table \ref{table:table4_2} shows demographic data for both ADNI and AIBI dataset separately. Values are presented as mean \( \pm \) Standard deviation.

	\begin{table}[H]
	\centering
	\begin{tabular}{||l|c|c|c||}
	\hline
	\multicolumn{4}{||l||}{ADNI(n=662)}\\ \hline
    \hline
		 & NC & MCI & AD \\ \hline
	Sample Size (Female) & 254(130) & 219(93) & 189(80) \\ \hline
	Age(\(\pm SD\)) & \(75.4\pm 6.5\) & \(74.1\pm 8.1\) & \(75\pm 8.0\) \\ \hline
	MMSE(\(\pm SD\)) & \(29.1\pm 1.2\) & \(27.6\pm 1.9\) & \(22.6\pm 3.2\) \\ \hline
 	\end{tabular}

	\begin{tabular}{||l|c|c|c||}
	\hline
	\multicolumn{4}{||l||}{AIBL(n=621)}\\ \hline
    \hline
		 & NC & MCI & AD \\ \hline
	Sample Size (Female) & 455(265) & 99(47) & 67(39) \\ \hline
	Age(\(\pm SD\)) & \(72.4\pm 6.2\) & \(74.3\pm 6.8\) & \(72.9\pm 6.5\) \\ \hline
 	\end{tabular} 	
 	
	\caption[Demographic of ADNI and AIBL datasets.]{Demographic of ADNI and AIBL datasets.}
  	\label{table:table4_2}
	\end{table}

\begin{figure}[H]
		\centering
		\includegraphics[width=0.9\textwidth]{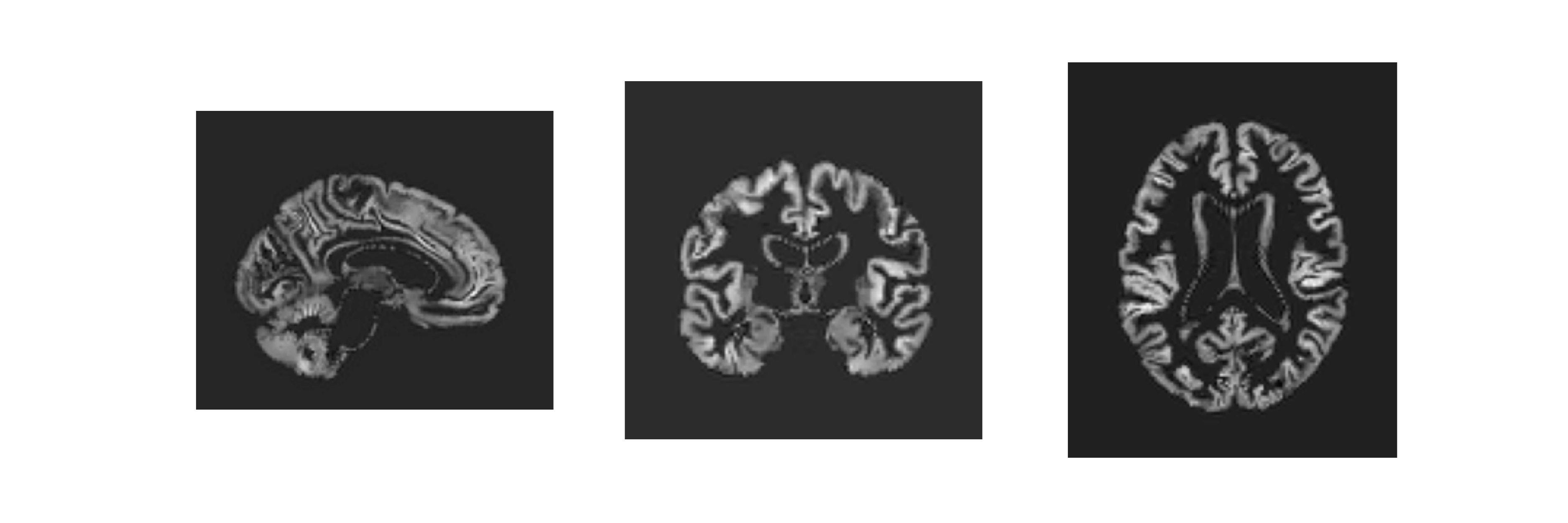}
		\caption[MRI scan of an AD subject (slice no. 55, 75, 60 from left respectively).]{MRI scan of an AD subject (slice no. 55, 75, 60 from left respectively).}
		\label{fig:fig4_3}
	\end{figure}

\noindent
Each MRI scan is a 3D Volume of grayscale intensity of dimension \( 121\times145\times121\) where \(145\) is the number of coronal slices of dimension \( 121\times121 \). Figure \ref{fig:fig4_3} shows a sample MRI scan from ADNI dataset.

	\section{Data Preprocessing}\label{chapter:4.2}
Input of a volumetric ConvNets usually requires volumetric Data. Although there are different methods for representing a volumetric data, \cite{Wu2015} \cite{Maturana2015} binary occupancy grid was used to encode the 3D shapes in this study because of the robustness of the structure than other representations. ModelNet and fruits CAD models were preprocessed to convert them from mesh structure to 3D volumetric data. As there were very few models per each category some augmentation was used to increase the training datasets. 3D MRI scans were also converted from intensity volumetric grid to binary volumetric grid using a threshold value. Following section describes the steps and tools that are used for preprocessing of the data.

	\subsection{Voxelization}\label{chapter:4.2.1}
To convert 3D mesh models into a 3D volumetric data a tool named \textit{binvox} \cite{patrickmin} was used.This is a very straight-forward command line program that reads a 3D model file and rasterized into a binary 3D voxel grid and writes the resulting voxel file as  \textit{filename.binvox}. Converted \textit{.binvox} file then can be read using a python library (\textit{binvox-rw-py} \cite{dimatura}). To convert all 3D models at once a simple \textit{shell script} was used. Voxelization can be done choosing any dimension while converting the model file. Higher dimension would result more accurate object representation with the cost of higher memory. Different dimension was investigated to find the optimal model representation while keeping the memory use manageable. Figure \ref{fig:fig4_4} below shows a 3D CAD model with different dimension.

	\begin{figure}[H]
		\centering
		\includegraphics[width=0.8\textwidth]{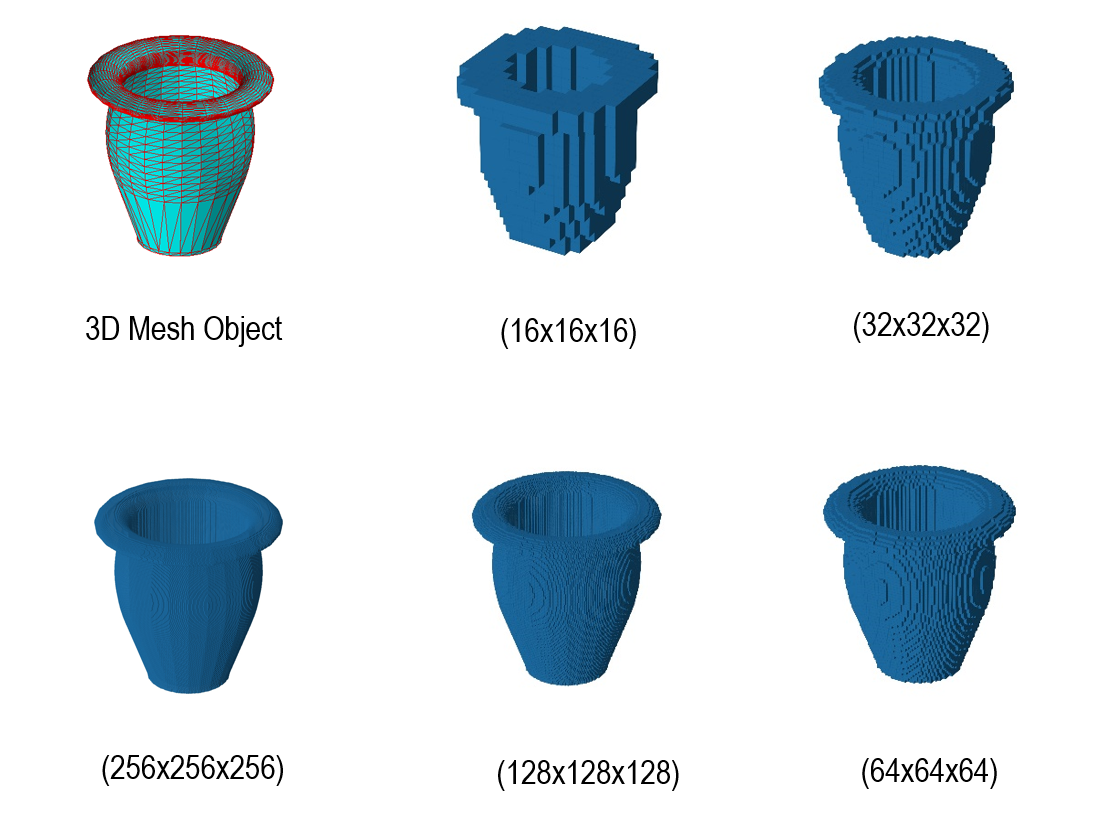}
		\caption[3D mesh model converted into 3D voxel grid with different dimension.]{3D mesh model converted into 3D voxel grid with different dimension.}
		\label{fig:fig4_4}
	\end{figure}

\begin{figure}[H]
    	\centering
		\includegraphics[width=0.8\textwidth]{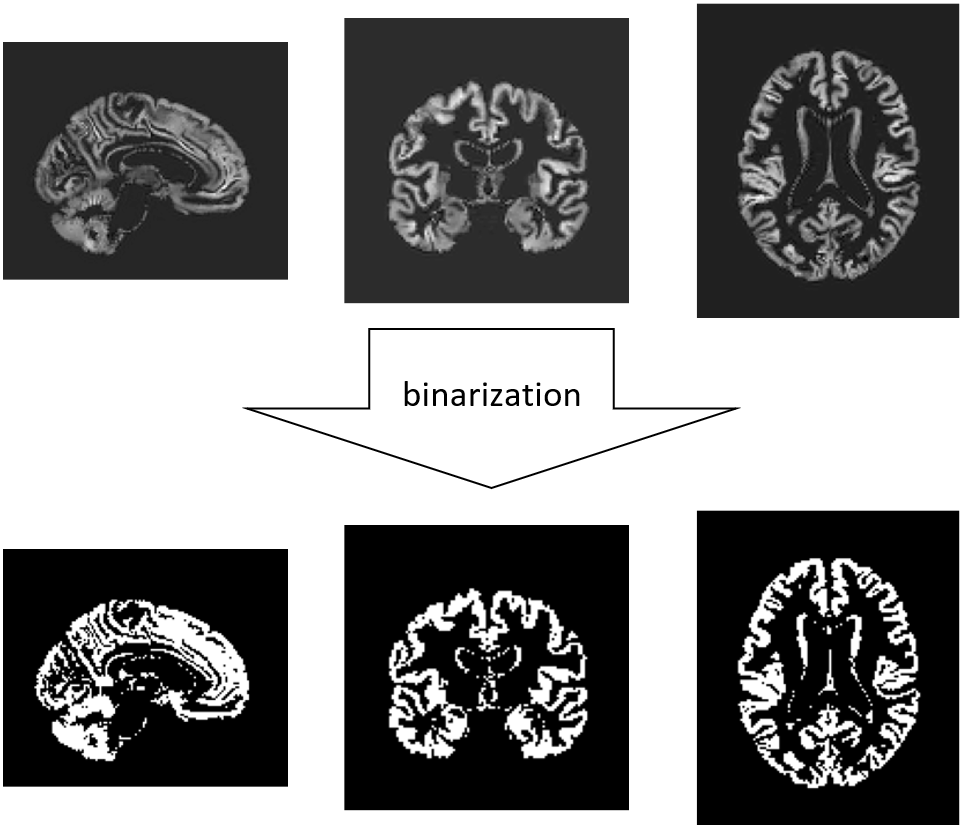}
		\caption[MRI scan before and after binary voxelization.]{MRI scan before and after binary voxelization.}
		\label{fig:fig4_5}
	\end{figure}

\noindent
From the figure \ref{fig:fig4_4} it can be observed that higher dimensional voxel 3D grids resulted near accurate to the original mesh structure keeping all the depth information, whereas lower dimensional 3D grids were less accurate and their depth information were also lost. Considering memory consumption and depth information, model with dimension \( (64\times64\times64)\) gave the best trade-off. 
ADNI and AIBL MRI scans were also converted into 3D binary voxel grid. MRI scans can be manipulated using python package \textit{NiBabel} \cite{NiBabel} which provides \textit{read/write} access to some common medical and neuroimaging file formats (GIFTI, NIfTI1, NIfTI2, CIFTI-2, MINC1, MINC2, AFNI BRIK/HEAD, MGH and ECAT)  including analysis. MRI scans used in this study were in textit{NIfTI} file format. As this thesis goal was to find the learned pattern by the ConvNet trained on MRI scans it was desirable to use the whole brain model rather than selecting a region of interest (ROI) or segment of brain scan. But whole brain model requires significant amount of memory and also time consuming. By converting the 3D intensity volumetric data into binary volumetric data, reduced the memory requirement significantly while keeping the structural information intact. Figure \ref{fig:fig4_5} illustrates an MRI intensity volumetric scan that was converted to binary voxel grid. For binarization threshold value 0.4 was used that means intensity of 0.4 and overs were converted to 1 otherwise 0.

	\subsection{Data Augmentation}\label{chapter:4.2.2}
Compared to state-of-the-art 2D ConvNets that are trained on millions of training samples, data available for this study was insignificant. Less numbers of training samples can lead to overfitting of the model. Data augmentation can help to reduce the overfitting problem by increasing the number of training samples. While there are multiple ways of augmenting the data such as rotating, shifting, cropping, flipping, translating, etc., not all of them were helpful for the type of data this study was working on. Simple shifting along different axis and diagonally was one of the methods that was used here that shifts the entire object along defined axis and value while keeping the model structure unchanged. After checking all the training samples manually, a range for shifting range was has been selected to make sure object was not getting out of the occupancy grid. The empty spaces after shifting were padded with 0. Flipping is another way of augmenting the data but as most of the ModelNet and Fruits CAD objects were symmetrical along some axes, flipping along those axes would result same output. This wasis also taken care of by looking into individual object manually. 

	\begin{figure}[H]
	\centering
	\begin{subfigure}[b]{0.4\textwidth} 
    	\includegraphics[width=\linewidth]{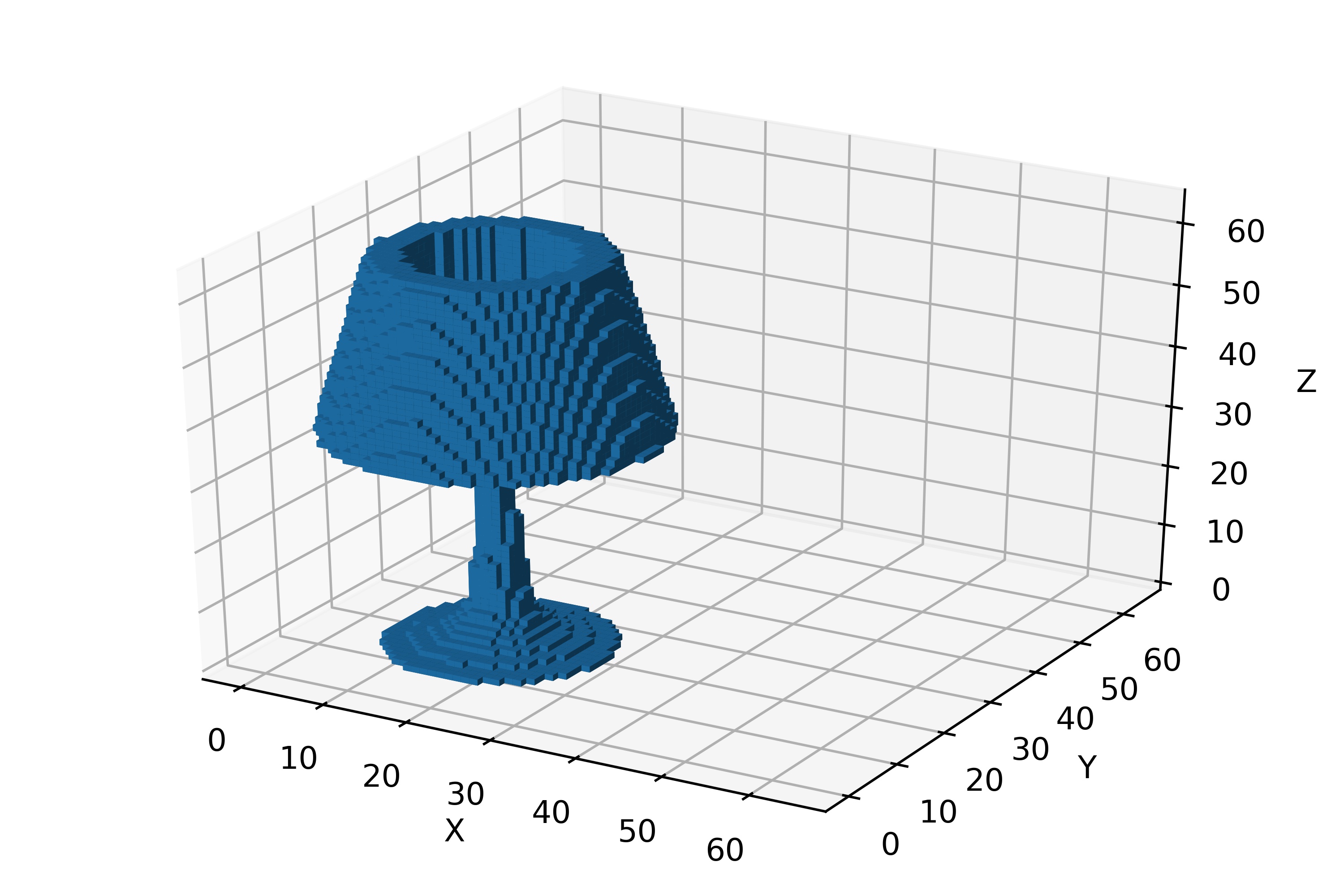}
    	\caption[]{original} 
	\end{subfigure}%
	\quad
	\begin{subfigure}[b]{0.4\textwidth} 
    	\centering
    	\includegraphics[width=\linewidth]{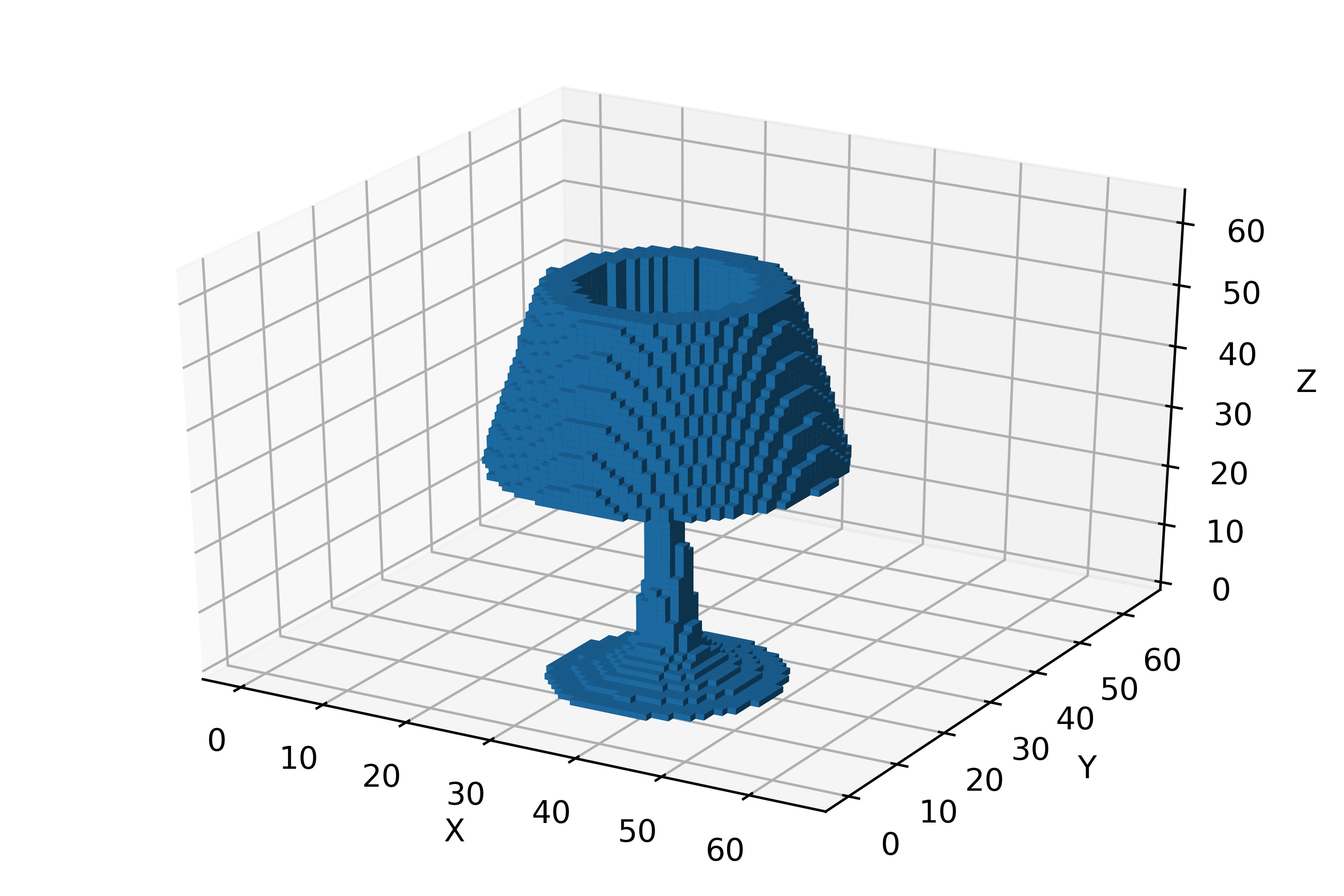}
    	\caption[]{shifted along X} 
	\end{subfigure}

	\begin{subfigure}[b]{0.4\textwidth} 
    	\includegraphics[width=\linewidth]{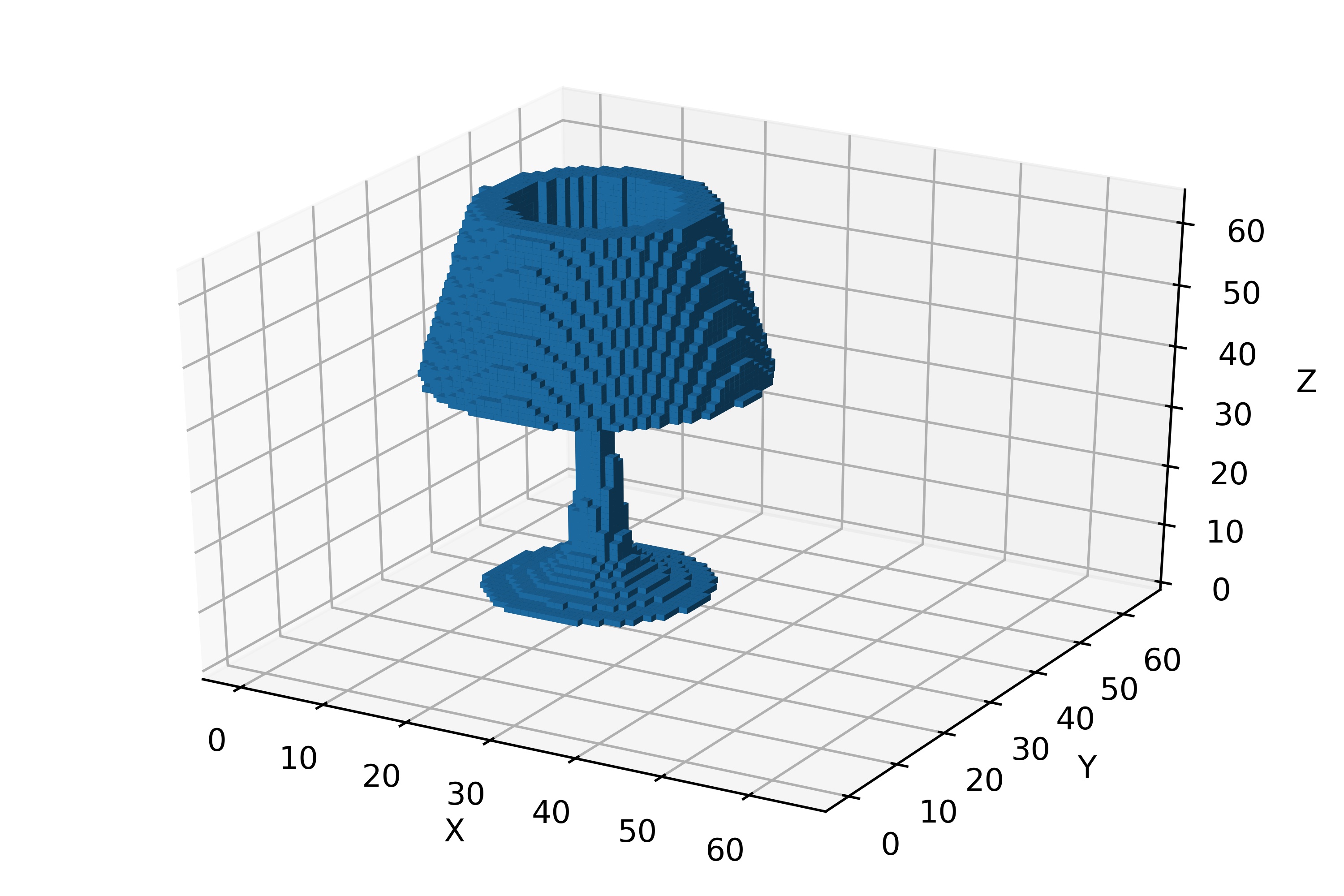}
    	\caption[]{shifted along Y} 
	\end{subfigure}%
	\quad
	\begin{subfigure}[b]{0.4\textwidth} 
    	\includegraphics[width=\linewidth]{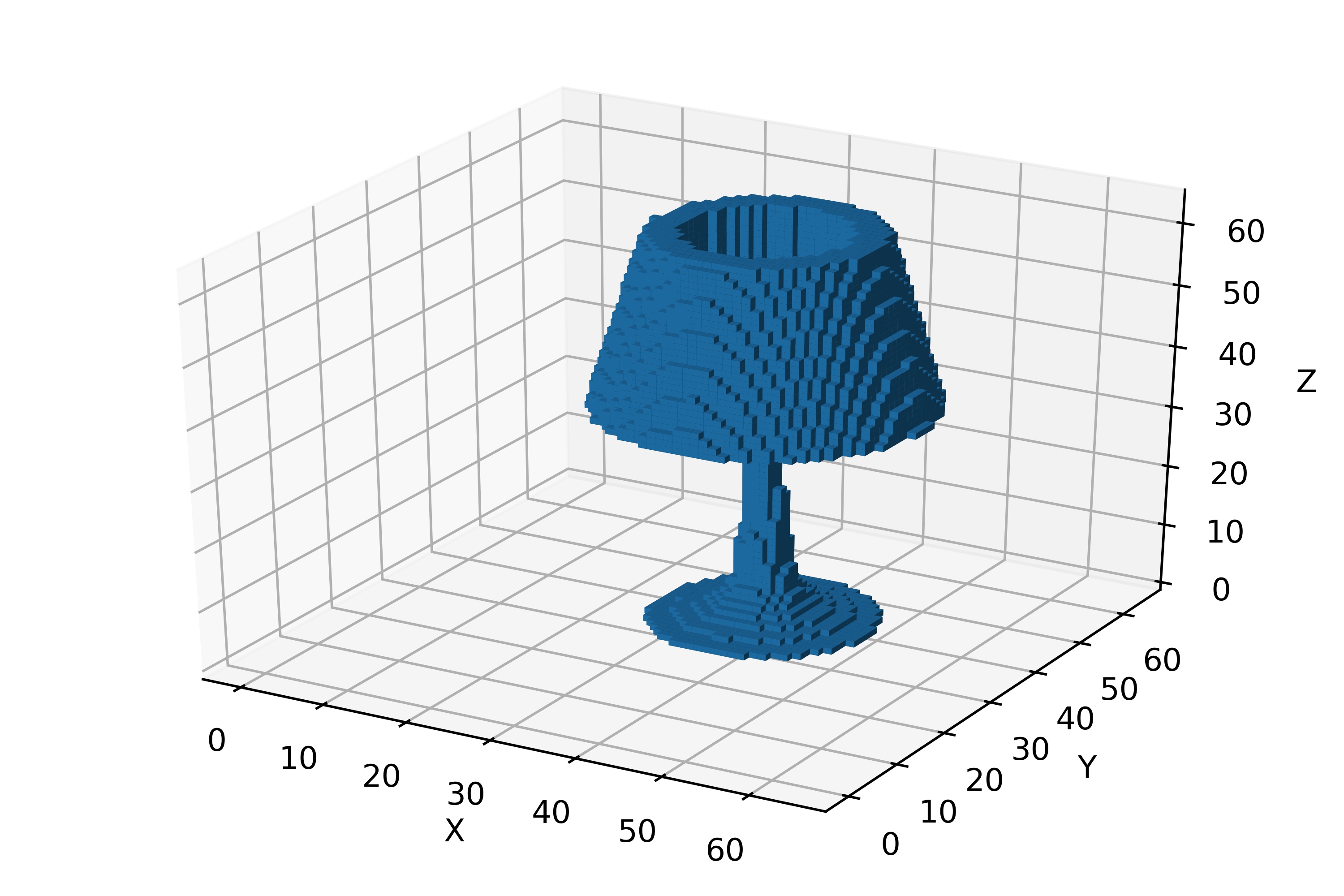}
    	\caption[]{shifted diagonally XY plane} 
	\end{subfigure}%
    \caption[Augmentation of a training sample Lamp. Shifted 20 voxel along X, Y, and XY surface.]{Augmentation of a training sample Lamp. Shifted 20 voxel along X, Y, and XY surface.} 
	\label{fig:fig4_6}
	\end{figure}

\noindent
Figure \ref{fig:fig4_6}(a) shows that most of the occupancy grid of the model is empty on the \textit{XY} surface so shifting along this surface would not crop the model but along \textit{Z} shifting is not desirable which would crop the model. For MRI scans simple shifting, flipping and cropping augmentation methods were used. Cropping for MRI scans was only used for training convolutional autoencoder which will be discussed later.

	\section{3D CNN Structures}\label{chapter:4.3}
\begin{figure}[H]
    	\centering
		\includegraphics[width=\textwidth]{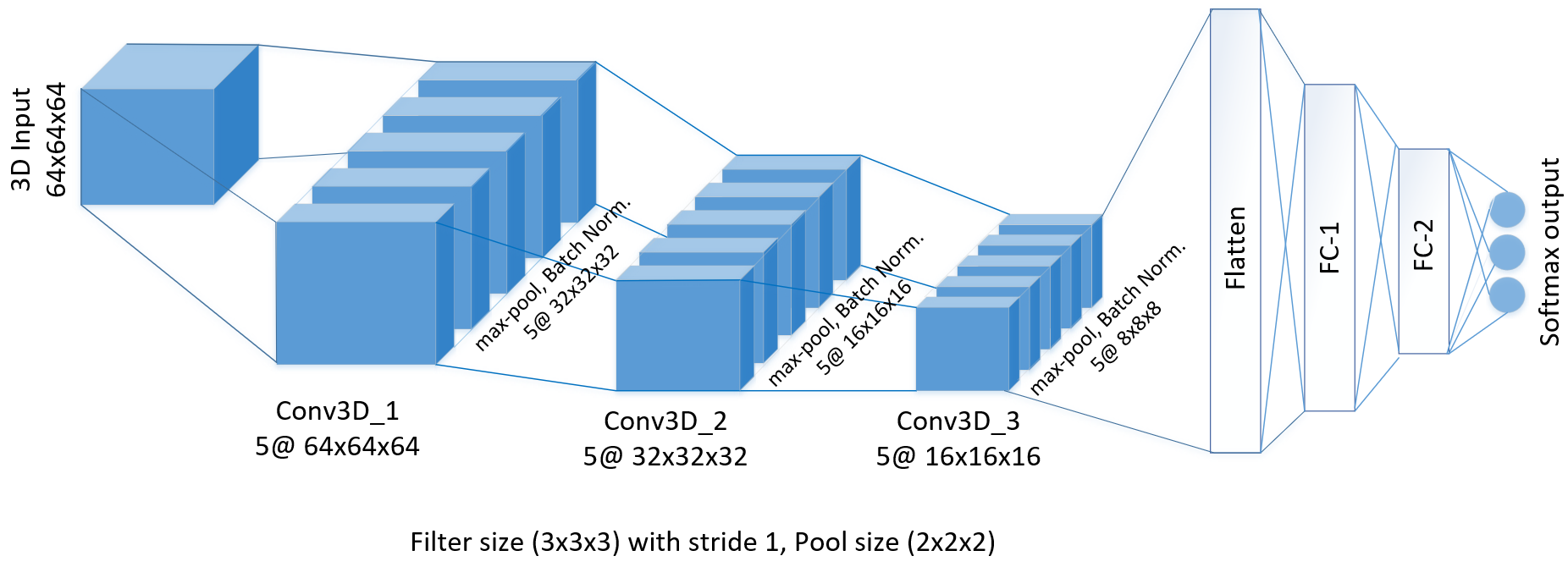}
		\caption[3D ConvNet with 3 Conv blocks shows how input of size \((64\times 64\times 64)\) transforms through the network. After each conv layer max-pooling operation reduces the input space by half. ]{3D ConvNet with 3 Conv blocks shows how input of size \((64\times 64\times 64)\) transforms through the network. After each conv layer max-pooling operation reduces the input space by half. }
		\label{fig:fig4_7}
	\end{figure}

The objective of this study was to evaluate the shape features learned by a 3D ConvNet with respect to the required layer structure. There were several network structures were trained and evaluated. Here, base structure of the 3D ConvNet is described which was modified to conduct different experiments. The base ConvNet was comprised with three to four 3D Conv layers. Each Conv layer was followed by a 3D max pooling layer, Activation layer (ReLU). After the last Conv layer, data was flattened connected with one or more fully-connected layers with ReLU activation function. The last layer always was a fully-connected layer with a Softmax activation. Sometimes after last Conv layer the output was connect directly to the classification layer. The number of filters, filter sizes, pool sizes were also changed for different structures. The study aims to find how based on different network structures and different hyperparameters, shape feature encodings and learning abstractions changes. In chapter 5, changes in the base 3D CNN along with their outcome is compared. A 3D network structure is presented in figure \ref{fig:fig4_7}.

\noindent
This 3D ConvNet was true with some slight modifications for supervised learning. For unsupervised learning the network structure was changed which will be discussed in next section (section \ref{chapter:4.4}).

	\section{Convolutional Autoencoder and Transfer Learning}\label{chapter:4.4}
Autoencoders (AE) are the unsupervised neural networks where output is same as the input. AE applies backpropagation to set the target to be equal to the inputs. Typically, application of AE includes information retrieval and dimensionality reduction which was one of the first use of representational learning. Discriminative feature representation in lower dimensional space improves the performance of the classifier \cite{Masci2011}. Besides, the learning process of AE is unsupervised that means it does not require labeled data. The usefulness of localized spatial features for 2D/3D image analysis has already been proven in research, so AE that does not acknowledge these features are not suitable. Convolutional Autoencoders (CAE) which extends AE, considers spatial features of image data just like a ConvNet does. A CAE is comprised of an encoding and a decoding phase. In encoding phase inputs are passed through the network to represent them into a lower-dimension feature space known as code, then the encoded data are reconstructed to its original space during decoding phase.
 
Figure \ref{fig:fig4_9} illustrates a CAE where an input image was encoded to a lower-dimensional feature space using three Conv blocks. Each Conv block in encoder part was comprised of a Conv layer with ReLU activation followed by a max-pooling and a batch normalization layer. In decoder part of the CAE, encoded data was transformed with Conv blocks that were same as encoder part but instead of max-pooling an upscaling layer was used. Motivation behind implementing CAE in this study was to transfer the learned features by the network for supervised learning.  

	\begin{figure}[H]
    	\centering
		\includegraphics[width=0.6\textwidth]{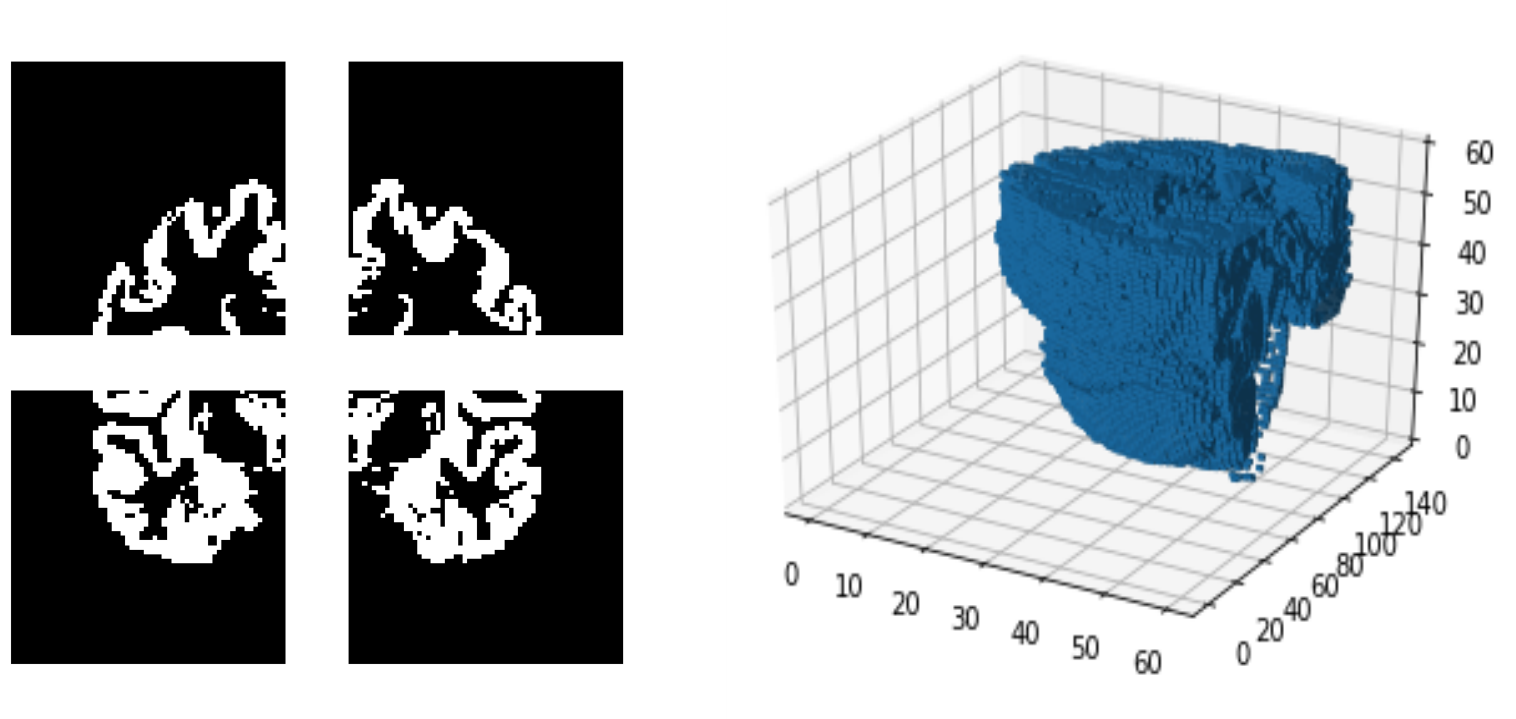}
		\caption[Divided parts of along coronal slice (left), lower left patch volume (right).]{Divided parts of along coronal slice (left), lower left patch volume (right).}
		\label{fig:fig4_8}
	\end{figure}

	\begin{figure}[H]
    	\centering
		\includegraphics[width=\textwidth]{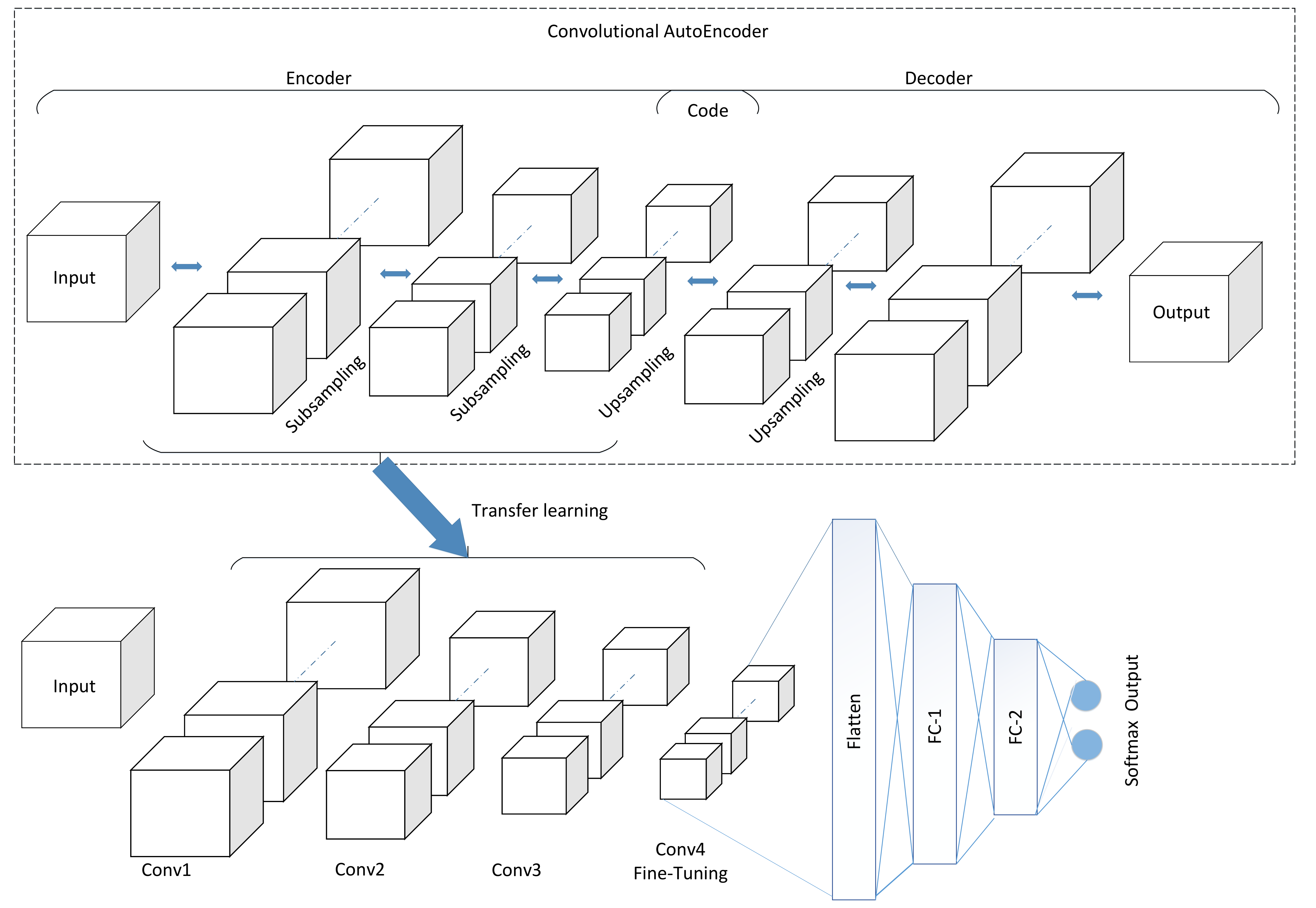}
		\caption[Transfer learning from convolutional autoencoder. ]{Transfer learning from convolutional autoencoder. }
		\label{fig:fig4_9}
	\end{figure}
	
\noindent	
Considering the fact that, lower level features such as edges, intensity changes are shared among all images, here the CAE was trained using multiple small patches of the available MRI scans to increase the size of training data. Each MRI scan was divided into four parts along the coronal slice that resulted each part a \((64\times 144\times 64)\) volume (Figure \ref{fig:fig4_8}). After dividing total number of training sample became \(5,132\) \((4\times1283)\) and after augmentation the number became \(30,792\). The Conv layers used in the CAE had 5 filters of size \((3\times3\times3)\) and selection of pool was \((2\times2\times2)\). Using the learned features from unsupervised learning a convnet was constructed for supervised learning. The first three conv layers were transferred from abovementioned CAE and a fourth Conv layer was added for fine-tuning the classification. After forth Conv block, 2 fully-connected layers with ReLU and a classification layer with Softmax were added. Figure \ref{fig:fig4_9} shows the network architecture with CAE.

	\section{Training Environment and Tools}\label{chapter:4.5}
All the implementations of this study were conducted on Google Colaboratoy\footnote{\url{https://colab.research.google.com/}} (\textit{colab}) which is a cloud hosted environment for running Python codes. At the moment, \textit{colab} offers 12 GB of free GPU access with maximum 25 GB of memory (RAM) per user. The choice of deep learning framework was \textit{Keras} where \textit{Tensorflow} was used as backend. \textit{Keras} offers very consistent and simple yet high level \textit{APIs} for deep learning and it is the most used framework. For converting 3D CAD mesh into binary volumetric data, a freely available command line program called \textit{binvox} \cite{patrickmin} was utilized. For MRI scans, an open source package \textit{Nibable} was used to read and write operation. To visualize the features leaned by the ConvNet \textit{Activation Maximization} technique was used which can be implemented using \textit{Keras-vis}. It is a high-level visualization toolkit for debugging neural net models build on \textit{Keras} framework. Relevance maps (LRPs) were extracted using \textit{iNNvestigate} visualization toolkit. Although there were several visualization methods implemented in \textit{iNNvestigate} module, for this study only LRP-\(\alpha_1\beta_0\) method was adjusted. For data preparation \textit{Scikit-learn}, an open source tool for data analysis was used as well. Other necessary library such as \textit{numpy} for array manipulation, \textit{matplotlib} for data visualization, etc. were also used. All the toolkit described above and used for the study are referenced along with their version in appendix section (See Appendix \ref{appendix:A}).

	\chapter{Results and Discussion }\label{chapter:5}
In this chapter the results of the various implementations are presented and discussed with the reference to the aim of the thesis, which is to learn the shape features and abstractions in 3D ConvNets. Different implementation strategies (already discussed in Chapter \ref{chapter:4}) were used to produce the respective results and methods that are described in chapter \ref{chapter:3} were utilized to interpret the learnings of a 3D ConvNet. At first, the outcomes of 3D ConvNets trained on 3D CAD objects (ModelNet and fruits CAD models) are shown. Next, models trained with 3D MRI scans are discussed with their outcomes. Finally, the results are compared to understand the learnings of the networks.

	\section{ModelNet and 3D Fruits }\label{chapter:5.1}	
\begin{figure}[H]
    	\centering
		\includegraphics[width=\textwidth]{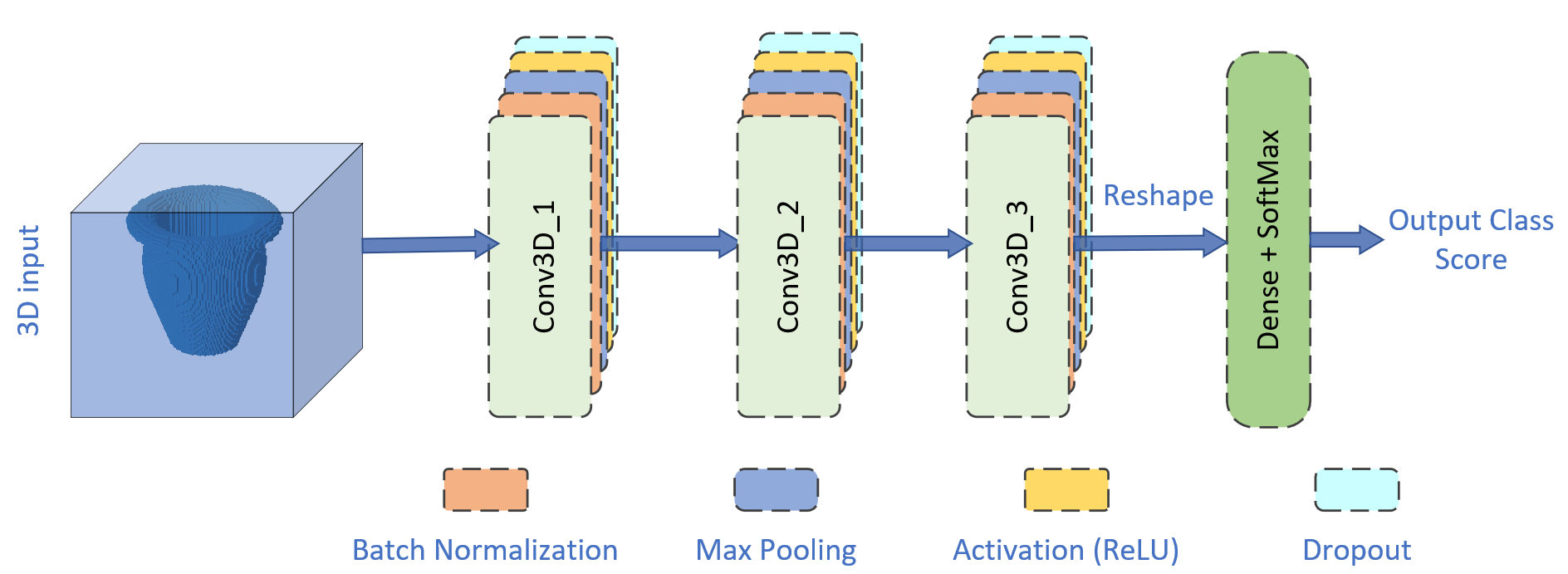}
		\caption[Simplified 3D ConvNet architecture for simulated CAD model classification. ]{Simplified 3D ConvNet architecture for simulated CAD model classification. }
		\label{fig:fig5_1}
	\end{figure}

To investigate the learning of 3D ConvNets on simple simulated 3D datasets, two classifiers with same architecture were trained. One with the ModelNet dataset that had three classes: Bottle, Lamp and Vase, another with fruits CAD models also with three classes: Apple, Banana and Pear. The 3D ConvNet was comprised of 3 Conv blocks and a FC layer with Softmax activation for classification. Each Conv block had a 3D Conv layer followed by  batch normalization, 3D maxpooling, activation(ReLU) and dropout layer (dropout rate 0.3). There were 5 kernels of size \((3\times3\times3)\) for each Conv layer with stride 1 and maxpool used pool size \((2\times2\times2)\). The inputs were binary 3D voxel grid of dimension of \((64\times64\times64)\) which were convolved with kernels in Conv layer and produces a series of volumetric feature maps. A simplified version of the network architecture is illustrated in figure \ref{fig:fig5_1}. Max-pooling operation reduced the size of the intermediate representation of the input. After final Conv block the \(64^3\) input became \(8^3\) voxel grid that was flattened to connect with FC layer with 3 classification neurons. Output volume of a layer depends on the number of filters used. Final Conv layer used 5 filters that produced \(5\times8^3\) feature volume after max-pooling. Table \ref{table:table5_1} shows the network demographic for Conv and max-pooling layers. 
	
	\begin{table}[H]
	\centering
	\begin{tabular}{||l|c|c|c|c|c|c||}
	\hline
		 & conv-1 & maxpool-1 & conv-2 & maxpool-2 & conv-3 & maxpool-3  \\ \hline\hline
	Filter \(Size^3\) & 3 & 2 & 3 & 2 & 3 & 2 \\ \hline
	No. of Filters & 5 & - & 5 & - & 5 & - \\ \hline
	{\vtop{\hbox{\strut Layer Input} \hbox{\strut \([voxels]^3\)}}} & 64 & 64 & 32 & 32 & 16 & 16 \\ \hline
	{\vtop{\hbox{\strut Layer Output} \hbox{\strut \([voxels]^3\)}}} & 64 & 32 & 32 & 16 & 16 & 8 \\ \hline
	{\vtop{\hbox{\strut Activation map} \hbox{\strut \(N\times[voxels]^3\)}}}  & \(5\times64\) & \(5\times32\) & \(5\times32\) & \(5\times16\) & \(5\times16\) & \(5\times8\) \\ \hline
	 	
 	\end{tabular} 	
 	
	\caption[3D ConvNet architecture details.]{3D ConvNet architecture details. \(^3\) denotes the dimension.}
  	\label{table:table5_1}
	\end{table}

\begin{figure}[H]
    	\centering
		\includegraphics[width=\textwidth]{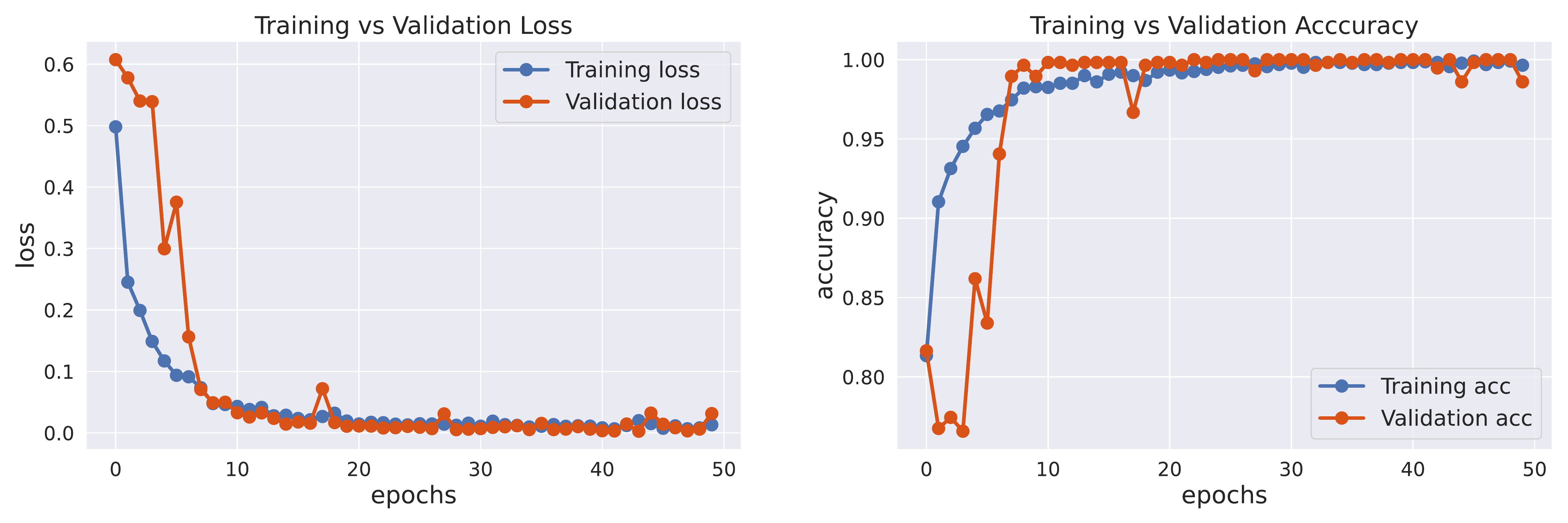}
		\caption[Training log of ModelNet classification. Train vs Validation Loss (left) and Accuracy (right).]{Training log of ModelNet classification. Train vs Validation Loss (left) and Accuracy (right).}
		\label{fig:fig5_2}
	\end{figure}

\begin{figure}[H]
    	\centering
		\includegraphics[width=\textwidth]{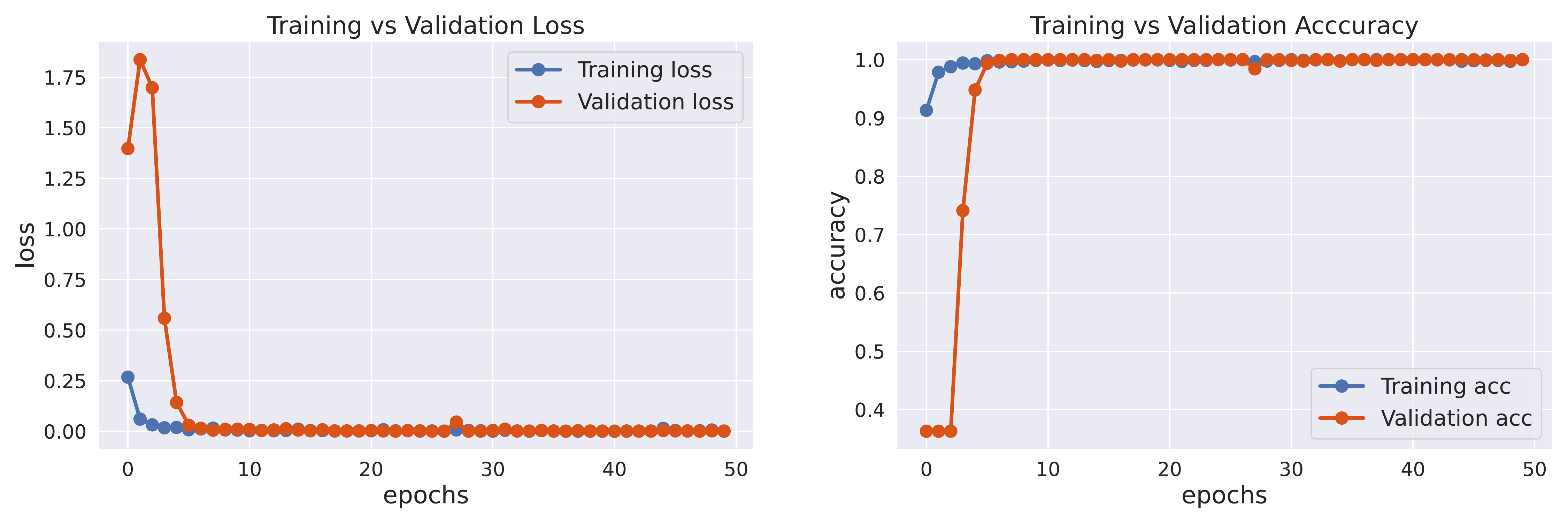}
		\caption[Training log of Fruits 3D CAD classification. Train vs Validation Loss (left) and Accuracy (right).]{Training log of Fruits 3D CAD classification. Train vs Validation Loss (left) and Accuracy (right).}
		\label{fig:fig5_3}
	\end{figure}

\noindent	
After data preparation using preprocessing steps and augmentation (described in 4.2), the network was trained for 50 epochs using batch size of 32 and learning rate 0.01. Stochastic Gradient Descent algorithm Adam was used for optimization where Categorical Cross-Entropy was the choice of loss function. 20\% of the dataset was used for validation set without any cross-validation. Figure \ref{fig:fig5_2} and \ref{fig:fig5_3} shows the loss and accuracy graph on training vs validation set for ModelNet and 3D fruits classifier respectively. For ModelNet classifier the network converged after 10 epochs with accuracy 98\%. Almost same result was obtained for 3D fruits classifier as well.

	\subsection{Filters and Feature maps }\label{chapter:5.1.1}	
\begin{figure}[H]
    	\centering
		\includegraphics[width=\textwidth]{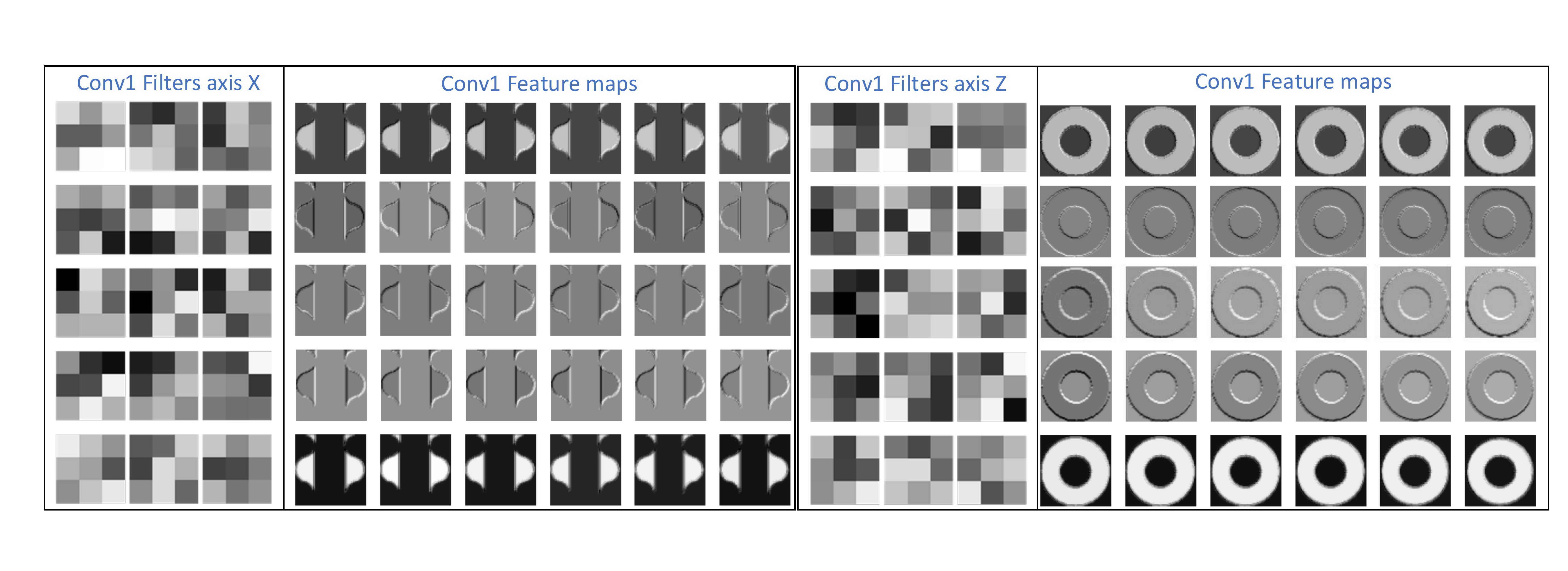}
		\caption[Conv1 filters and feature maps of ModelNet classifier. Feature maps generated using a vase CAD model from test set (shown in figure \ref{fig:fig5_5}).]{Conv1 filters and feature maps of ModelNet classifier. Feature maps generated using a vase CAD model from test set (shown in figure \ref{fig:fig5_5}).}
		\label{fig:fig5_4}
	\end{figure}

\begin{figure}[H]
    	\centering
		\includegraphics[width=\textwidth]{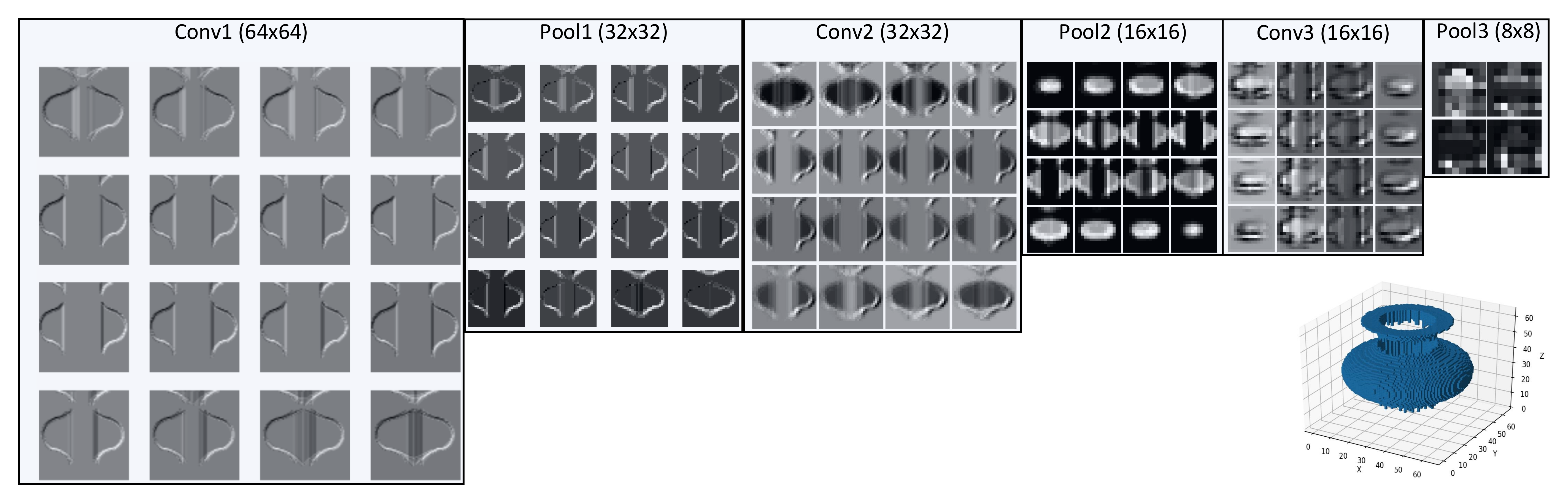}
		\caption[Activation maps for a test model vase through the network.]{Activation maps for a test model vase through the network.}
		\label{fig:fig5_5}
	\end{figure}

Visualizing learned filters is a common strategy to interpret the result of a Conv layer. Usually Conv filters of first layer are the most interpretable cause they directly examine the input pixels.  Figure \ref{fig:fig5_4} displays extracted filters and feature maps from Conv layer 1 of ModelNet classifier. Each 2D \((3\times3)\) slice of \((3^3\times1\times5)\) filter is plotted where along \textit{X} axis means \((x\times3^2\times1\times row)\) and along \textit{Z} axis means \((3^2\times z\times1\times row)\). Conv1 feature map was a volume of \((64^3\times5)\). In figure \ref{fig:fig5_4} the \(3^{rd}\) volume with \(64^2\) 2D slices are demonstrated which were generated by a Vase sample (shown in figure \ref{fig:fig5_5}) from test set. 
	
\begin{figure}[H]
    	\centering
		\includegraphics[width=\textwidth]{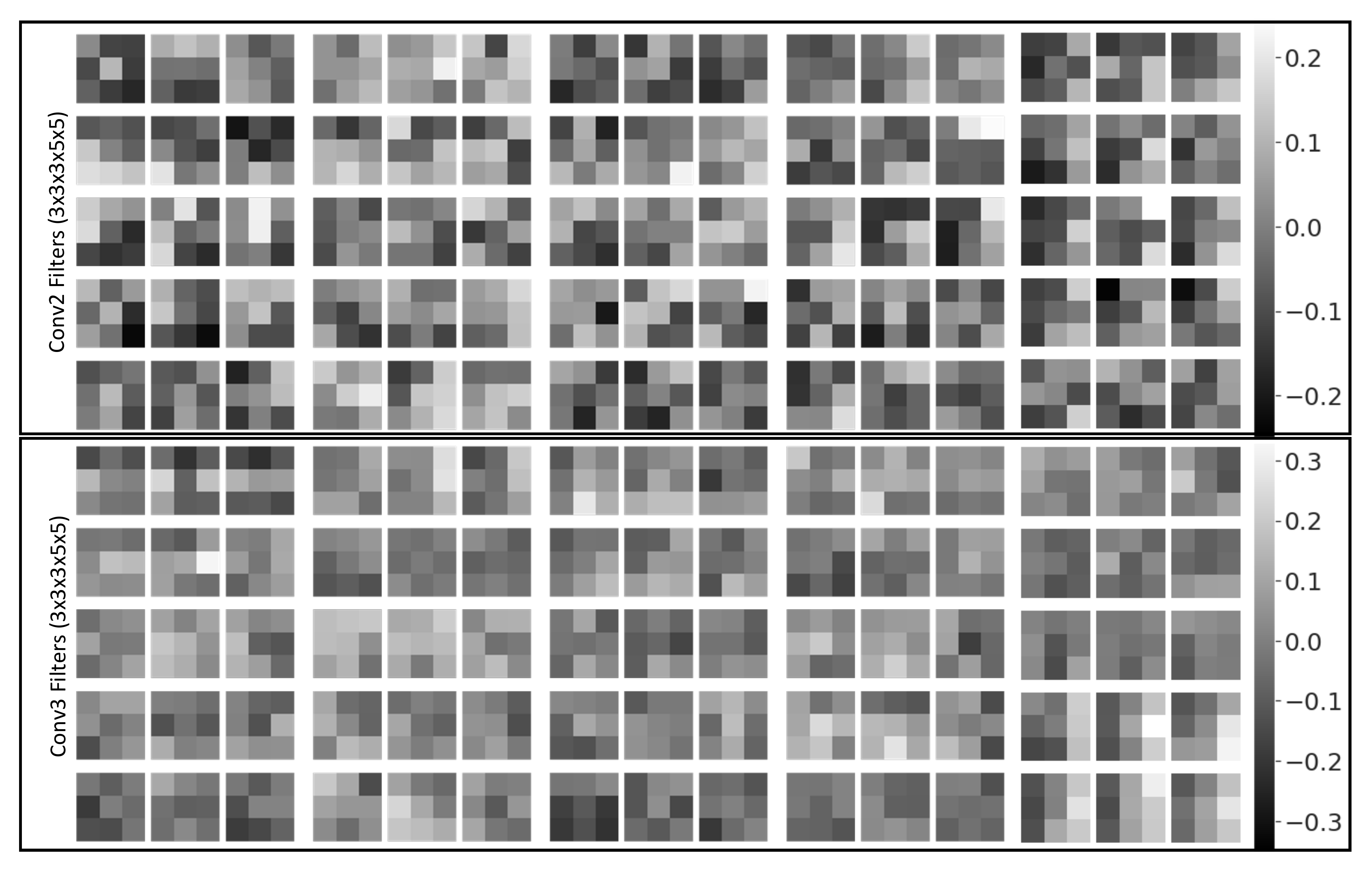}
		\caption[Learned filters by the ModelNet classifier.]{Learned filters by the ModelNet classifier.}
		\label{fig:fig5_6}
	\end{figure}

\noindent	
In figure \ref{fig:fig5_5} activation maps for all three Conv blocks are shown for a test sample Vase which was predicted with 100\% accuracy by the classifier. For the ease of visualization only middle volume is presented with inner most slices. See table \ref{table:table5_1} for activation volume details for each layer. In deeper layers the features are sparser and more localized, visualizing them could help us to identify potential dead filters for which the activation might be zero. Figure \ref{fig:fig5_6} plots all the learned filters by Conv1 and Conv2 layer. Typically, Smooth pattern in learned filters is the indication of a well-trained network. In this case, looking at the filters it was hard to find smooth patterns without noise. However, in deeper layers the patterns became smoother and last couple of filters (figure \ref{fig:fig5_6} lower right) in Conv3 layer, patterns of potential edge detector filters were found. It is also important to mention that these are only the slices of 3D filters which is very hard to interpret. Nevertheless, looking at the activation maps revealed how the input was transformed through the network layers using the learned filters. 
	
\noindent	
Results from the 3D fruits CAD classifier were comparable to ModelNet thus omitted here. The amount of training samples was insignificant to have a smooth pattern in the learned filters. On top of that, most of the CAD models had very less occupancy in the volumetric grid and were hollow. Although both classifiers yielded a very high accuracy with area under the curve (AUC) greater than 90\%, maybe networks were not learning the features that were hoped for to make the classification decision.

	\subsection{LRP Visualization }\label{chapter:5.1.2}	
As explained in chapter \ref{chapter:3}, LRP calculates the relevance of each neuron in the network to make the decision. LRP relevance map visualization can give a better idea to understand which pixels/voxels of the input image are getting more activated for a class prediction. For this experiment, LRP-\(\alpha{_1} \beta{_0}\) method was used to produce the relevance map for different class neurons with different input test models.  

	\begin{figure}[H]
    	\centering\offinterlineskip
		\includegraphics[width=0.6\textwidth]{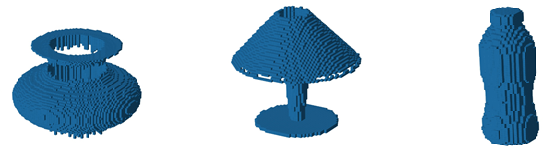}
		\includegraphics[width=\textwidth]{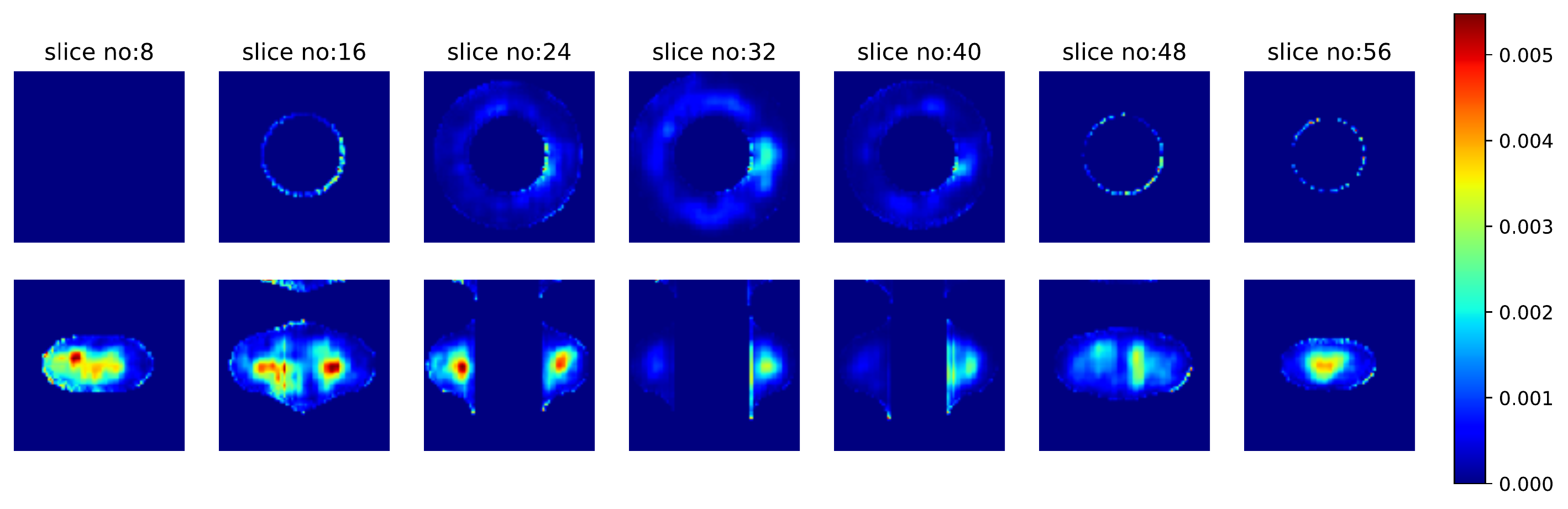}
		\includegraphics[width=\textwidth]{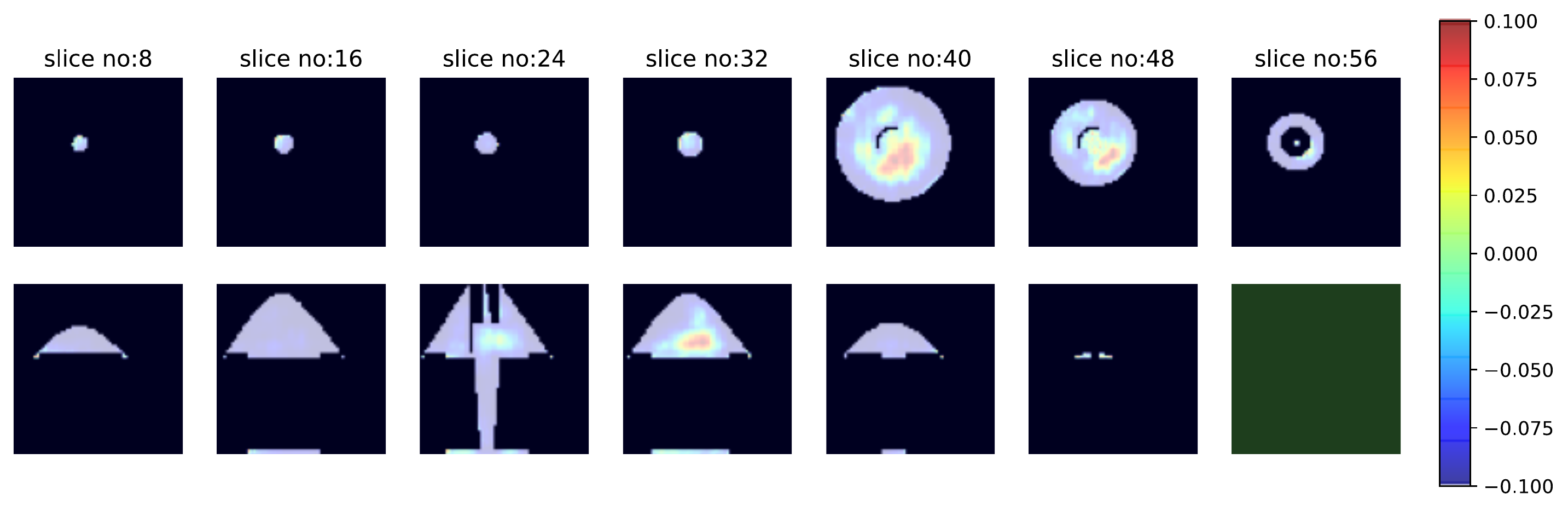}
		\includegraphics[width=\textwidth]{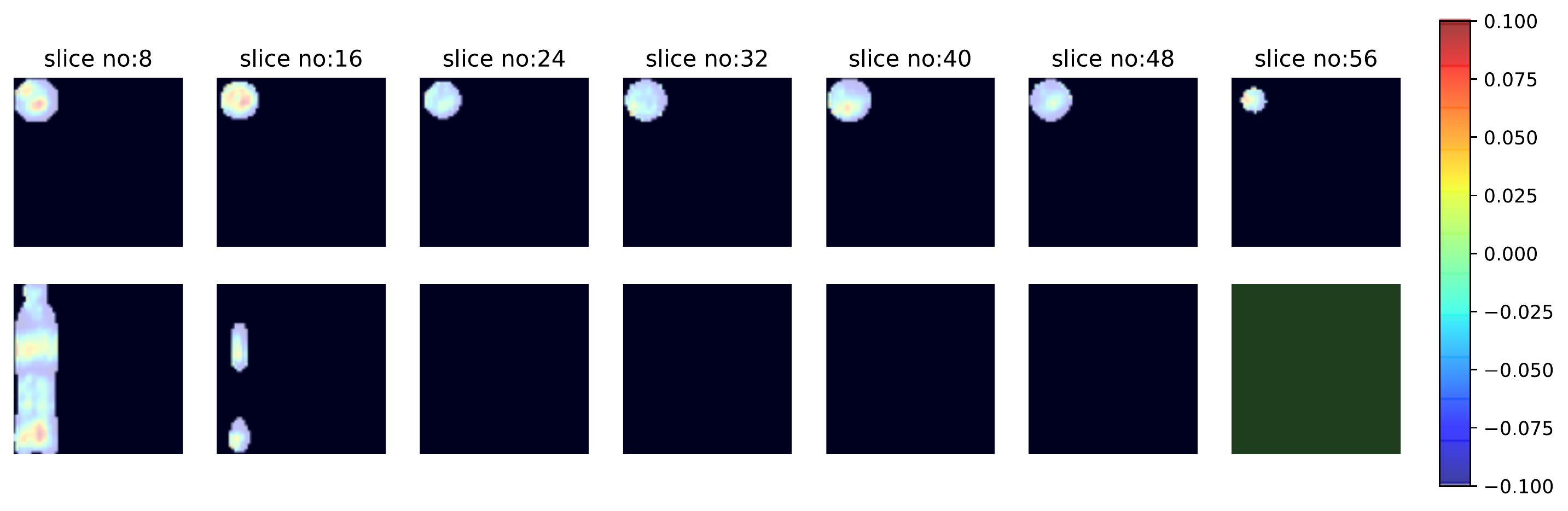}
		\caption[LRP maps for three classes (Vase, Lamp and Bottle) of ModelNet classifier.]{LRP maps for three classes of ModelNet classifier. Top rows: vase, center: lamp, bottom: bottle class.}
		\label{fig:fig5_7}
	\end{figure}
	
\noindent	
LRP maps for each class was investigated with couple of models from test set. Figure \ref{fig:fig5_7} illustrates LRP maps for each class of ModelNet. Only 7 slices from top and side view have been plotted for each volume of LRP map. From the relevance maps it appeared that the model was focusing on arbitrary parts of the object which were not intuitively discriminating between the object. It was largely unknown which features actually drove the decision. It can also be the case that the properties that LRP maps were pointing at are not directly visible such as thickness or other shape features. Results from fruits model were also same and therefore omitted here. As simulated 3D data that were used for the classification was very small and most of the object covered only a small part of the volume, it is most likely that the network was making its decision based on the property that were not relevant. LRP maps from 3D fruits model were also comparable to these results.  	
	
	\section{3D ConvNets with MRI Scans  }\label{chapter:5.2 }
In this section, 3D ConvNets trained with MRI brain scan are discussed. Available MRI scans from ADNI and AIBL datasets were preprocessed and then used to train the network to classify between AD and NC subjects. There were several models with different structure which were trained to extract the shape features learned by the model. With respect to the changes in the network structure such as number of layers, size and shape of the filters, the learning abstractions of the trained model were investigated. 

	\subsection{Training of 3D CNN  }\label{chapter:5.2.1 }
Keeping the thesis goal in mind each model was trained with the same set of train, validation and test set. Although for better accuracy evaluation \textit{k-fold} cross validation could have been used, due to limited computational resource and time it was not considered. Moreover, this study aimed to find learning abstraction based on changes in network layer structure. Table \ref{table:table5_2} shows different network architecture along with their accuracy.  

	\begin{table}[H]
	\centering
		\begin{tabular}{||l|c|c|c|c|c||}
		\hline
	
		 & {\vtop{\hbox{\strut Architecture} \hbox{\strut of}\hbox{\strut Conv block}}} & {\vtop{\hbox{\strut No. of} \hbox{\strut Conv}\hbox{\strut block}}} &  {\vtop{\hbox{\strut No. \& size} \hbox{\strut of filter}\hbox{\strut per block}}} & {\vtop{\hbox{\strut No. and size} \hbox{\strut of}\hbox{\strut Dense layer(s)}}} & Acc.  \\ \hline \hline
		 
Model-1	& {\vtop{\hbox{\strut Conv3D, B.Norm,} \hbox{\strut MaxPool, ReLU}\hbox{\strut DP(0.3)}}} & 3 &  \(5(3\times3\times3)\) & Dense(2) & 71.32\%  \\ \hline		 

Model-2	& {\vtop{\hbox{\strut Conv3D, B.Norm,} \hbox{\strut MaxPool, ReLU}\hbox{\strut DP(0.3)}}} & 4 &  \(5(3\times3\times3)\) & Dense(2) & 72.09\%  \\ \hline

Model-3	& {\vtop{\hbox{\strut Conv3D, B.Norm,} \hbox{\strut MaxPool, ReLU}\hbox{\strut DP(0.3)}}} & 4 &  \(5(5\times5\times5)\) & Dense(2) & 62.02\%  \\ \hline

Model-4	& {\vtop{\hbox{\strut Conv3D, B.Norm,} \hbox{\strut MaxPool, ReLU}\hbox{\strut DP(0.3)}}} & 4 &  \(5(7\times7\times7)\) & Dense(2) & 72.09\%  \\ \hline

Model-5	& {\vtop{\hbox{\strut Conv3D+ReLU,} \hbox{\strut MaxPool, B.Norm}}} & 4 &  \(5(7\times7\times7)\) & Dense(2) & \textbf{76.74\%}  \\ \hline	

Model-6	& {\vtop{\hbox{\strut Conv3D+ReLU,} \hbox{\strut MaxPool, B.Norm}}} & 4 &  \(5(3\times3\times3)\) & {\vtop{\hbox{\strut Dense(64)+DP(0.1)} \hbox{\strut Dense(32)+DP(0.1)}\hbox{\strut Dense(2)}}} & 65.89\%  \\ \hline
	 
 		\end{tabular} 	
 	
	\caption[Different 3D ConvNet architectures. Training setup: epochs=50, batch size=32, learning rate=0.01, loss function=\textit{categorical cross entropy}, optimizer=\textit{Adam}, Average training time(approximately)=68min(without augmentation) \& 144min(with augmentation).]{Different 3D ConvNet architectures. Training setup: epochs=50, batch size=32, learning rate=0.01, loss function=\textit{categorical cross entropy}, optimizer=\textit{Adam}, Average training time(approximately)=68min(without augmentation) \& 144min(with augmentation). [DP denotes dropout rate in the table]}
  	\label{table:table5_2}
	\end{table}
	
	\pdfbookmark[3]{Evaluation of Model Structure}{evaluation of model structure}
	\subsubsection{Evaluation of Model Structure}\label{chapter:5.2.1.1}	
\begin{figure}[H]
    	\centering\offinterlineskip
		\includegraphics[width=0.8\textwidth]{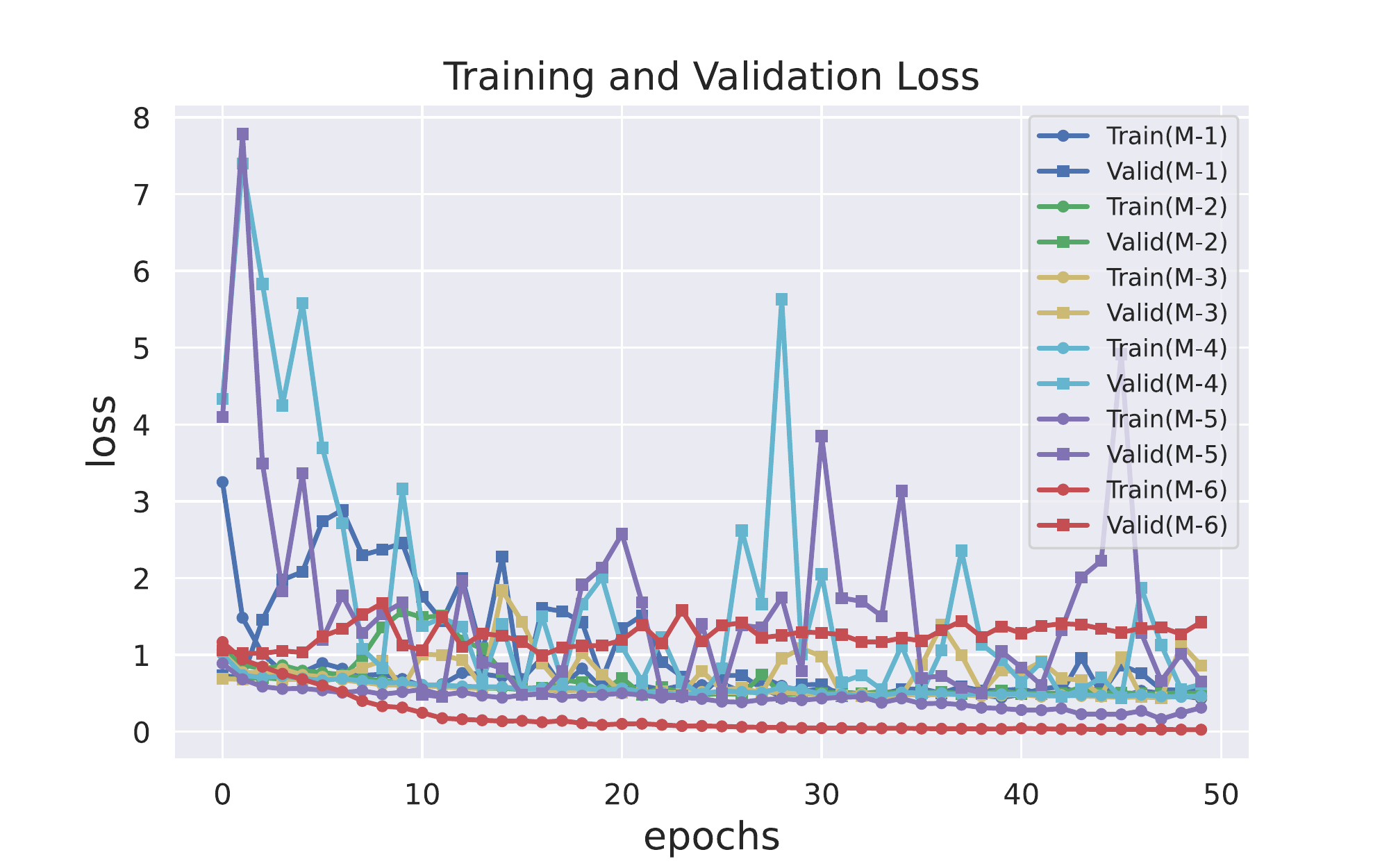}
		\includegraphics[width=0.8\textwidth]{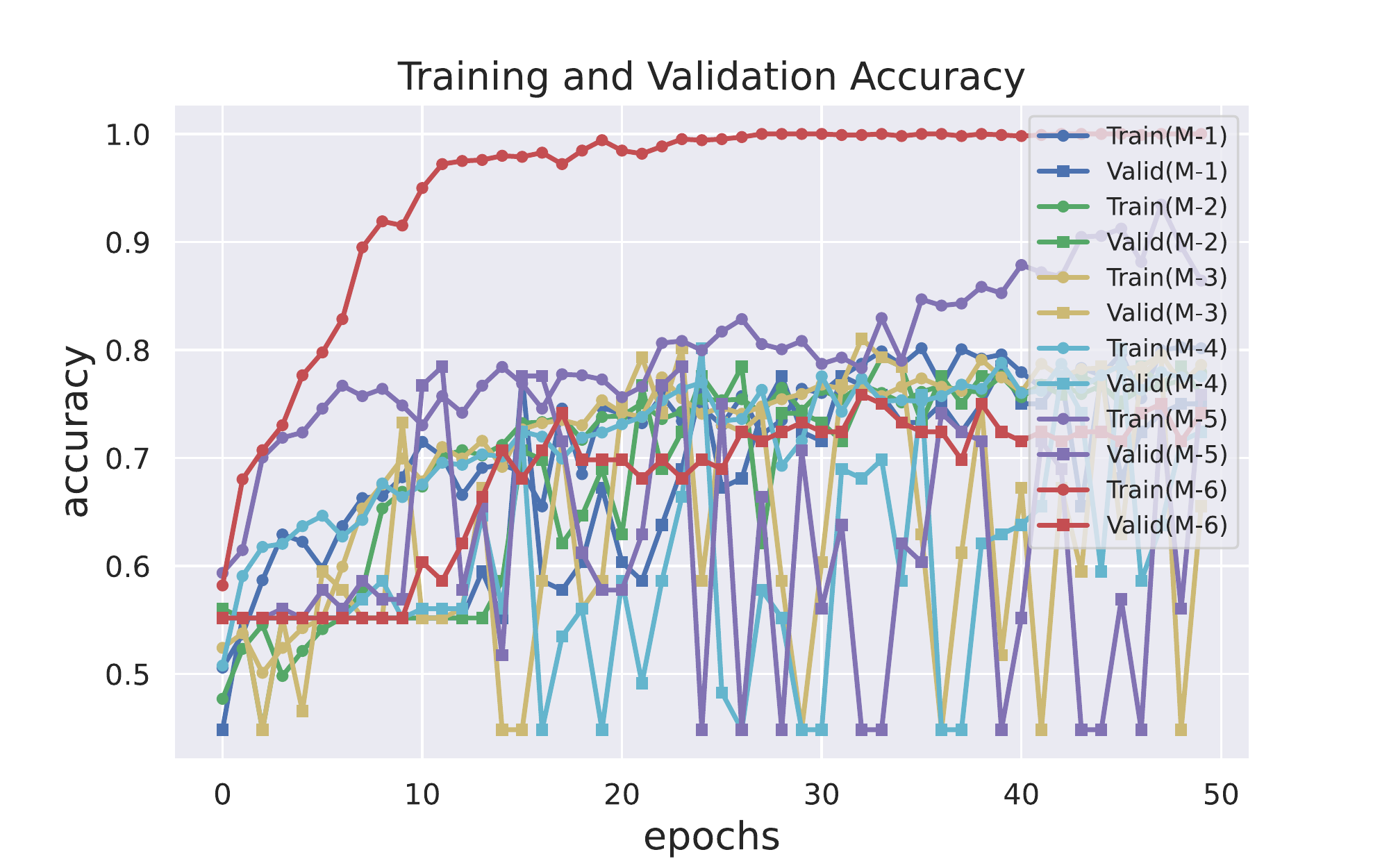}
		\caption[Mean loss and accuracy curve for Modes-1 to Model-6(Table \ref{table:table5_2}).] {Loss and Accuracy curve for Model-1 to Model-6(Table \ref{table:table5_2}). [M denotes Model in the figure] }
		\label{fig:fig5_8}
	\end{figure}

The motivation was to find a base network structure that has reasonable accuracy and somewhat stable loss and accuracy curve. First the same architecture (Model-1) that was used for CAD models (described above) was considered but as brain scans had larger volume which resulted higher trainable parameters in dense layer after \(3^{rd}\) Conv block. Adding a fourth Conv block (Model-2) reduced the trainable parameters with a little increase in accuracy as well. Increasing the filter size to \(5^3\) (Model-3) reduced the accuracy but \(7^3\) filter size (Model-4) yielded same accuracy like Model-2. Although Model-4 performed same as model-2, LRP maps (see figure \ref{fig:fig5_16}) suggest that it was looking for irrelevant part of the brain scan for the classification. By changing Conv block structure in Model-5 but keeping the same filter number and size \((5\times7^3)\) produced the best accuracy out of all models in table \ref{table:table5_2}. However, all the models had a very unstable loss and accuracy curve. Figure \ref{fig:fig5_8} plots the loss and accuracy curve for Model-1 to Model-6. All the models were trained using data without augmentation.

\noindent	
From figure \ref{fig:fig5_8} it appeared that all the models training log was very unstable except model-6 which resulted a bit stable training, although it faced overfitting problem like all other models and had a very low accuracy (see table \ref{table:table5_2} for details). This base model was used to investigate learning of the network by training with augmented dataset  while validation and test set remained same.  

	\pdfbookmark[3]{Evaluation of Data Augmentation}{evaluation of data augmentation}
	\subsubsection{Evaluation of Data Augmentation}\label{chapter:5.2.1.2}
To improve learning of a network a large training dataset is very crucial. As large dataset was not available for this study, with some data augmentation size of the training data was increased. Simple 10 voxel shifting along each axis before and after left/right flip was the augmentation scheme for all the models. After augmentation, training size became 8,304 samples where validation and test set were excluded. Number of samples in validation set (126) and test set (129) remained unchanged for all the training process.

	\begin{figure}[H]
    	\centering\offinterlineskip
		\includegraphics[width=0.8\textwidth]{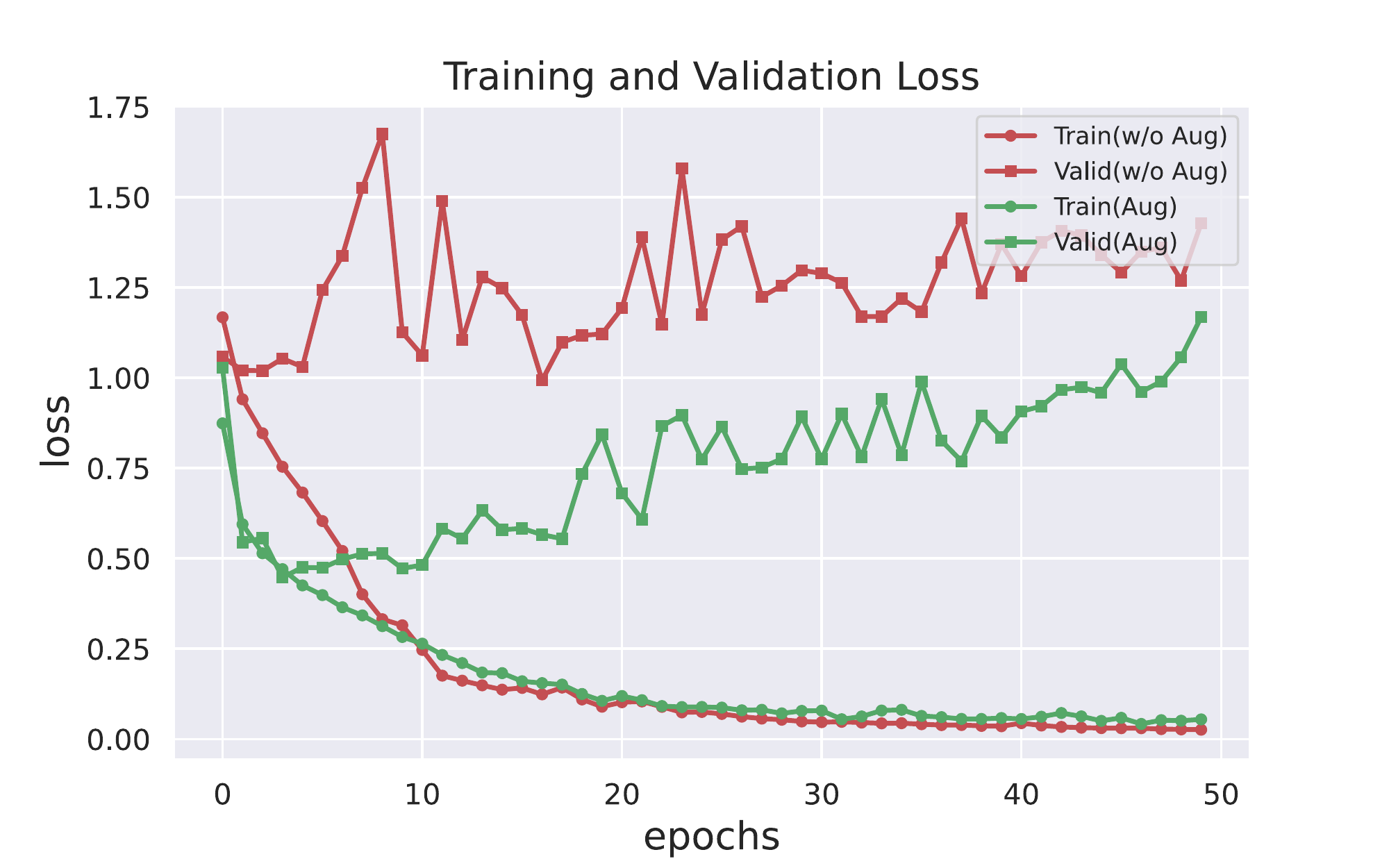}
		\includegraphics[width=0.8\textwidth]{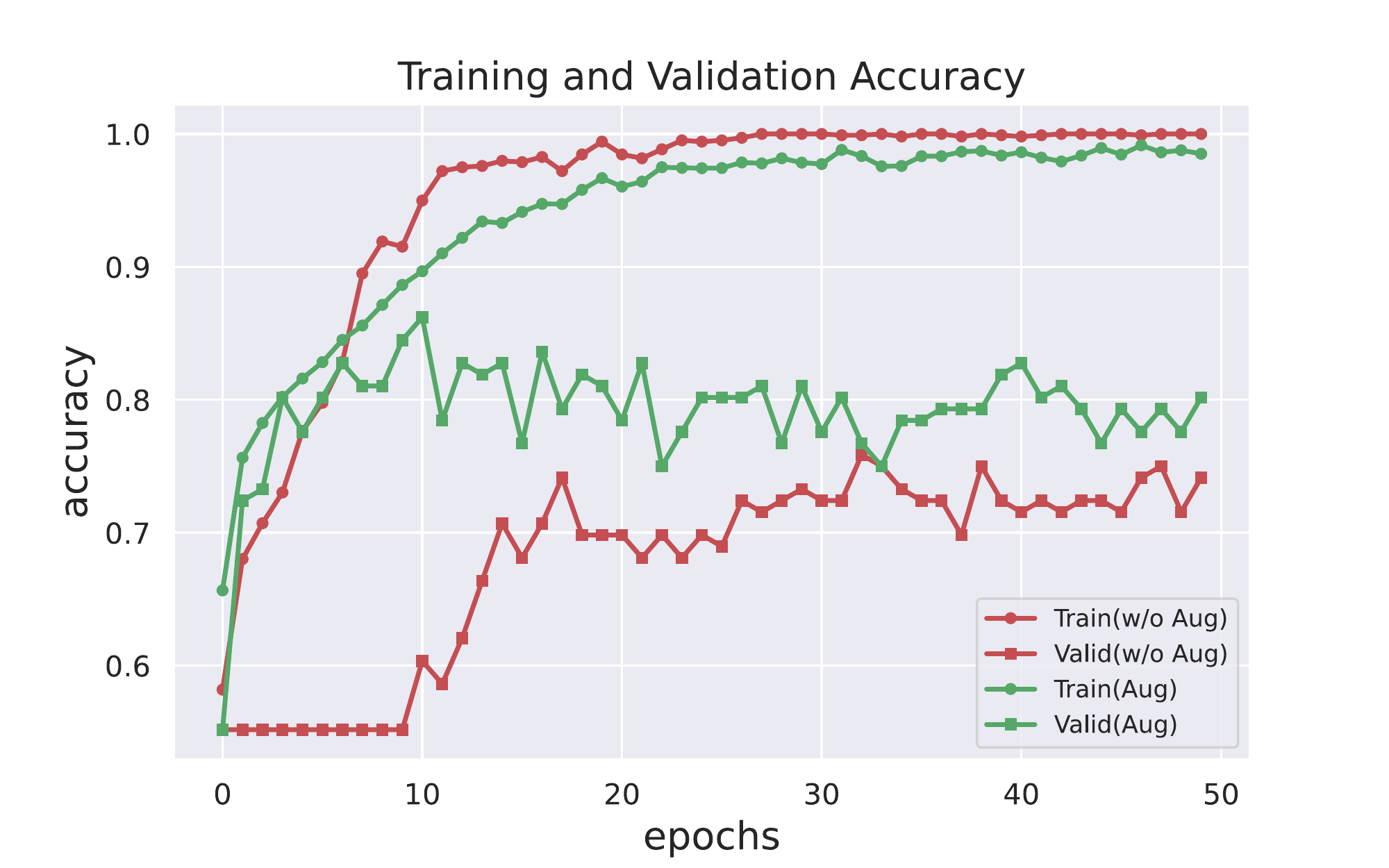}
		\caption[Loss and Accuracy curve for Model-6 (given in table \ref{table:table5_2}) trained with (green plot) and without (red plot) data augmentation.]{Loss and Accuracy curve for Model-6 (given in table \ref{table:table5_2}) trained with (green plot) and without (red plot) data augmentation.}
		\label{fig:fig5_9}
	\end{figure}

\noindent	
Training the same Model-6 architecture with augmented data increased the accuracy to 74.42\% from 65.89\%. Figure \ref{fig:fig5_9} shows the training log of Model-6 (green plots) with augmented dataset. Training the other models (Model-1 to Model-5) with augmented dataset showed similar pattern and therefore omitted here.   

	\pdfbookmark[3]{Evaluation of Conv Filter Shape}{evaluation of conv filter shape}
	\subsubsection{Evaluation of Conv Filter Shapes}\label{chapter:5.2.1.3}
 
	\begin{figure}[H]
    	\centering\offinterlineskip
		\includegraphics[width=0.8\textwidth]{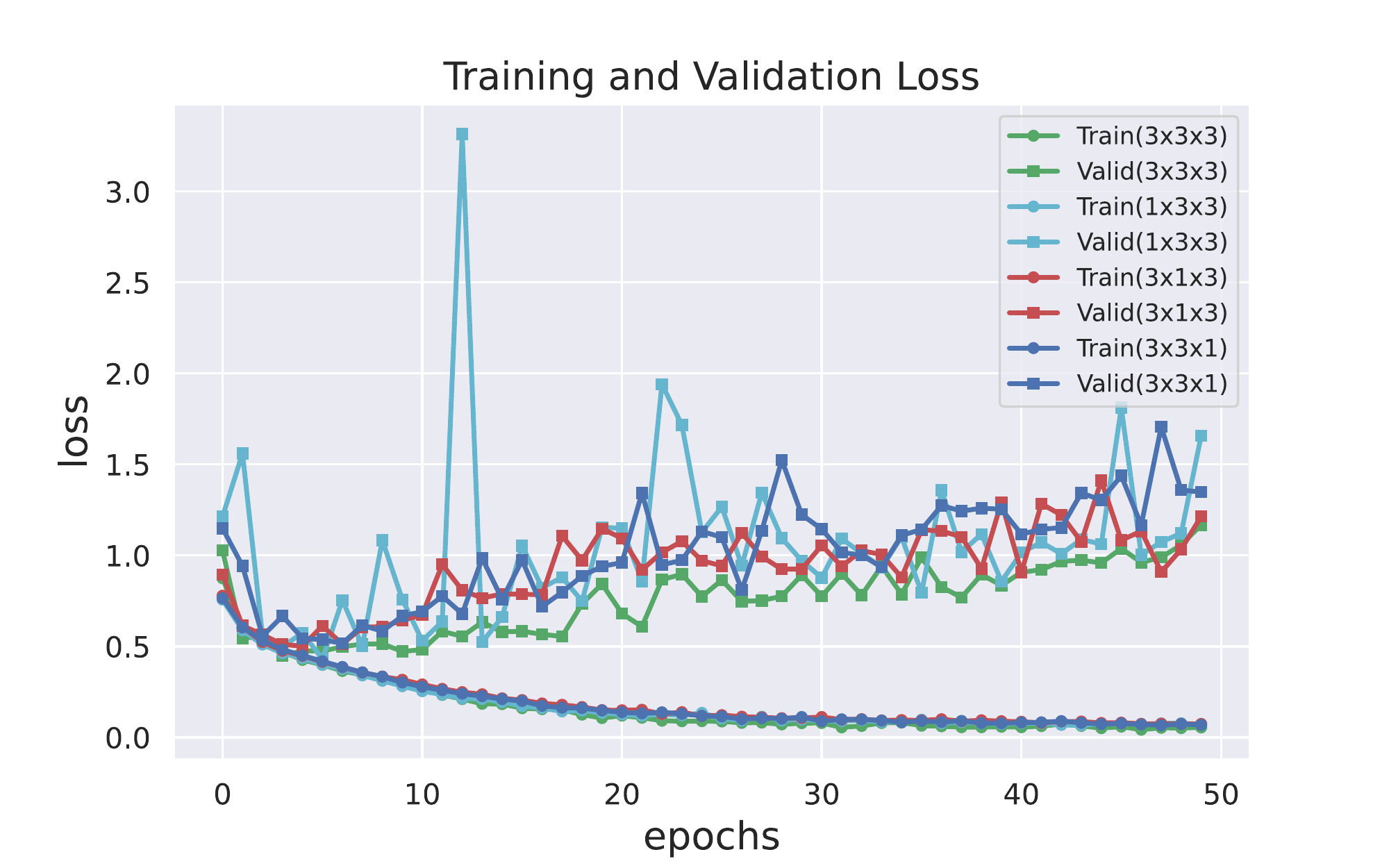}
		\includegraphics[width=0.8\textwidth]{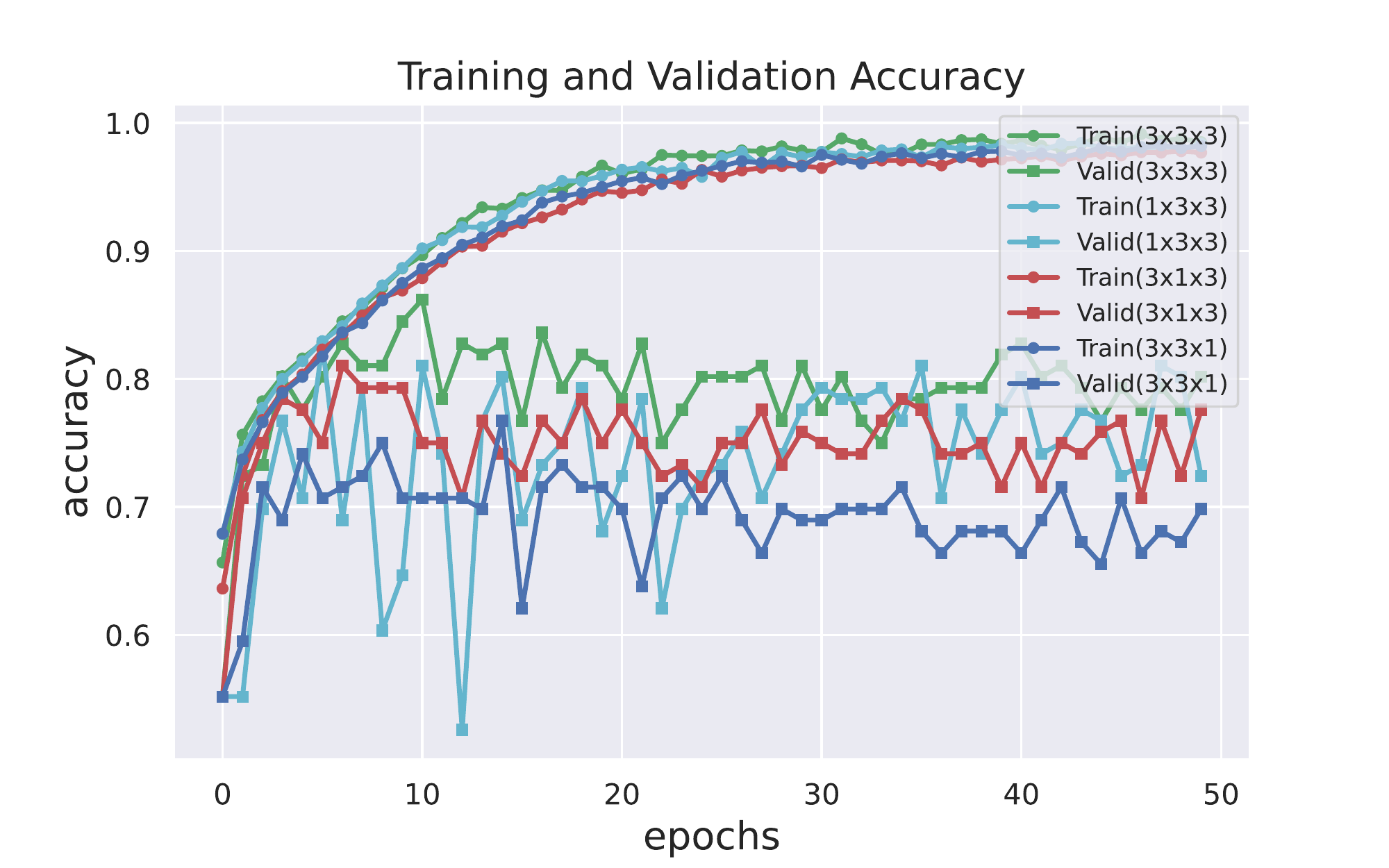}
		\caption[Training log of \(3^3\) filters model and other shaped filters model.]{Training log of \(3^3\) filters model and other shaped filters model.}
		\label{fig:fig5_10}
	\end{figure}

So far, all the models discussed  above have used cubic filters of different sizes but the shape remained same. To explore the effect of various shapes of the filter on 3D inputs, previously found best accuracy model was trained with different shapes of the filters. Focusing on the sagittal, coronal and axial slices of the brain MRI, filters of shape \((1\times3\times3)\), \((3\times1\times3)\) and \((3\times3\times1)\) ) were selected respectively. While accuracy of the model with coronal \((3\times1\times3)\) filter was comparable (75.19\%) with \(3^3\) filter model (Model-6), other two filters showed a significant decrease in accuracy (\((1\times3\times3)\)= 63.5\% and \((3\times3\times1)\)=68.99\%). In figure \ref{fig:fig5_10} training log of model with different shaped filters and cubic filters are illustrated and it appeared that they are similar . Learned features and visualization of relevance maps for these models are compared and discussed in next sections.

	\subsection{Feature maps Visualization}\label{chapter:5.2.2}
Feature maps are the result of an input image after convolutional operation by ConvNet filters in a layer. Multiple filters in a layer generate a feature map volume by stacking them together. A 3D input convolved with 3D filters produce a 4D feature map where \(4^{th}\) dimension is the number of filters used. The ConvNets used in this study had 5 filters in each Conv layer. To see how the input was transformed passing through the layers, feature maps of each Conv layer were visualized here.     

	\begin{figure}[H]
    	\centering
		\includegraphics[width=0.65\textwidth]{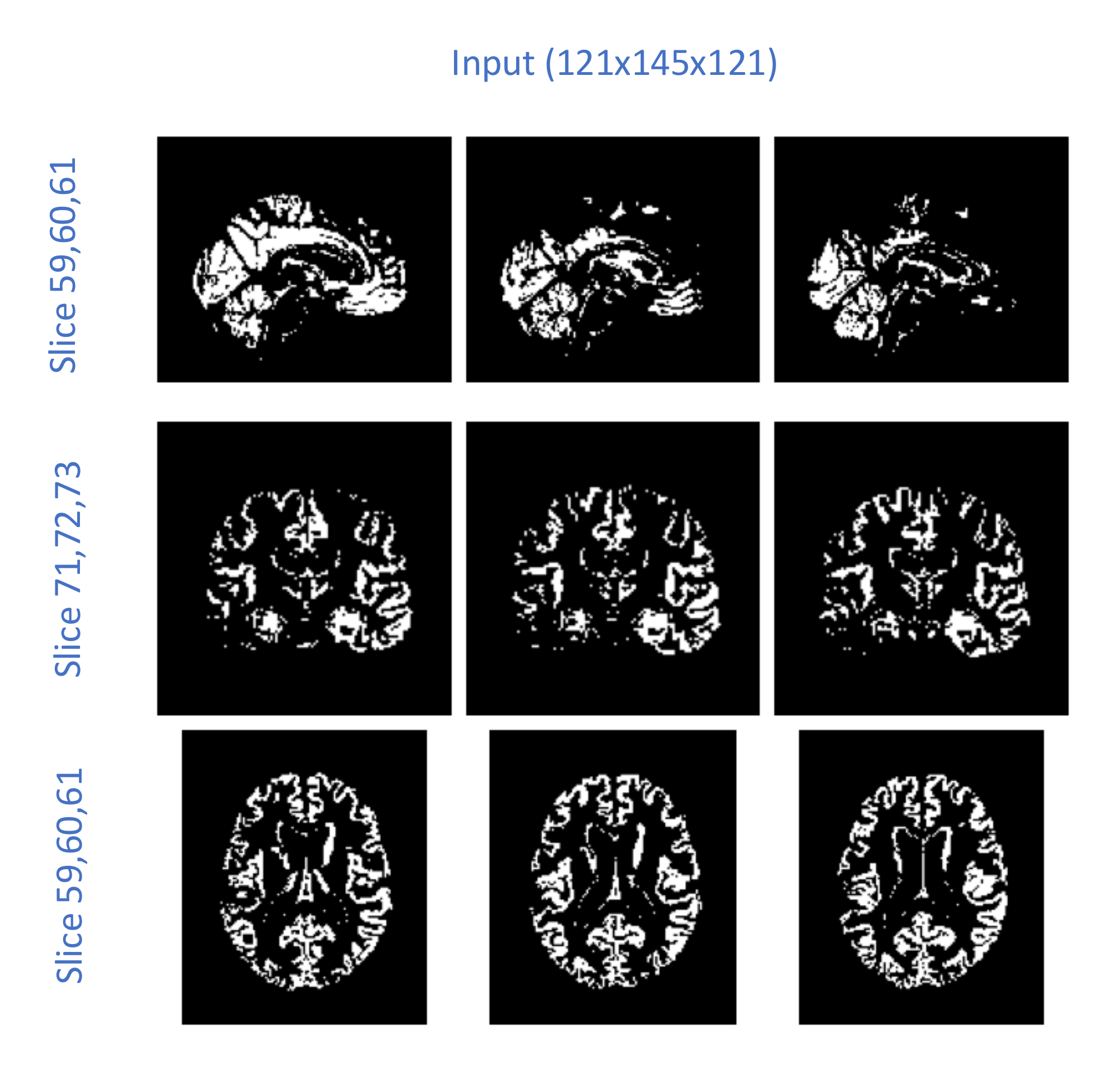}
		\caption[Inner most sagittal, coronal and axial (top to bottom) slices of MRI scan.]{Inner most sagittal, coronal and axial (top to bottom) slices of MRI scan. }
		\label{fig:fig5_11}
	\end{figure}

	\begin{figure}[H]
    	\centering
		\includegraphics[width=0.9\textwidth]{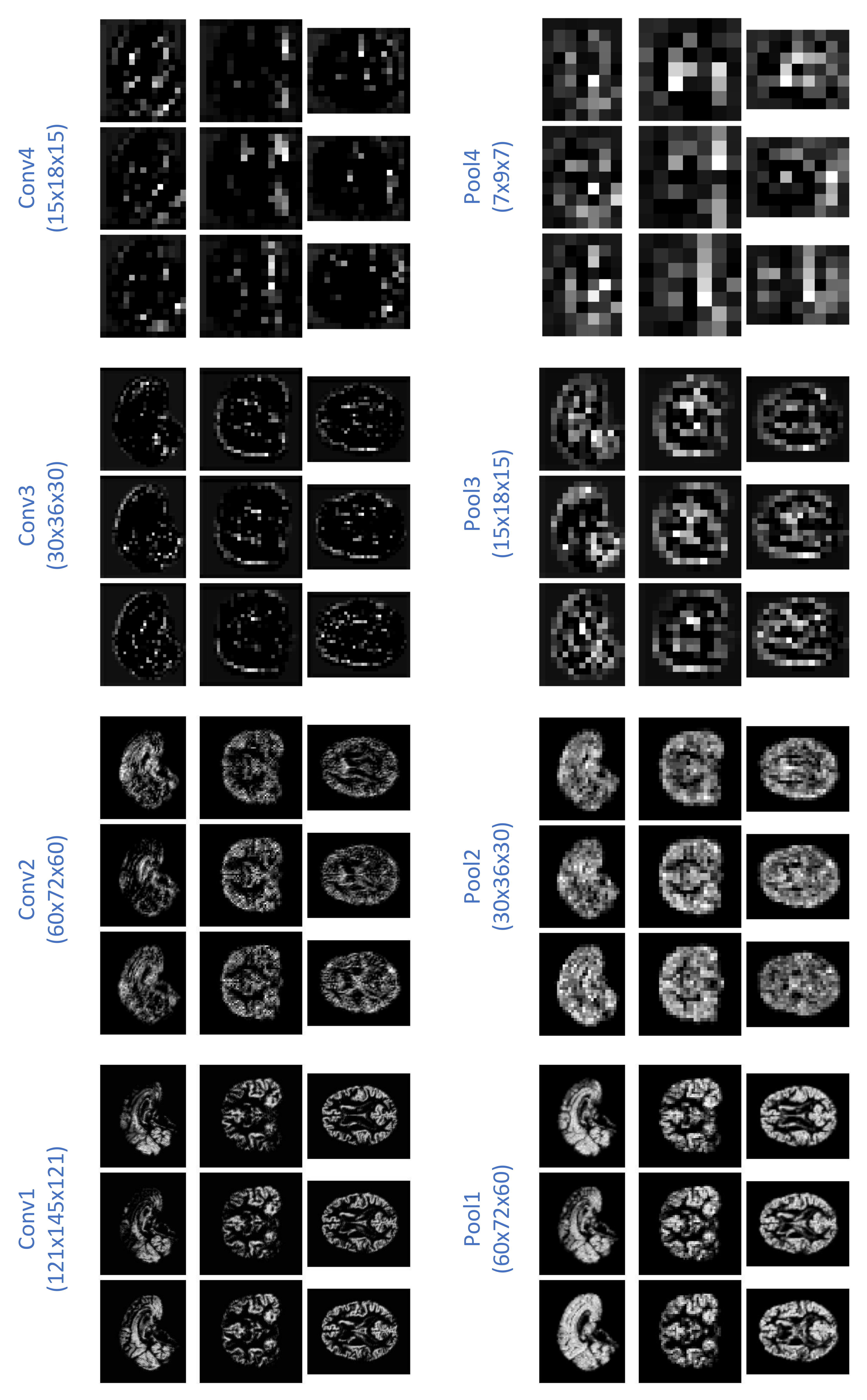}
		\caption[Feature maps of Conv and Pool layers in the network.]{Feature maps of Conv and pool layers in the network.}
		\label{fig:fig5_12}
	\end{figure}

\begin{figure}[H]
    	\centering
		\includegraphics[width=\textwidth]{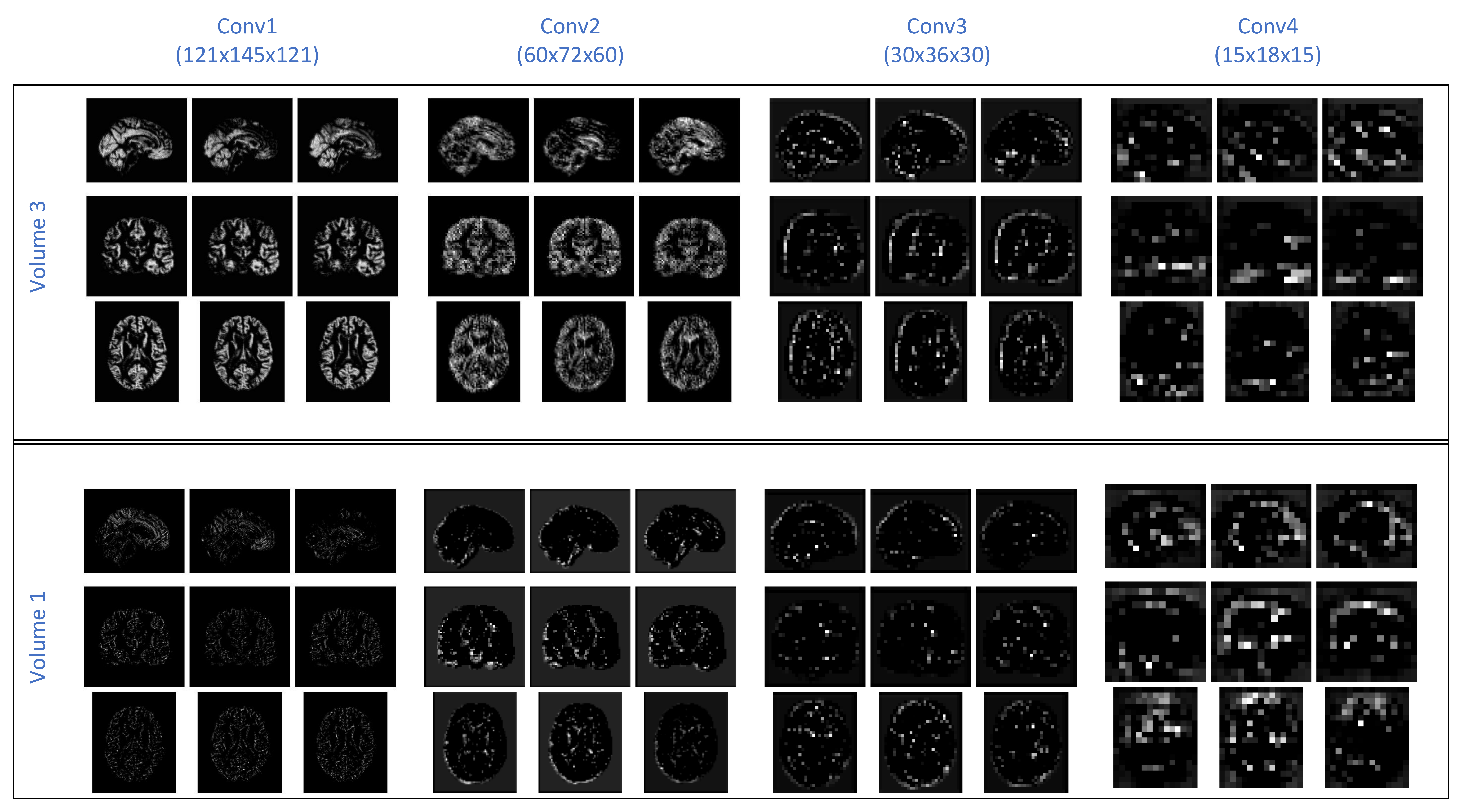}
		\caption[Feature maps of different filters in a layer output volume.]{Feature maps of different filters in a layer output volume.}
		\label{fig:fig5_13}
	\end{figure}

\noindent	
Three sequential slices from the middle of sagittal, coronal and axial MRI scan (figure \ref{fig:fig5_11}) were used to visualize the feature map transformation in the network layers. Max-pooling followed by a Conv layer reduced the input resolution by half. The selected ConvNet for feature map visualization had same network structure as Model-6 (Table \ref{table:table5_2}) trained with augmented dataset. See section \ref{chapter:5.2.1 } for more training details. Figure \ref{fig:fig5_12} \& \ref{fig:fig5_13} presents the feature maps of all four Conv layers. In figure \ref{fig:fig5_12} pooling layer activation maps are also visualized. As mentioned before every Conv layer had 5 filter that means feature map was a volume of \((input\times5)\). In figure \ref{fig:fig5_12}, feature map volume generated by \(3^{rd}\) filter and middle three slices of the volume are shown for each layer. The first layer (Conv1) retained all the features of the input image which suggest that Conv filter was activated at every part of the image. Considering the fact that complex images are made up of small edges with different orientations all put together, preserving most of the features from input image pose an intuition that the filters might be acted like elementary edge detectors. Deeper layers feature maps appeared to be more like an abstract representation of the original input image. In \(2^{nd}\) Conv layer the input images were still somewhat recognizable but from \(3^{rd}\) Conv layer they were not because deeper feature maps became sparser so the filter detected less features of an input image. While deeper layers detected high level features, lower layers detected simple features. The \(4^{th}\) Conv layer held more about the class of the input and less about the input image. Activation map for every pooling layer followed by a Conv layer are also visualized in figure \ref{fig:fig5_12} indicating how the pooling operation transformed the input image by subsampling.

\noindent	
Figure \ref{fig:fig5_13} demonstrates feature maps of different volume of a layer. By conventions reason for using multiple filters in a layer is that different filter should learn different features. Some might get activation for vertical edges, some for horizontal or diagonal edges or some for complex edges. Looking at the different feature map volumes in figure \ref{fig:fig5_13} generated by different filter, it can be observed that not all the filters were activated to extract the same features. While volume 3 retained almost all the features from input, volume 1 filter was less activated for the same input. Feature maps for other trained models including different shaped filters yield comparable results. Appendix \ref{appendix:B.1} shows more feature map results.  	
	
	\subsection{Filters Visualization}\label{chapter:5.2.3 }
Filters/kernels are the main building block of a ConvNet. Feature maps are the results of a Conv layer where filters are used as learning parameters. Training of a ConvNet adjusts the filters so that they can extract features from an input. Filters of a well-trained convent learn to detect different features in different layers.    
		
	\begin{figure}[H]
    	\centering
		\includegraphics[width=0.75\textwidth]{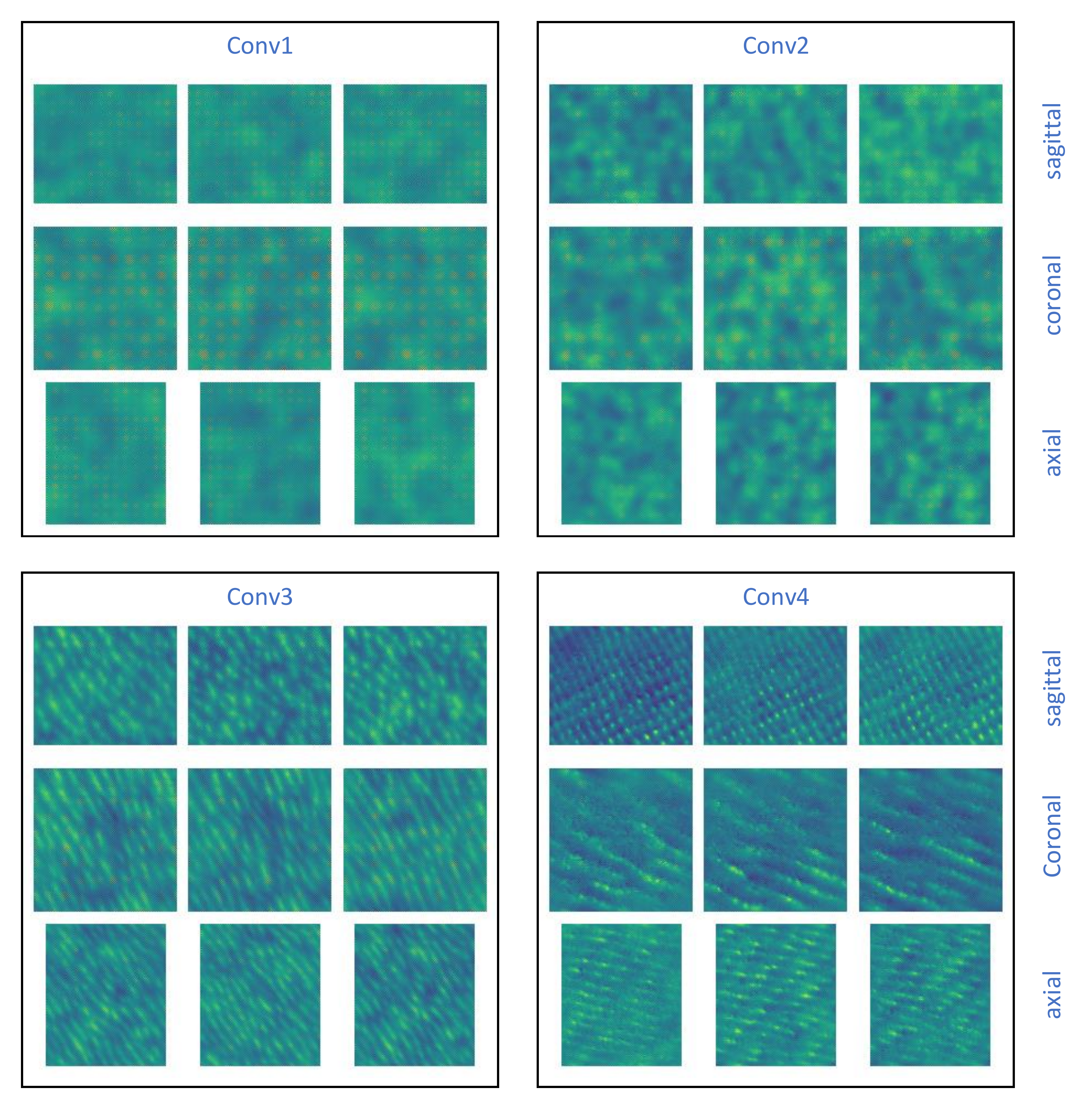}
		\caption[Filter visualization of each layer of ConvNet.]{Filter visualization of each layer of ConvNet. }
		\label{fig:fig5_14}
	\end{figure}

	 \begin{figure}[H]
    	\centering
		\includegraphics[width=\textwidth]{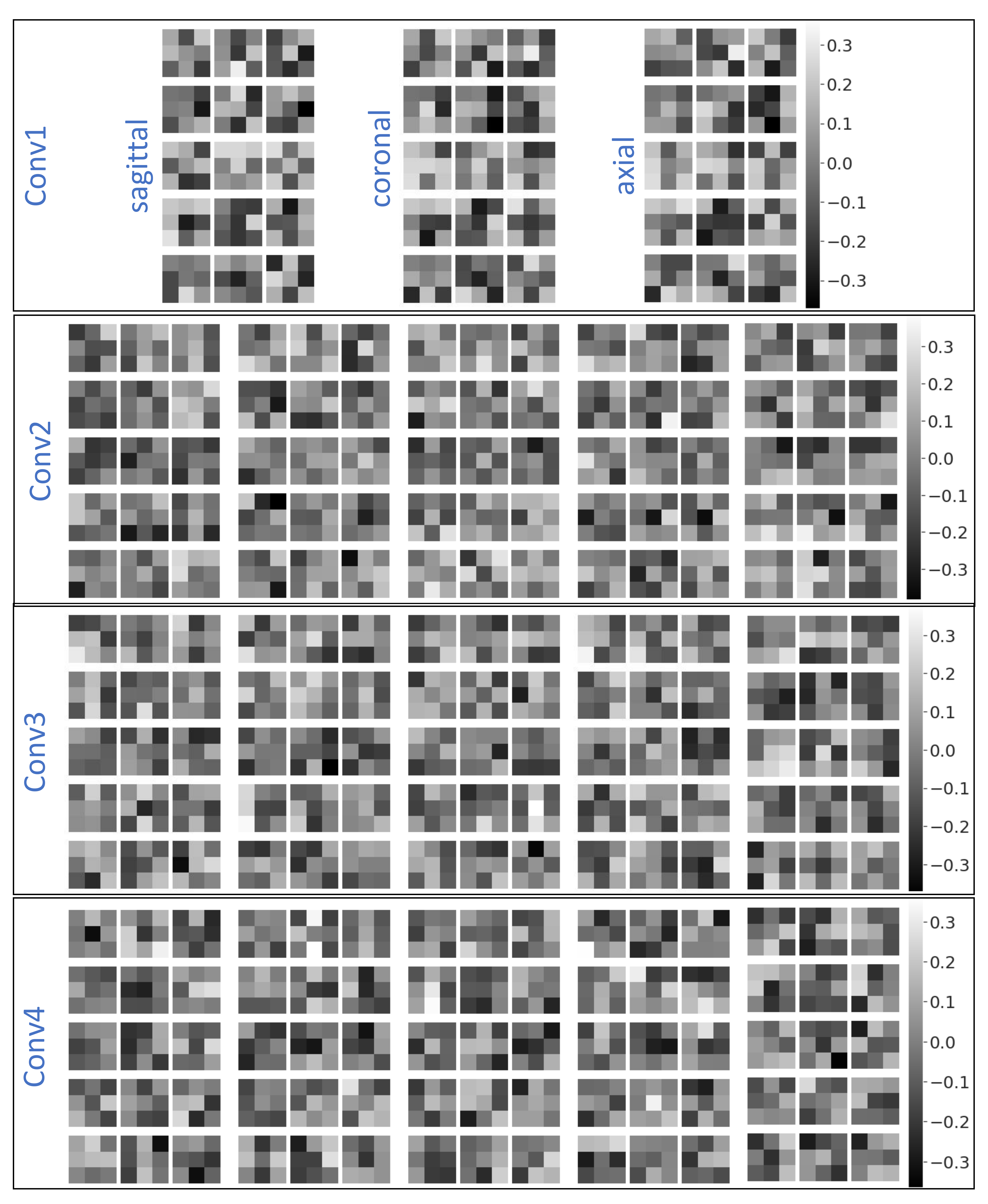}
		\caption[Leaned filters of all four Conv layers in the network. ]{Leaned filters of all four Conv layers in the network. }
		\label{fig:fig5_15}
	\end{figure}\vspace{2em}

In previous section, features maps of all Conv layers are visualized which were the results of applying some filters by convolution operation. Here those filters are discussed to get an intuition about their learning. The ConvNet that was used to visualize the features map in the previous section, is the same here for discussing the learned filters by the network. Although results of a particular model are presented here, other models that are mentioned in section \ref{chapter:5.2.1 } are also investigated. Results of those models can be seen in appendix \ref{appendix:B.2}. Each filter was a volume of size \((3\times3\times3)\) and there were 5 of them in every Conv layer. First Conv layer took a single 3D volume as input whereas \(2^{nd}\), \(3^{rd}\) and \(4^{th}\) Conv layer had to handle a 4D input \((3D volume\times5)\) generated by the previous layer. Thus, filters of \(2^{nd}\) to \(4^{th}\) layer were a volume of \((3\times3\times3\times5\times5)\). Figure \ref{fig:fig5_15} shows filters of all 4 Conv layers. Considering the MRI views, Conv1 filters are displayed for sagittal, coronal and axial separately while for other layers only coronal view of the filters is presented. Looking at the filters directly reveals very less information
about what they have learned. May be some of them can be recognized as a primitive edge detector but not all of them are interpretable. Alternative way of interpreting ConvNet filters is to use Activation Maximization (See section \ref{chapter:3.3 }). The process started with a random input image and tried to generate an output where the filters were more activated. Figure \ref{fig:fig5_14} displays images that are generated by activation maximization. For every Conv layer image generated by a randomly selected filter is presented with 3 random sagittal, coronal and axial slices. From figure \ref{fig:fig5_14} it can be observed that deeper blocks encoded complex pattern than the initial layers. These observations are relatable to the feature maps where early filters were activated on simple shapes and deeper filters were built on top of each other to encode complex features which are not general for input images. Although encoded patterns were not recognizable like state-of-the-art ConvNets where inputs are commonly known everyday object, these patterns were the indication that network was trained to learn some useful features. As mentioned in the section \ref{chapter:5.2.1 } that models with different architecture were trained to investigate their learning abstraction, filters of those models were also  visualized but only model trained with \((3\times3\times3)\) filters showed complex pattern in deeper layers like figure \ref{fig:fig5_14}.

	\subsection{Visualization with LRP}\label{chapter:5.2.4}	
\begin{figure}[H]
    	\centering
		\includegraphics[width=\textwidth]{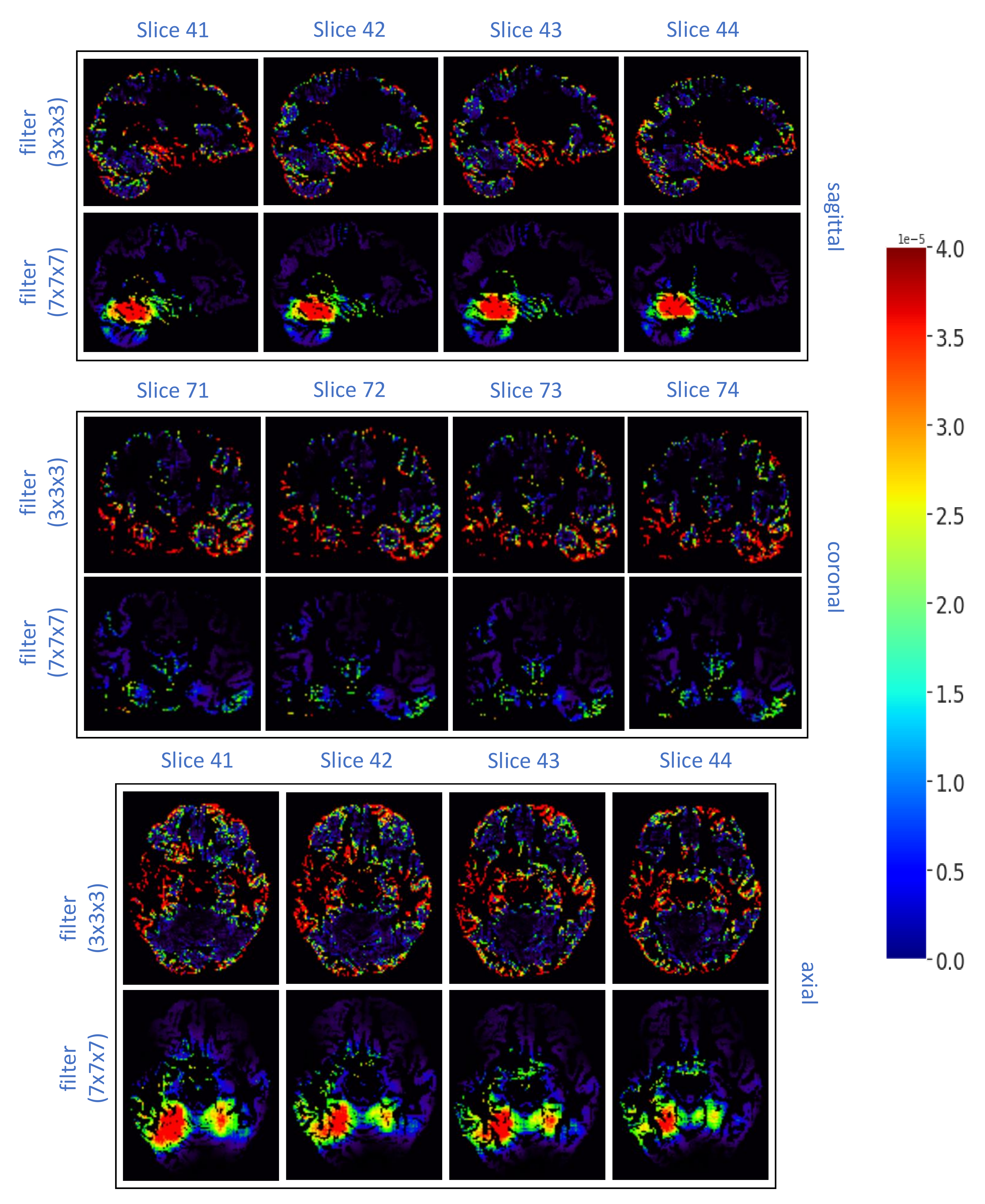}
		\caption[LRP heatmaps for ConvNet trained with filters of size \((3\times3\times3)\) and \((7\times7\times7)\).]{LRP heatmaps for ConvNet trained with filters of size \((3\times3\times3)\) and \((7\times7\times7)\).}
		\label{fig:fig5_16}
	\end{figure}

LRP is a decomposition-based method that calculated the relevance of each pixel of an image for classification decision. This study utilized LRP-\(\alpha_1\beta_0\)  rule for creating relevance heatmaps for ConvNets that were trained with different sets of network architecture. In table \ref{table:table5_2}, it appeared that using larger filter size increased the accuracy of the network although their training log was very unstable. Visualizing LRP maps of these networks revealed that using larger filters was not feasible for this study. Figure \ref{fig:fig5_16} shows the comparison of relevance maps between model trained with filters of size \((3\times3\times3)\) and \((7\times7\times7)\).   	

	\begin{figure}[H]
    	\centering
		\includegraphics[width=\textwidth]{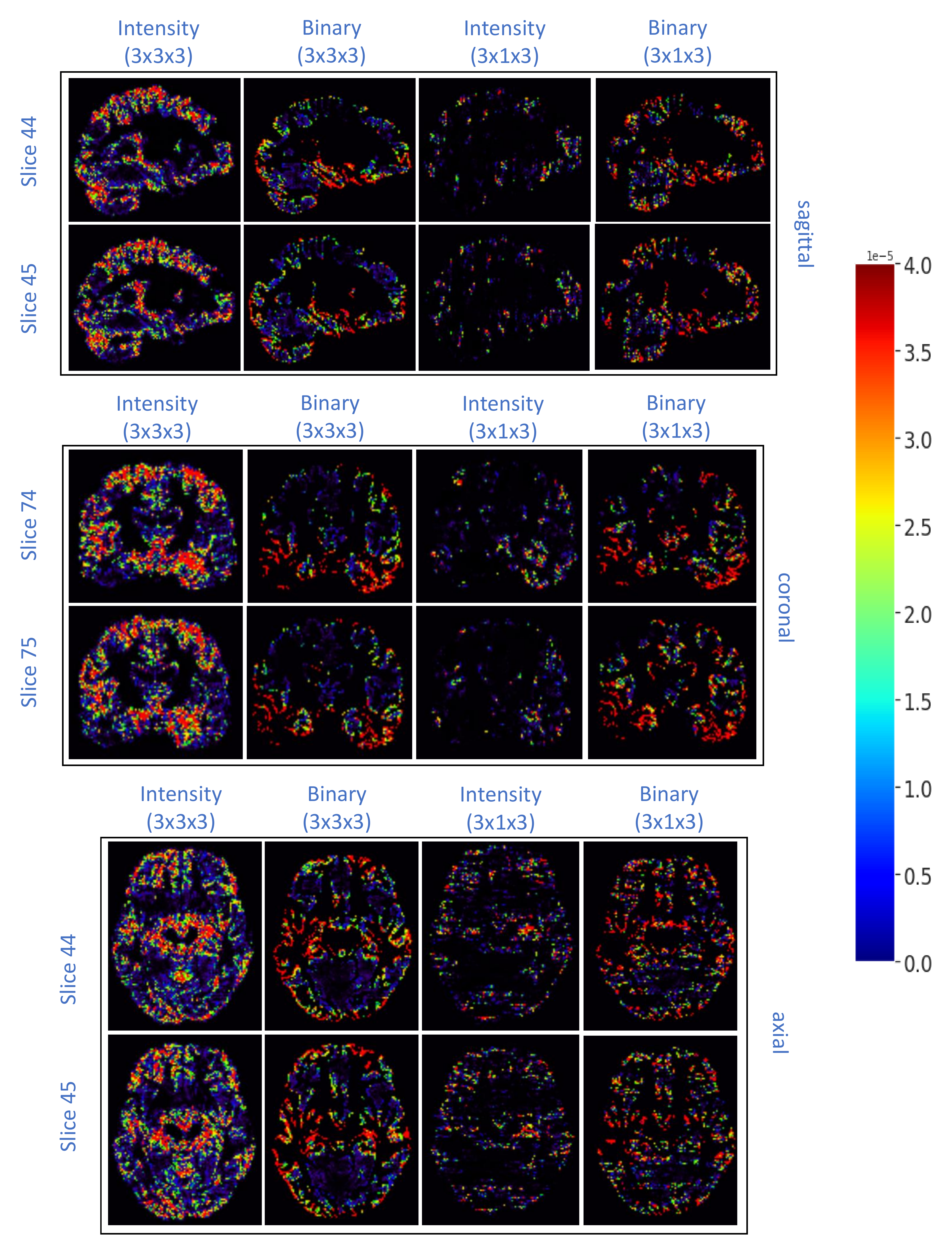}
		\caption[Relevance maps of intensity and binary voxel model with different filter shapes.]{Relevance maps of intensity and binary voxel model with different filter shapes.}
		\label{fig:fig5_17}
	\end{figure}

\noindent
The relevance maps of convent model trained with \((3\times3\times3)\) filter size showed that the voxels with high relevance were spread most of regions of the scan whereas the other model focused on some particular parts of the brain which were not relevant for AD classification. Four sagittal, coronal and axial slices for both models are visualized in the figure \ref{fig:fig5_16}. Slices were selected based on the hippocampus region of the brain. Looking at the slices of the brain relevance it can be observed that voxels near the temporal lobe had higher relevance which is desired for AD classification. As mentioned in data pre-processing part, brain MRI scans were binarized using a threshold of 0.4 that means all the voxels in the original scans were converted to either 1 or 0 depending on their intensity value. After binarizing, reduction of volume indicated by lower intensity values of brain region was lost. To check how binarization affected the relevance maps, same network was trained with intensity valued MRI scans. Figure \ref{fig:fig5_17} illustrates the LRP maps form two intensity scan models with two binary scan models. One model used filter size \((3\times3\times3)\) and other one used \((3\times1\times3)\). While accuracy of both models trained with intensity scans can be compared with binary scanned model, relevance map was less similar for \((3\times1\times3)\) filter model. The observation in the figure \ref{fig:fig5_17} reveals that intensity model with \((3\times3\times3)\) filter size has stronger relevance than the binary model of same filter size. But for filter size \((3\times1\times3)\) intensity model has less strong relevance than binary model. All the relevance maps presented in figure \ref{fig:fig5_16} \& \ref{fig:fig5_17} were for a true positive test subject with an accuracy of 99\%.

	\subsection{Transfer Learning with CAE }\label{chapter:5.2.5}	
\begin{figure}[H]
    	\centering
		\includegraphics[width=0.8\textwidth]{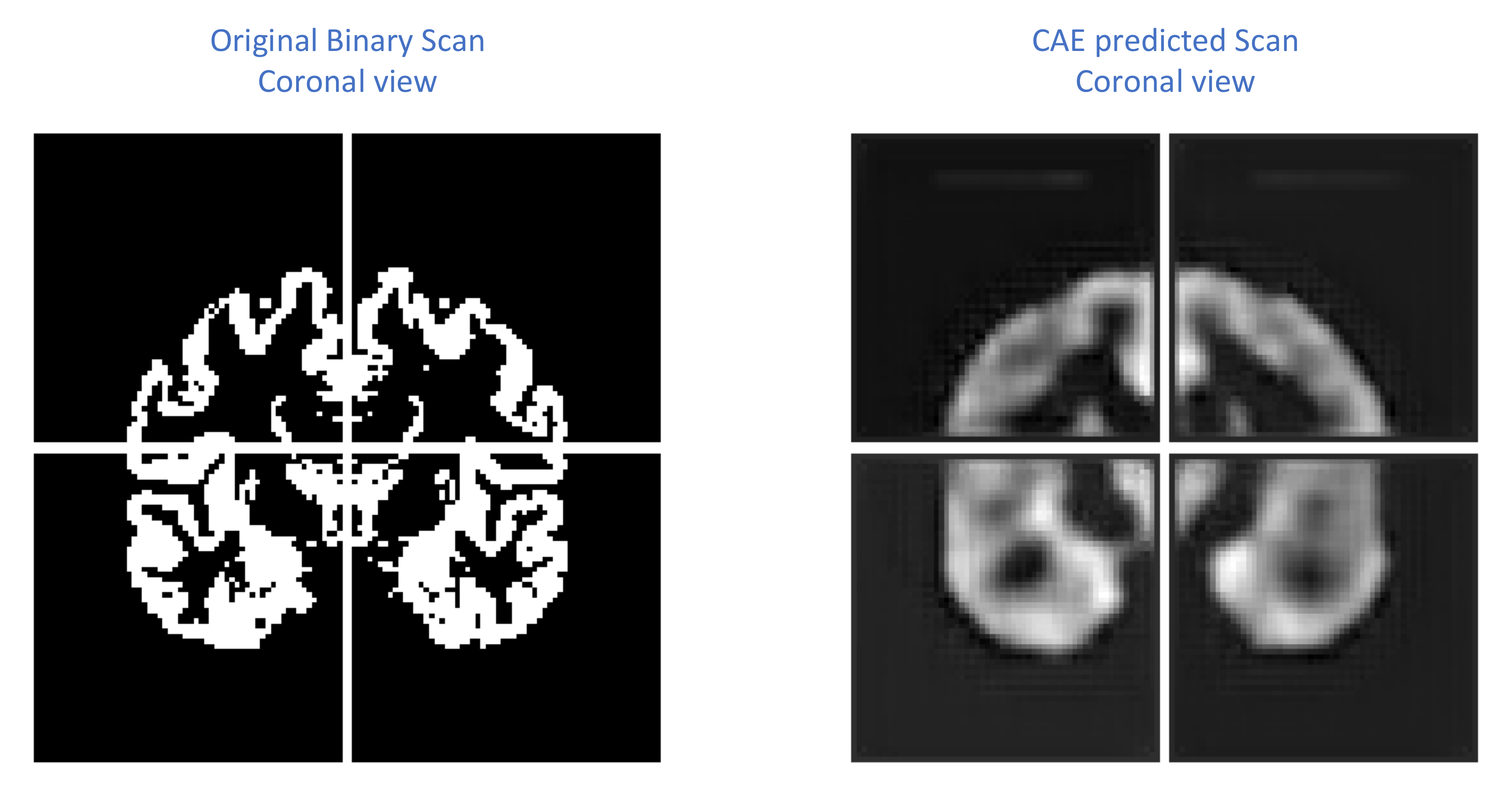}
		\caption[Coronal slice in patches original and predicted.]{Coronal slice in patches original and predicted.}
		\label{fig:fig5_18}
		
		\includegraphics[width=0.9\textwidth]{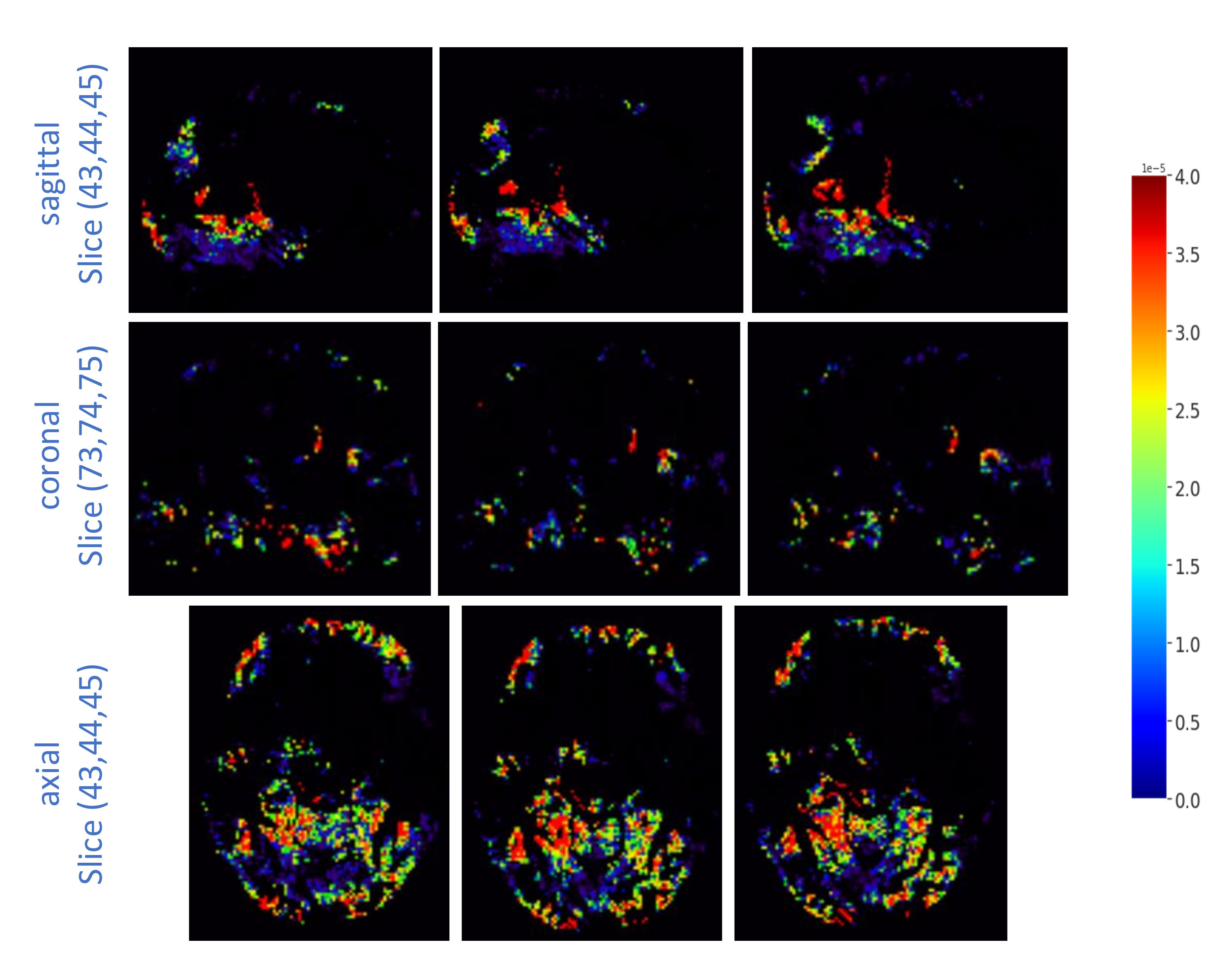}
		\caption[LRP visualization of ConvNet that uses pretrained weights from CAE.]{LRP visualization of ConvNet that uses pretrained weights from CAE. }
		\label{fig:fig5_19}
	\end{figure}

In section \ref{chapter:4.4}, implementation of a ConvNet has been discussed where first 3 Conv layers were transferred from encoder part of unsupervised CAE (figure \ref{fig:fig4_9}). The idea behind was to use the already learned filters from CAE that was trained on a larger data set. Data size was increased by dividing the brain MRI scan along coronal slice. Lower level filters usually do not look for distinct features of an image for classification decision rather detects the simple features that are common for all complex images such as edges in different orientation. This intuition intrigued the investigation of transferred weights from CAE to check whether it improves the learning of 3D ConvNet for AD classification task. Figure \ref{fig:fig5_18} shows the input patches of a coronal slice for training the CAE and reconstructed patches of the same slice from the decoder part of the network. The 3D ConvNet that used pretrained weights for the first 3 Conv layers from CAE and a fourth Conv layer for fine-tuning was trained with the same set of train, validation and test data. The hyperparameters for training were the same like other ConvNets mentioned in section \ref{chapter:5.1}. After training the network yielded an accuracy of 72.87\% which was quite similar (see section \ref{chapter:5.1}) to the 3D ConvNets without transfer learning. Relevance maps generated by this model with the same test subject as above showed less activated regions in coronal slices than the other models (Figure \ref{fig:fig5_19}).             	
	
\noindent
While axial slices showed activated region similar to previous models, sagittal slices were pointing to a region which less comparable. Learned features and feature map visualization were also comparable to the previously trained models.

	\chapter{Summary and Outlook  }\label{chapter:6 }	
In this section the overall findings based on different investigation strategies to achieve the thesis goal which was to learn the shape features and abstractions in 3D ConvNet for AD classification are summarized. In the previous chapter (chapter \ref{chapter:5}) results of the study are presented and discussed in details, here only the standout outcomes are reviewed in brief. 

Even though learned filters and LRP maps of ModelNet and 3D fruits CAD model revealed inconclusive shapes and features, the model showed very high accuracy \((98\%\pm1)\) in classification which was the motivation to apply the same network structure to the MRI data. However, training the same network structure resulted an accuracy of 71.32\% with very unstable training log over 50 epochs. An updated network structure with modified Conv blocks and added FC layers before final classification layer increased the accuracy (74.42\%) with a bit stable training log. It was also observed that increasing the filter sizes in the Conv layers improved the model accuracy but looked for only a particular part of the brain for the classification. Effect of changes in filter shape used in the Conv layers were also investigated. While model with filter shape \((3\times1\times3)\) reported similar accuracy, for models with \((1\times3\times3\)) and \((3\times3\times1)\) filter shape showed significant drop (8-10\%) in accuracy with the same training setup. Considering the orientation of MRI data this observation can give an intuition that 3D ConvNet has encoded more features along coronal slices thus hold more information for classification then other views (sagittal, axial).

Looking inside the learned filters and feature maps of the network suggested typical outcome from a ConvNet. In the first 2 Conv layers filters were looking for low level features and different filters were encoding different features of the input image which is expected because by convention different filters should get activated on different features such as edges of different orientation or other complex patterns. For better interpretation, filter visualization technique \textit{activation maximization} was used for different models described above but only model with \((3\times3\times3)\) shaped filters revealed some interesting pattern in deeper layers. Results from \textit{activation maximization} claimed that in deeper layers more complex structure were encoded then the lower layers in the network. Although patterns were not as interpretable as state-of-the-art models, finding complex patterns in deeper layers suggested that the network was learning as it supposed to be. 

Based on LRP relevance maps found for above mentioned models described how changing the filter size and shape affected relevant voxels in the MRI scan. While the relevance maps for model with \((3\times3\times3)\) filters showed strong relevance near temporal lobe and edges of the scan, larger filters focused on a certain region of the scans for classification. Model with \((3\times1\times3)\) filters also considered similar relevance in coronal slices. This study used binarized MRI scans for all implementation which allowed the entire scans of the brain to be used for shape and feature leaning. However, for LRP relevance map comparison, intensity valued voxel grid of MRI scan was used for training the same ConvNets that was used with binary valued scans. The observation between binary scan model and intensity scan model revealed that, intensity model had more activated relevance voxels than binary model which was reasonable because of the fact that binary model had a smaller number of voxels compared to intensity model. Despite having less voxels in the binary scans, relevance maps were comparable to intensity scan relevance map in the case of \((3\times3\times3)\) filter size, but \((3\times1\times3)\) filter size models for these two types of data had very different relevance map. Intensity model showed a smaller number of relevant voxels than binary model which showed same result as \((3\times3\times3)\) filter size model in coronal slices. 

Finally, transfer learning from a CAE which was trained with multiple patches on input binary scan to extract the low-level feature, was implemented. Even though number of training patches of MRI scan was over 30k, transferring the learned features for supervised learning did not improve accuracy. Moreover, LRP relevance maps for this model were less interpretable than the model trained without transfer learning. 

This study analyzed the state-of-the-art methods for 3D shape detection and learning the abstractions in 3D ConvNet for AD classification. Even though some valuable outcomes were observed there is much room for improvements. Binarizing the MRI scans removed lots of valuable information from the data so processing data with other algorithm and methodology can be investigated for shape detection. Due to limited computational resources and time, optimization of hyperparameters for training the network was not considered. By optimizing the network learning results from this study can be compared. Most obvious reason for not learning the shape features in the network as expected, was the lack of available data. State-of-the-art models are trained on millions of data which enables the earlier stages of a deeper network to learn more generalized features in the input space. Data available for this study was insignificant for learning shape features which are general across the input space. Moreover, there are less research in the domain of 3D shape feature extraction. Even though shapes were encoded there was no standard way to represent the encoded 3D shape, so this study always relied on 2D shape representation techniques. Investigating the way to represent the 3D shape encoded by a 3D ConvNet can be a major research topic in computer graphics. The issue of limited explainability and knowledge extraction from deep learning models has also been a new target topic of German Research Foundation (DFG, Deutsche Forschungsgemeinschaft) \cite{DFGDeep}.

			
	\clearpage\pagenumbering{Roman}
	\bibliographystyle{unsrtdin}
	\bibliography{library}
	
	\appendix
	\chapter{Description of Used Modules}\label{appendix:A}
This appendix acknowledges the modules and tools that were used for various implantation of the thesis. Section \ref{chapter:4.5} described the training environment and purpose of these tools in brief. Here, those modules and tools along with some other modules are mentioned that were crucial for this thesis work. 

	\begin{table}[H]
	\centering
	\begin{tabular}{||l|l|l|l||}
	\hline
	{\vtop{\hbox{\strut Module} \hbox{\strut Name}}} & Version & Description & URL \\ \hline\hline
	Keras       & 2.3.1 & {\vtop{\hbox{\strut Deep learning} \hbox{\strut framework}}}  & \url{https://keras.io/}\\ \hline
	Tensorflow	& 1.15.2 & {\vtop{\hbox{\strut Deep learning} \hbox{\strut backend}}}  & \url{https://www.tensorflow.org/}\\ \hline
	NiBabel     & 3.0.2 & {\vtop{\hbox{\strut read/write} \hbox{\strut MRI files}}}  & \url{https://nipy.org/nibabel/}\\ \hline
	Keras-vis & 0.4.1	& {\vtop{\hbox{\strut Filter} \hbox{\strut visualization}}}  & \url{https://github.com/raghakot/keras-vis}\\ \hline
	iNNvestigate & 1.0.8	& {\vtop{\hbox{\strut LRP} \hbox{\strut visualization}}}  & \url{https://pypi.org/project/innvestigate/}\\ \hline
	Scikit-learn & 1.0.8	& {\vtop{\hbox{\strut Data} \hbox{\strut preparation}}}  & \url{https://scikit-learn.org/stable/}\\ \hline
	Numpy & 1.18.5	& {\vtop{\hbox{\strut Array} \hbox{\strut manipulation}}}  & \url{https://numpy.org/}\\ \hline
	Matplotlib & 3.3.0	& {\vtop{\hbox{\strut Data} \hbox{\strut visualization}}}  & \url{https://matplotlib.org/}\\ \hline
	Seaborn & 0.10.1	& {\vtop{\hbox{\strut Data} \hbox{\strut visualization}}}  & \url{https://seaborn.pydata.org/}\\ \hline
	H5py & 2.10.0	& {\vtop{\hbox{\strut Data} \hbox{\strut storing}}}  & \url{https://www.h5py.org/}\\ \hline
	 	
 	\end{tabular} 	
 	
	\caption[Most used modules to implement various models for the thesis.]{Most used modules to implement various models for the thesis.}
  	\label{table:tableA_1}
	\end{table}

	\chapter{Result Visualization}\label{appendix:B}
Here the visualization results of some models that was used throughout the this work has been shown. 
	\section{Feature maps}\label{appendix:B.1}
Feature maps of model trained with \((1\times3\times3)\), \((3\times1\times3)\), \((3\times3\times1)\) filter shape are shown here. Only the inner most slices of \(3^{rd} \) volume of 4D feature maps of a Conv layer are visualized.

	\begin{figure}[H]
    	\centering
		\includegraphics[width=\textwidth]{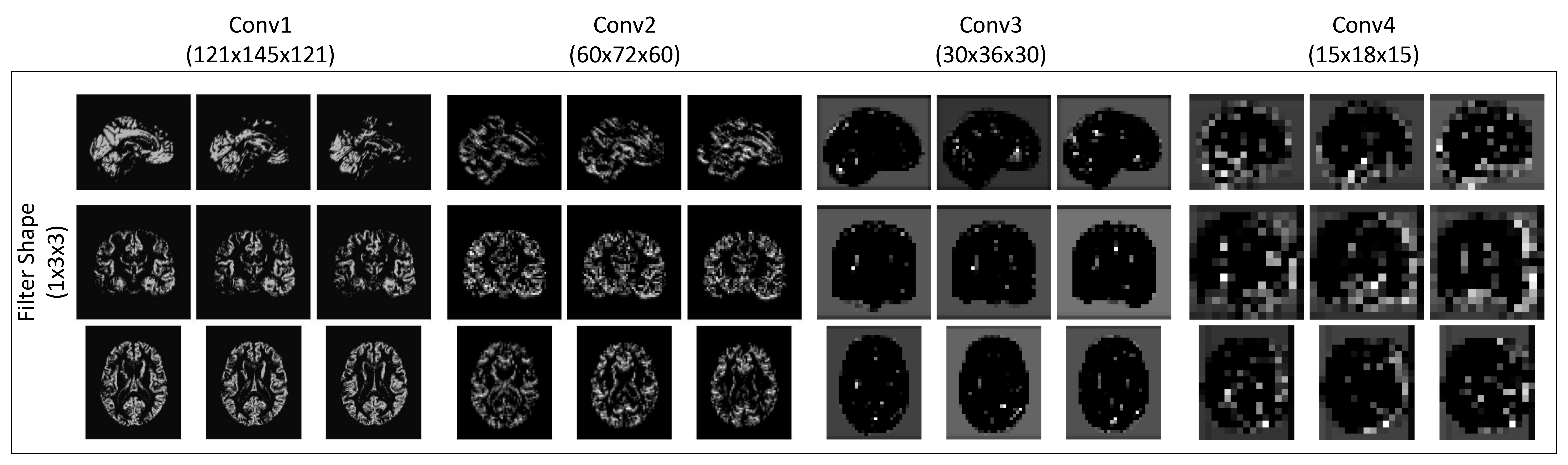}
		\caption[Feature maps of model trained with \((1\times3\times3)\) filters.]{Feature maps of model trained with \((1\times3\times3)\) filters.}
	\label{fig:figB_1}
	\end{figure}

	\begin{figure}[H]
    	\centering
		\includegraphics[width=\textwidth]{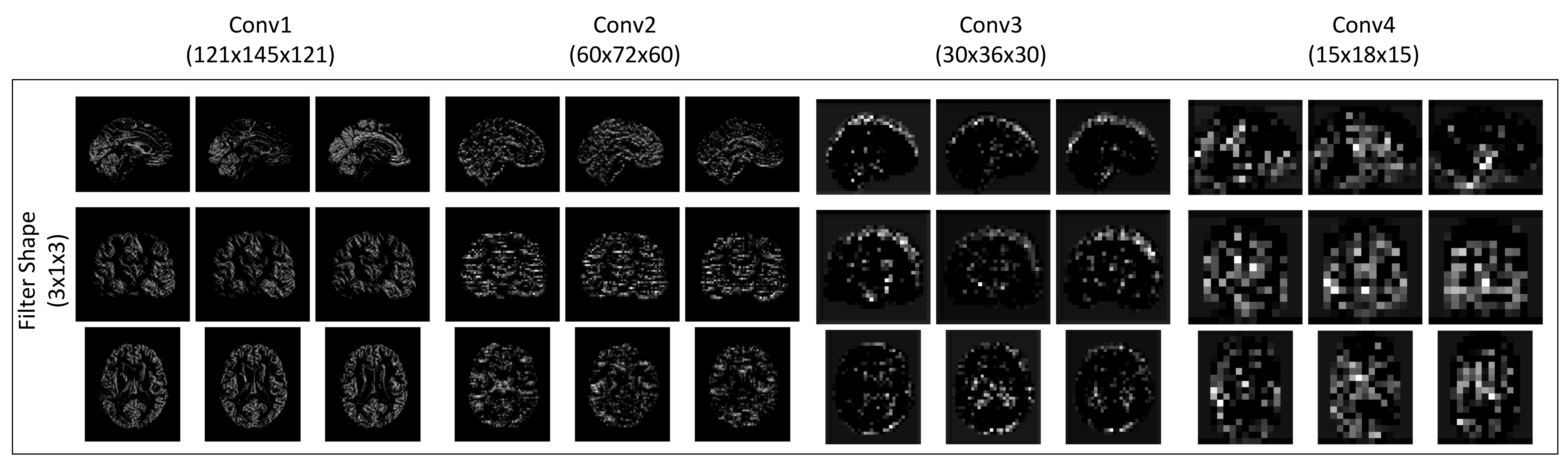}
		\caption[Feature maps of model trained with \((3\times1\times3)\) filters.]{Feature maps of model trained with \((3\times1\times3)\) filters.}
	\label{fig:figB_2}
	\end{figure}
	
	\begin{figure}[H]
    	\centering
		\includegraphics[width=\textwidth]{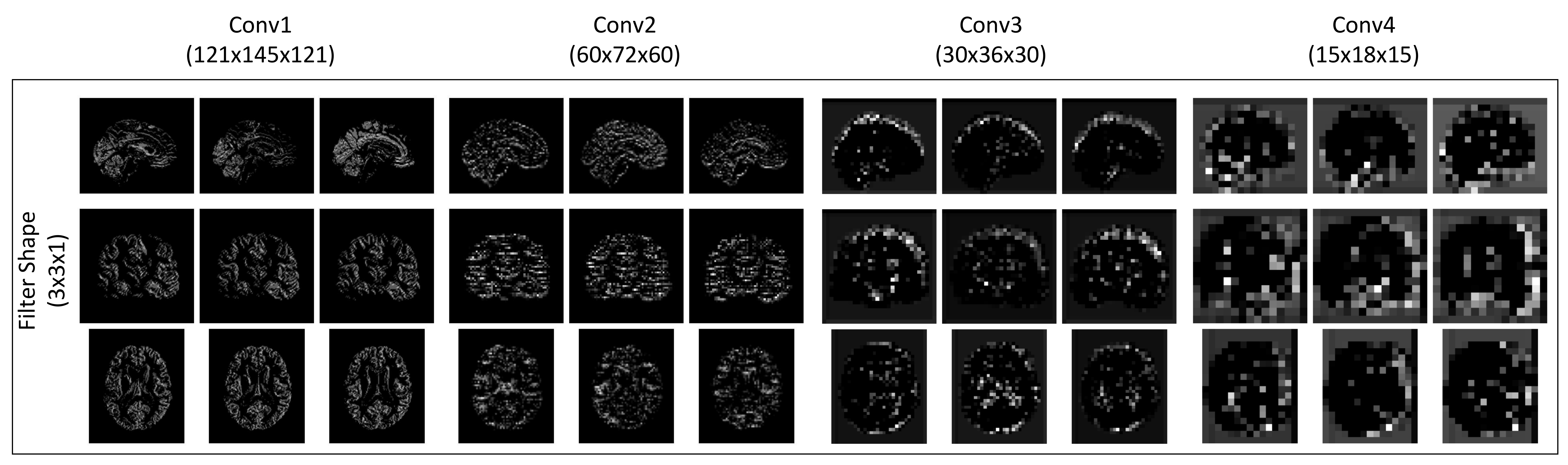}
		\caption[Feature maps of model trained with \((3\times3\times1)\) filters.]{Feature maps of model trained with \((3\times3\times1)\) filters.}
	\label{fig:figB_3}
	\end{figure}

	\section{Filters visualization}\label{appendix:B.2}
Filter visualization with \textit{Activation Maximization} of model trained with \((1\times3\times3)\), \((3\times1\times3)\), \((3\times3\times1)\) filter shape are shown here. The inputs were generated with a random noise and choice of filters per layer were also random. Visualized slices for each input in a layer are selected randomly as well.   

	\begin{figure}[H]
    	\centering
		\includegraphics[width=\textwidth]{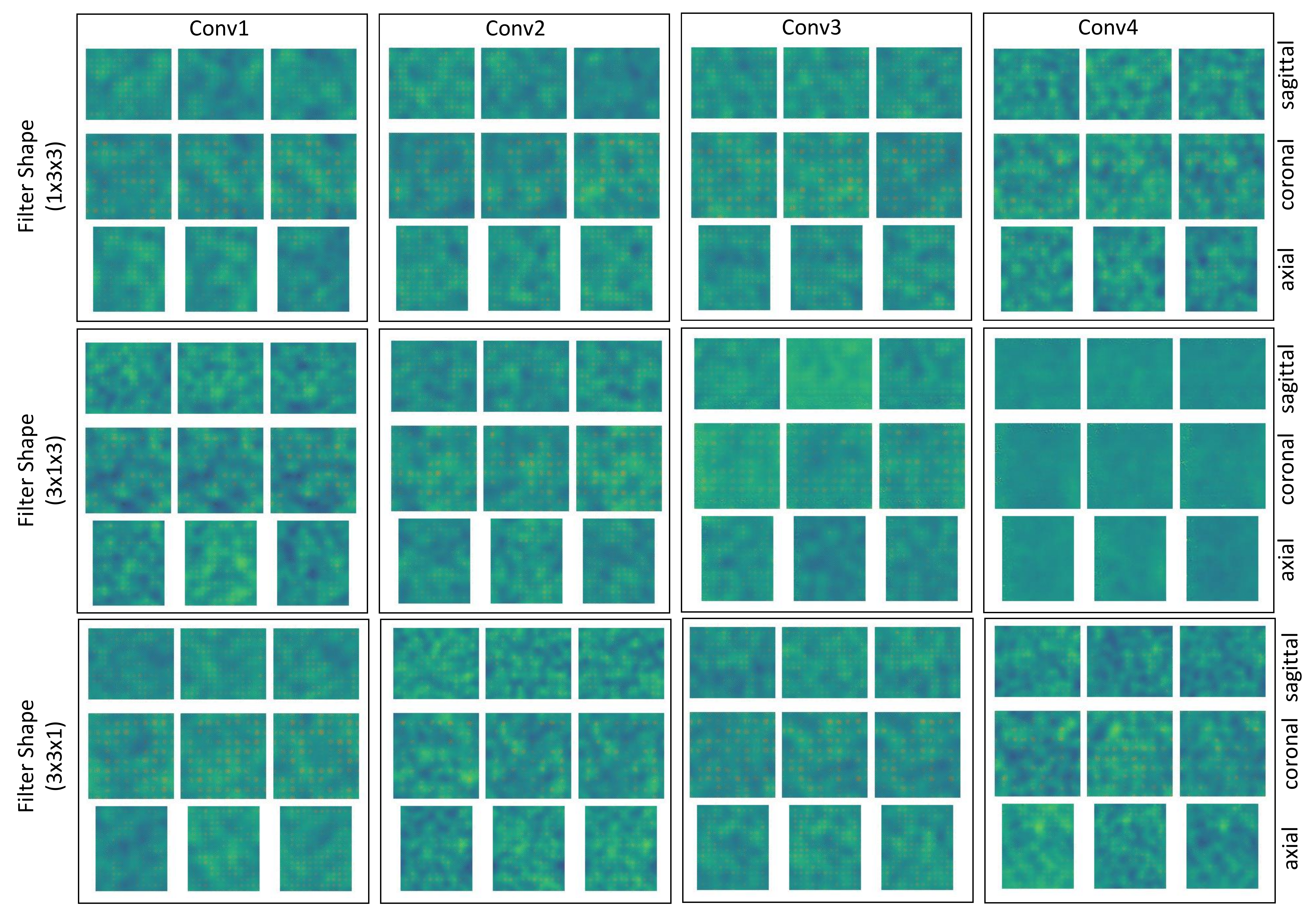}
		\caption[Activation Maximization output of different shaped filters.]{Activation Maximization output of different shaped filters.}
	\label{fig:figB_4}
	\end{figure}

\end{document}